%% file: main.tex
\documentclass{article} 
\usepackage{iclr2027_conference,times}
\iclrfinalcopy

\input{math_commands.tex}

\usepackage{hyperref}
\usepackage{url}
\usepackage{booktabs}
\usepackage[most]{tcolorbox}

\usepackage{amsmath}
\usepackage{amssymb}
\usepackage{tabularx}
\usepackage{marvosym}
\usepackage{xcolor}
\usepackage{pifont}
\usepackage{enumitem}

\usepackage{graphicx}
\usepackage{subcaption}
\usepackage{adjustbox}
\usepackage{multirow}
\usepackage{multicol}
\usepackage{rotating}
\usepackage{wrapfig}
\usepackage{caption}
\usepackage{minted}
\usepackage{listings}
\usepackage{float}
\usepackage{fvextra}
\usepackage{array}
\usepackage{makecell}
\usepackage{vcell}
\usepackage{ragged2e}
\usepackage{changepage}
\usepackage{xurl}
\usepackage[T1]{fontenc}
\usepackage[utf8]{inputenc}
\usepackage[greek,english]{babel}

\usepackage[toc,page,header]{appendix}
\usepackage{minitoc}

\doparttoc 
\faketableofcontents 

\newcommand{\factualerr}[4]{%
\shortstack{
$#1 {\scriptstyle \pm #2}$ \\
[-1pt]
{\scriptsize $(#3 \pm #4)$}
}}

\newcommand{\bestfactualerr}[4]{%
\shortstack{
$\mathbf{#1} {\scriptstyle \mathbf{\pm #2}}$ \\
[-1pt]
{\scriptsize $(#3 \pm #4)$}
}}

\newcommand{\resultcell}[2]{%
\shortstack{
$#1$ \\
[-1pt]
{\scriptsize $\pm #2$}
}}

\newcommand{\bestresultcell}[2]{%
\shortstack{
$\mathbf{#1}$ \\
[-1pt]
{\scriptsize $\mathbf{\pm #2}$}
}}

\newcommand{\daterange}[2]{%
\shortstack{
{\scriptsize #1} 
{\scriptsize -- #2}
}}

\title{\textit{What If TSF}: Evaluating Scenario-Guided Multimodal Time Series Forecasting}

\author{\NoHyper \noindent
Jinkwan Jang$^{*}$ \,\, Hyunbin Jin$^{*}$ \,\, Hyungjin Park \,\,\,\, Kyubyung Chae \,\,\,\, Taesup Kim$^{\dagger}$\\[0.3em]
Graduate School of Data Science, Seoul National University\\
\endNoHyper
}
\begin{document}

\maketitle
\begingroup
\makeatletter
\NoHyper
\renewcommand\thefootnote{*}
\footnotetext{
    Equal contribution.\hspace{1em} $^{\dagger}$Corresponding author.
}
\endNoHyper
\makeatother
\endgroup

\input{scripts/0_abstract}
\input{scripts/1_introduction}
\input{scripts/2_related_work}

\input{scripts/3_WIT}

\input{scripts/4_evaluation}
\input{scripts/5_experiments}
\input{scripts/6_conclusion}

\newpage

\subsection*{AI use statement}



Generative AI tools were used for language polishing and stylistic refinement of the manuscript, and for selected benchmark-construction tasks, including source-grounded evidence extraction and factual scenario composition.
AI-generated content was used only as intermediate assistance and not as a source of factual evidence.
The authors take full responsibility for the final benchmark artifacts, manuscript content, and reported results.

\subsection*{Ethics statement}



WIT is intended as a research benchmark for evaluating scenario-guided forecasting and not as a decision-making system for financial, political, or policy applications.
Source coverage and scenario construction may introduce biases, so benchmark results should be interpreted as comparative research evaluations rather than evidence for real-world decisions or policy actions.

\subsection*{Reproducibility statement}



Section~\ref{sec:wit_construction} and Appendices~\ref{app:sources_and_stats} and~\ref{app:construction_protocol} describe the data sources and benchmark-construction protocol, including reference-guided candidate screening, factual scenario construction and anchor selection, alternative and irrelevant scenario construction and validation, and empirical matched-set construction.
Section~\ref{sec:eval_details} and Appendix~\ref{app:eval_details} describe the evaluation metrics, aggregation rules, and sensitivity analyses.
Section~\ref{sec:experiments} and Appendix~\ref{app:exp_details} document the evaluated model configurations, knowledge cutoffs, inference settings, and repeated-run protocol.
Prompt templates and model inputs are detailed in Appendix~\ref{app:prompts_inputs}.
Together, these specifications support reproducibility of WIT construction and evaluation.




\bibliography{reference}
\bibliographystyle{iclr2027_conference}

\newpage

\renewcommand \thepart{}
\renewcommand \partname{}

\appendix

\addcontentsline{toc}{section}{Appendix} 

\part{} 
{%
  \vspace{-3.5em}
  \renewcommand{\baselinestretch}{1.1}\selectfont 
  \Large 
    \setcounter{tocdepth}{3}
  \setcounter{parttocdepth}{3}

  \parttoc
}

\newpage

\input{appendix/a_data}
\input{appendix/b_construction}

\input{appendix/c_prompt}
\input{appendix/d_eval_detail}
\input{appendix/e_experiment}

\input{appendix/f_analysis}

\input{appendix/g_limitation}

\end{document}

%% file: math_commands.tex
\usepackage{amsmath,amsfonts,bm}

\def\eqref#1{equation~\ref{#1}}

\def\1{\bm{1}}

\DeclareMathAlphabet{\mathsfit}{\encodingdefault}{\sfdefault}{m}{sl}
\SetMathAlphabet{\mathsfit}{bold}{\encodingdefault}{\sfdefault}{bx}{n}



%% file: scripts/0_abstract.tex
\vspace{-20pt}
\begin{abstract}

Recent advances in large language models (LLMs) have enabled time series forecasting to move beyond numerical observations and incorporate external information in multimodal settings. 
Such information can improve forecasting performance, but accuracy alone may not reveal whether models appropriately respond to it: a model may ignore relevant information, fail to distinguish scenarios with different implications, or overreact to irrelevant or weak signals. 
We introduce \textit{What If TSF (WIT)}, a benchmark for evaluating whether models effectively respond to future scenarios.
WIT constructs \emph{controlled scenario sets} by fixing the forecasting state while systematically varying future scenarios, enabling three complementary evaluations: 
\emph{Factual Comparison}, which measures the predictive utility of factual scenario information; 
\emph{Relational Comparison}, which evaluates whether forecasts satisfy expected direction, contrast, intensity, and restraint relations across scenarios;
and \emph{Grounded Comparison}, which provides a complementary empirical assessment of scenario-induced revision trajectories where comparable real cases are available.
Experiments show that factual accuracy gains do not consistently translate into appropriate responses to alternative scenarios, demonstrating the need to evaluate multimodal forecasting beyond predictive accuracy.

\end{abstract}

%% file: scripts/1_introduction.tex
\section{Introduction}

Most forecasting methods infer future dynamics primarily from historical numerical observations, either by modeling dependencies across observed variables or by learning transferable temporal patterns at scale with time-series foundation models (TSFMs)~\citep{liu2024itransformer, wang2025timemixer++, jang2026towards, liu2025moirai2, liu2025sundial, liu2026falconx, khwaja2026toto2}.
Recently, covariate-aware TSFMs have extended this scope beyond historical observations by conditioning forecasts on known future covariates, such as calendar features and planned interventions available over the prediction horizon~\citep{ansari2025chronos2, jain2026timesfm3, podest2026tirex2, meyer2026t0}.
While this substantially broadens the information available to numerical forecasters, many consequential future drivers are difficult to represent as numerical covariates or may simply not be captured by the available covariate set.
Information about expected shifts in supply or demand, as well as forecasts of unusual weather conditions, may instead be available through reports or news articles even when the time series itself provides no indication of such external changes.
Such text can describe future conditions without specifying the magnitude, onset, or duration of the target response.
Models must therefore use domain knowledge to interpret these conditions and translate their implications into numerical forecast revisions.

Recent advances in large language models (LLMs) and multimodal alignment have made it increasingly feasible to incorporate natural-language context into time series forecasting~\citep{kong2025achieving}.
A growing body of work explores prompting, representation fusion, and context-aware forecasting methods that allow external information to influence numerical predictions~\citep{requeima2024llmpro, jia2024gpt4mts, jin2024timellm, liu2024unitime, li2026language, wu2026aurora, das2026nexus}.
Benchmarks such as Time-MMD~\citep{liu2024timemmd} and TimesX~\citep{liu2026rethinking} further support this direction by providing richer and better-aligned textual context.
Together, these developments have established an important premise: external context can provide predictive information beyond numerical extrapolation and can potentially improve forecasting performance.

Yet, a model can predict the observed future more accurately without necessarily reflecting what the provided context implies.
A factual scenario may reduce forecast error relative to a no-scenario baseline, establishing its predictive utility without showing whether the forecast reflects the implications of that scenario. 
A model can therefore perform well on factual forecasting while still failing to distinguish how alternative contexts should change its forecasts.
Thus, even when context is genuinely relevant and useful for prediction, an important question remains: \textit{when only the future scenario changes, does the forecast revise in a way that is consistent with that scenario?}

Recent work has begun to move beyond factual context evaluation. 
CiK~\citep{williams2025cik} evaluates forecasting with future information, constraints, and other types of context, while TADiff~\citep{gu2026what} explicitly studies counterfactual forecasting under alternative future conditions.
However, these settings do not directly isolate scenario-induced forecast changes by holding the forecasting state fixed and comparing each scenario against the same neutral forecast.
For example, a scenario with an upward implication should induce an upward revision, opposing scenarios should produce contrasting revisions, stronger versions of the same condition should produce larger revisions, and irrelevant information should leave the forecast largely unchanged.
This motivates evaluating not only the predictive utility of context, but also whether forecasts respond consistently across scenarios and remain empirically grounded.

\begin{figure*}[t!]
  \centering
  \vspace{-12pt}
  {\includegraphics[width=\linewidth]{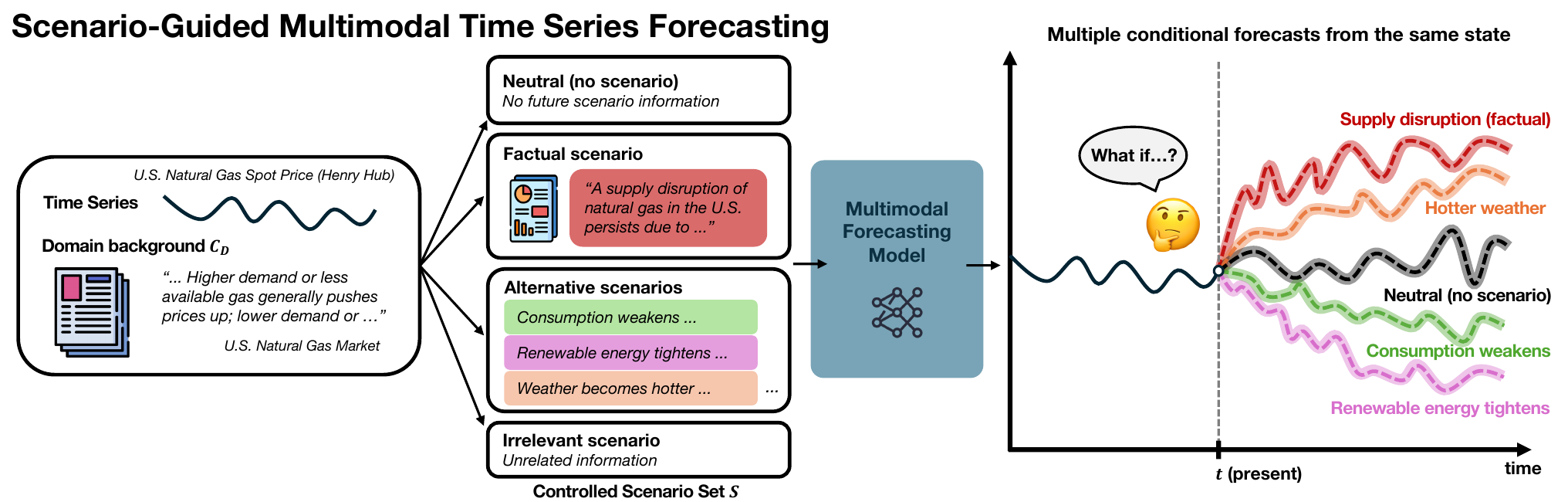}}
    \caption{{Formulation of scenario-guided multimodal time series forecasting.} The numerical history and domain background remain fixed while future scenarios vary, allowing comparison of scenario-conditioned forecasts under the same forecasting state.}
  \label{fig1:scenarios}
  \vspace{-14pt}
\end{figure*}

We introduce \textit{What If TSF (WIT)}~\footnote{Data: \url{https://github.com/jinkwan1115/WhatIfTSF}}, a benchmark designed to evaluate whether forecasts reflect the implications of contextual information under controlled future scenarios.
WIT evaluates forecasts using \textbf{controlled scenario sets}, holding the numerical history and domain background context fixed while systematically varying the future scenario (See Figure~\ref{fig1:scenarios}). 
By expanding evaluation from individual text-conditioned forecasts to structured comparisons under the same forecasting state, WIT assesses both factual predictive utility and consistency in scenario-induced forecast responses.

WIT covers Energy, Economy, and Politics, where interpreting scenario implications requires reasoning about supply and demand, relative macroeconomic conditions, and how voters assign responsibility and respond to political competition, respectively.
We examine scenario-guided forecast revisions under distinct mechanisms.
In these domains, multiple interacting factors make it difficult to determine the magnitude and timing of the target's response to individual future conditions.
For alternative scenarios, WIT therefore uses qualitatively justified relations to evaluate forecast revisions without requiring exact numerical responses or counterfactual trajectories as ground truth.
WIT follows a common construction procedure that supports extension to additional targets.



WIT evaluates these forecasts through three complementary comparisons.
\textbf{Factual Comparison} asks how accurately the model predicts the observed future, how much factual scenario information reduces forecast error on average relative to the model's neutral forecast, and how often it does so.
\textbf{Relational Comparison} tests whether scenario-induced revisions relative to the model's neutral forecast satisfy explicit expectations about \emph{Direction}, \emph{Contrast}, \emph{Intensity}, and \emph{Restraint}.
Finally, \textbf{Grounded Comparison} assesses how closely these neutral-relative revision trajectories align with deviations observed in comparable real cases.
Together, these comparisons evaluate not only whether context improves observed-future prediction, but whether forecasts respond coherently and in empirically plausible ways as future scenarios change.

%% file: scripts/2_related_work.tex
\section{Related Work}
\label{sec:related_work}

\input{tables/table-comparison}

\paragraph{Context-aware forecasting datasets and benchmarks.}

Time-MMD~\citep{liu2024timemmd} temporally aligns real-world time series with text collected from reports and web search results, evaluating text-conditioned forecasts against numerical-only baselines.
CiK~\citep{williams2025cik} manually constructs context across five information types---intemporal, future, historical, covariate, and causal---and evaluates context effects using Region of Interest (RoI)-weighted errors and constraint penalties, including removal and irrelevant or misleading context.
Its tasks use both observed and synthetic numerical futures.
GIFT-CTX~\citep{yan2026moiraiagent} combines selected CiK examples with additional synthetic cases designed so that accurate forecasting requires both numerical history and textual context.
CAF-7M~\citep{zheng2026overcoming} generates synthetic context from observed histories and futures, retains cases where context improves forecasting, and compares original, absent, and swapped context.
TimesX~\citep{liu2026rethinking} constructs context from observed sources, organizing metadata, calendar, covariate, and event information and evaluating their incremental effects through forecast errors.
TADiff~\citep{gu2026what} extends evaluation to factual and alternative scenarios under a shared numerical history.
For real-world data, alternative conditions are selected from other samples, while its Synth dataset provides synthetic futures for alternative scenarios.
Its evaluation combines forecast error with consistency scoring that measures the alignment of forecasts with intrinsic historical features and extrinsic future conditions.

\paragraph{Understanding and improving text use in multimodal forecasting.}
TESS~\citep{li2026tess} generates text describing statistical features extracted from observed future windows in its diagnostic experiments and reports that models often assign more attention to redundant tokens than to predictive ones, even among samples where text yields positive predictive gains.
WIT shares the concern that accuracy gains alone provide an incomplete picture of text use and builds on it by examining relations among numerical forecast revisions across scenarios at the same anchor.
CAMEF~\citep{zhang2025camef} constructs alternative event descriptions by modifying numerical values and sentiment-related wording and uses them in contrastive learning.
WIT uses alternative scenarios as evaluation inputs to test predefined relations among the resulting forecast revisions.

\paragraph{Positioning WIT.}
Table~\ref{tab:comparison} compares prior work in terms of context comparisons, context effect evaluation, and numerical references.
Among different types of context, scenarios are especially important because they describe future conditions that may affect the forecast.
Prior benchmarks show that context can improve forecast accuracy, but forecast error provides limited insight into how a model responds across scenarios, such as the direction, relative ordering, strength, and restraint of its revisions.
Moreover, when an alternative scenario did not actually occur, its synthetic future must be constructed under additional assumptions, making the resulting trajectory difficult to justify or validate against real observations.

WIT therefore keeps numerical history and domain background context fixed, compares scenario-induced revisions against the same neutral forecast, and evaluates their relations using real-case-informed alternatives.
These responses are further grounded by comparing their neutral-relative revision trajectories with deviations observed in matched real cases, without relying on synthetic alternative futures.

%% file: tables/table-comparison.tex
\begin{table*}[t]
\centering
\caption{
Comparison with prior context-aware forecasting datasets and benchmarks.
\textbf{WIT evaluates scenario responses relative to the same model's neutral forecast and, for supported pairs, assesses their empirical alignment with deviations observed in matched real cases.}
}
\label{tab:comparison}

\begingroup
\footnotesize
\setlength{\tabcolsep}{2.5pt}
\renewcommand{\arraystretch}{1.12}
\renewcommand{\tabularxcolumn}[1]{m{#1}}

\begin{tabularx}{\textwidth}{@{}
>{\centering\arraybackslash}m{0.12\textwidth}
>{\centering\arraybackslash}m{0.30\textwidth}
>{\centering\arraybackslash}m{0.30\textwidth}
>{\centering\arraybackslash}X
@{}}
\toprule

\shortstack{\textbf{Dataset /}\\\textbf{Benchmark}} &
\textbf{Context comparisons} &
\textbf{Context effect evaluation} &
\textbf{Numerical reference} \\

\midrule
\midrule
\addlinespace[3pt]
Time-MMD &
\shortstack{Whole context} &
Error &
Observed \\

\addlinespace[3pt]

CiK &
\shortstack{Types + irrelevant or misleading} &
\shortstack{RoI$^{\ast}$ error + penalties} &
\shortstack{Observed / synthetic} \\

\addlinespace[3pt]

GIFT-CTX &
\shortstack{Whole context} &
RoI$^{\ast}$ error &
\shortstack{Observed / synthetic} \\

\addlinespace[3pt]

CAF-7M &
\shortstack{Whole context + swapped} &
Error &
Observed \\

\addlinespace[3pt]

TimesX &
\shortstack{Types} &
Error &
Observed \\

\addlinespace[3pt]

TADiff &
\shortstack{Scenarios; factual / alternative} &
Error, consistency &
\shortstack{Observed / synthetic} \\

\midrule

\textbf{WIT} &
\shortstack{
\textbf{Scenarios\,(each vs.\ neutral);}\\
\textbf{factual / alternative / irrelevant}
} &
\shortstack{
Error, win rate$^{\dagger}$,\\
\textbf{neutral-relative relations}
} &
\shortstack{
\textbf{Observed}\\
\textbf{(Empirical matched sets)}
} \\

\bottomrule
\end{tabularx}

\raggedright
\begin{minipage}{\textwidth}
\fontsize{7pt}{9.5pt}\selectfont
\textsuperscript{\(\ast\)}Region of Interest; CiK additionally includes constraint penalties. \\
\textsuperscript{\(\dagger\)}Win-rate metrics in CiK and CAF-7M (DoubleCast) measure how often a model outperforms a forecasting baseline. 
WIT's Context Win Rate instead measures, for the same model and case, whether adding the factual scenario reduces forecast error relative to its neutral forecast.
\end{minipage}
\endgroup
\vspace{-10pt}
\end{table*}

%% file: scripts/3_WIT.tex
\section{What If TSF Benchmark Construction}
\label{sec:wit_construction}

WIT follows a common construction protocol.
Given a target time series and a corpus of time-stamped textual sources, the protocol constructs the domain background context, forecasting anchors, controlled scenario sets, and empirical matched sets using numerical and semantic checks appropriate to each component.
Although time series and textual context are domain-specific, the same construction principles apply across domains.
This allows WIT to be extended to new targets and observation periods without redesigning the benchmark case by case.

Construction proceeds in five stages.
First, we specify the forecasting target and relevant external mechanisms and generate candidate forecasting states.
Second, frozen history-only forecasters identify candidates whose observed futures show persistent deviations from calibrated numerical expectations.
Third, source evidence is verified and condensed into factual scenarios, and the final anchors are selected from evidence-qualified candidates.
Fourth, alternative and irrelevant scenarios are constructed separately using the fixed domain background and combined with each anchor's factual scenario to form controlled scenario sets.
Finally, empirical matched sets are constructed separately for supported anchor--scenario pairs.

\subsection{Domain Background Context and Candidate States}

\begin{wrapfigure}{r}{0.50\textwidth}
  \vspace{-1.5em}
  \centering
  \includegraphics[width=\linewidth]{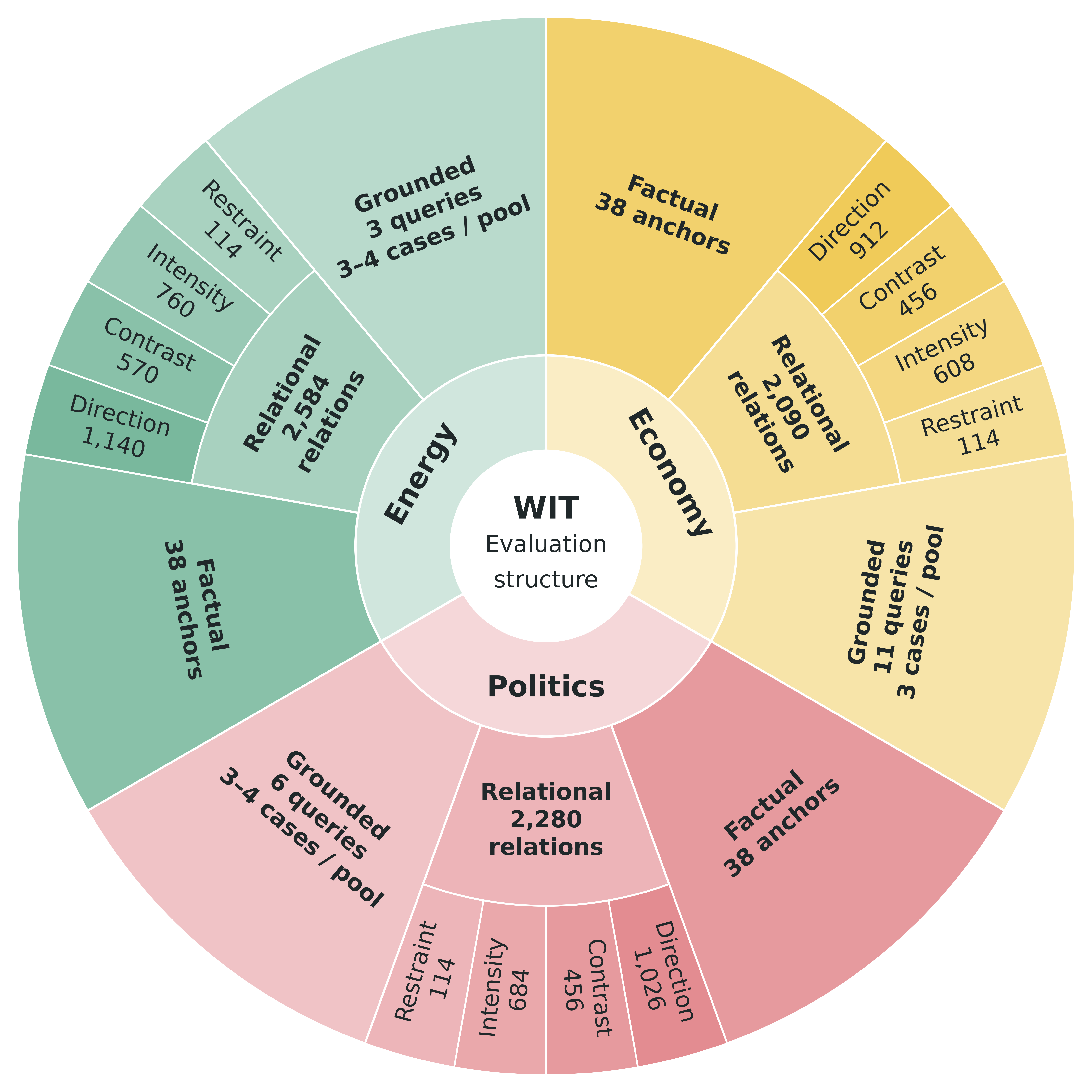}
  \vspace{-20pt}
  \caption{Composition of the current WIT release by domain and evaluation category.}
  \label{fig2:pie}
  \vspace{-2.0em}
\end{wrapfigure}

For each domain $D$, we define a domain background context $\mathcal{C}_D$ containing the target definition and a compact set of mechanisms through which external conditions can plausibly affect it.
In the current release (Figure~\ref{fig2:pie}), the targets are the Henry Hub natural gas spot price (Energy), the Nominal Broad U.S.\ Dollar Index (Economy), and reported polling support for a selected party (Politics).
Details of data sources and benchmark statistics are reported in Appendix~\ref{app:sources_and_stats}.

$\mathcal{C}_D$ serves both as the shared background supplied to models and as the semantic reference used during construction.
It contains only general domain knowledge and no episode-specific future information.
Each candidate state is $z_i=(\mathbf{x}_i,\mathcal{C}_D)$, where $\mathbf{x}_i$ is the numerical history.
Candidate windows are generated from available observations while preserving the original observation calendar.
For each selected anchor, the same forecasting state $z_i$ is reused across all scenario conditions so that only the future scenario changes.
Detailed window construction and domain-background construction are provided in
Appendices~\ref{app:data_windows} and~\ref{app:background_context}, respectively.
Examples of $\mathcal C_D$ for each domain are shown in Appendix~\ref{app:input_domain_background}.

\subsection{Reference-Guided Candidate Screening}

Not every forecasting window is equally informative for evaluating scenario responsiveness.
WIT therefore uses frozen history-only forecasts from Chronos-2, TimesFM-3, and TiRex-2 as construction references.
We identify candidates for which all three reference forecasts persistently deviate from the observed future in the same direction, as assessed using calibrated reference-relative residuals.

We estimate each reference's typical forecast errors using a separate construction-only pool of calibration windows, each containing a numerical history and its observed future.
These calibration residuals are used to robustly center and scale candidate residuals at each forecast step, making deviations comparable across domains.
Because stride-one windows can share much of the same history and future, calibration for each candidate excludes windows whose history--future intervals overlap that of the candidate.
Cross-model agreement reduces dependence on any single forecaster, while the persistence criterion filters out cases whose deviation is concentrated in only a small part of the forecast horizon.

Candidates passing this numerical screening become eligible for final anchor selection only after passing the factual evidence review.
The construction-only calibration, cross-fitted reference-relative deviations, and candidate eligibility criteria are detailed in Appendix~\ref{app:reference_screening}.



\subsection{Factual Scenario Construction and Anchor Selection}

\paragraph{Source-grounded evidence extraction.}
For each numerically eligible candidate, time-stamped textual sources covering the forecast horizon are examined for source-supported conditions relevant to mechanisms in $\mathcal{C}_D$.
We separate source-grounded evidence extraction from scenario composition.
Events and conditions are first extracted without exposing the observed target trajectory, reference-relative deviations, or any preferred response direction (Appendix~\ref{app:event_extraction}).

\paragraph{Factual scenario composition and review.}
Factual scenarios are then composed from source-supported conditions according to source support, target relevance, temporal relevance, and mechanism applicability.
Scenario composition does not use the observed reference-relative deviations.
Any qualitative alignment between the fixed scenario and those deviations is assessed only afterward and does not affect scenario wording or sample inclusion (Appendix~\ref{app:scenario_composition}).
This prevents factual scenarios from being selected simply because they align with the direction of the observed future.

The verified evidence is condensed into a \emph{compact, mechanism-preserving scenario} containing only the information needed to characterize the relevant future condition.
Observed target trajectories, reference forecasts, quantified target effects, and explicit forecasting instructions are excluded, while unnecessary identifying details are removed without discarding information required to interpret the scenario.
LLM assistance is used for evidence extraction and factual scenario composition, followed by a separate source and semantic review.
Detailed review and sample-acceptance criteria are provided in Appendix~\ref{app:scenario_review}.

\paragraph{Final anchor selection.}
Candidates without sufficiently supported factual scenarios are excluded from anchor selection.
The final anchor set is then selected from evidence-qualified candidates to cover the observation period broadly while avoiding near-duplicate cases within the same forecasting entity (Appendix~\ref{app:final_anchor_selection}).

\subsection{Controlled Alternative and Irrelevant Scenarios}
\label{sec:alternative_irrelevant}

Controlled alternative scenarios are constructed separately from factual scenarios by translating mechanisms in the fixed $\mathcal C_D$ into explicit, target-relevant hypothetical conditions.
Irrelevant scenarios are constructed separately to test responses to information unrelated to the target.
Both types of scenarios are constructed once per domain and reused across its finalized anchors.

\paragraph{Mechanism-grounded alternative scenarios.}
Each alternative scenario explicitly includes the information required to determine its qualitative implication, so that the expected relation can be interpreted without relying on additional anchor-specific context.
An expected upward or downward revision relative to the neutral forecast defines \emph{Direction}; paired scenarios with opposing implications define \emph{Contrast}; and stronger and weaker versions of the same condition define \emph{Intensity}.
Only relations that can be directly determined from the stated scenarios and $\mathcal C_D$ are retained.

\paragraph{Irrelevant scenarios.}
Irrelevant scenarios are checked for target irrelevance at both the topic and content levels during construction.
They support \emph{Restraint} tests, which assess whether the forecast remains largely unchanged relative to the neutral forecast.

\paragraph{Relation validation and cross-anchor applicability.}
For each domain, two domain experts who are not involved in scenario construction independently assess whether the scenario texts and $\mathcal C_D$ justify the expected Direction labels and Intensity orderings.
Both experts then check every retained Direction and Intensity relation across all finalized anchors in that domain for applicability without additional unstated assumptions.
Contrast pairs are formed from validated opposing Direction relations (Appendix~\ref{app:controlled_scenarios}).
This enables controlled comparisons without constructing synthetic future trajectories for alternative scenarios.


\subsection{Empirical Matched Sets}
\label{sec:empirical_matched_set}

Grounded Comparison uses separately constructed matched sets of real cases.
For each supported anchor--scenario pair, matched cases must have a similar numerical history, similar frozen construction-reference forecasts, and source evidence that the same scenario mechanism occurred with compatible timing, scope, roles, and applicability.
Matching does not use observed future target trajectories.
The matched cases provide empirical references for neutral-relative revision trajectories, rather than counterfactual outcomes for the anchor or estimates of causal effects.

Grounded Comparison provides complementary empirical evidence when supported pairs are available.
Coverage could be constrained by the difficulty of finding cases with both similar forecasting states and compatible scenario occurrences within the limited observation period permitted by strict knowledge-cutoff requirements.
Further matching details are provided in Appendix~\ref{app:matched_sets}.

%% file: scripts/4_evaluation.tex
\section{Benchmark Evaluation Details}
\label{sec:eval_details}

\begin{figure*}[t!]
  \centering
  {\includegraphics[width=\linewidth]{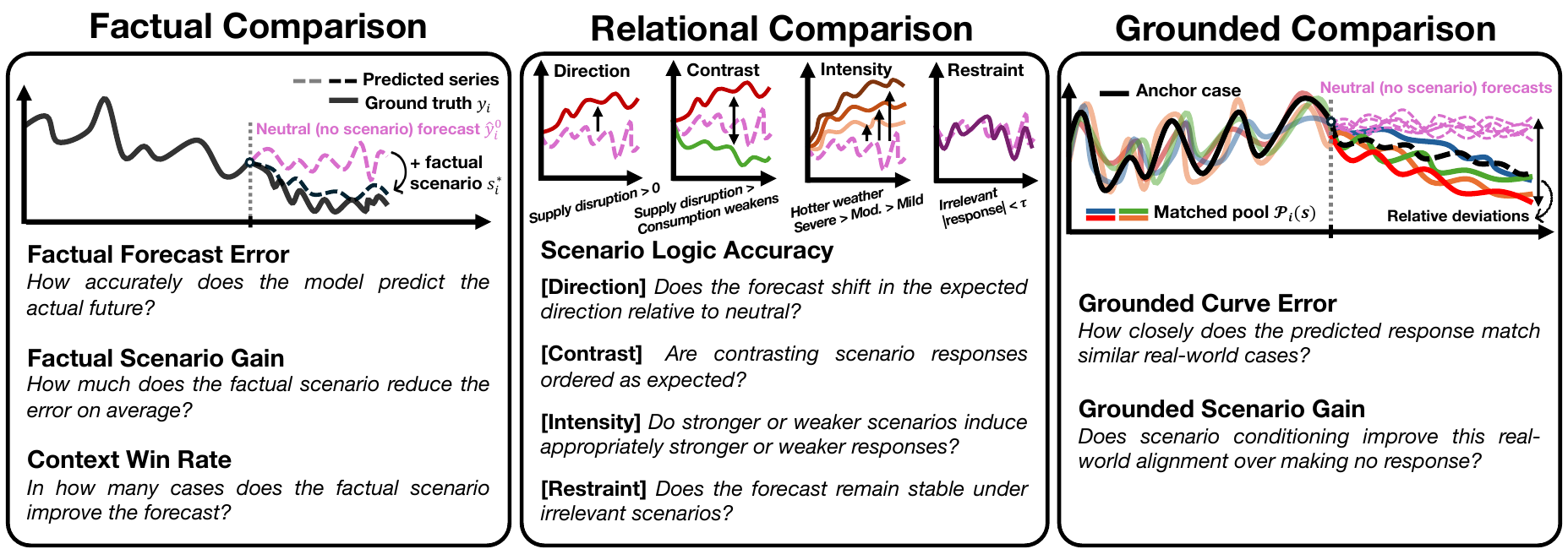}}
  \vspace{-10pt}
    \caption{{Three complementary comparisons in the WIT benchmark.}}
  \label{fig3:witmetrics}
  \vspace{-7pt}
\end{figure*}

WIT evaluates scenario-guided forecasting through three complementary comparisons: \emph{Factual}, \emph{Relational}, and \emph{Grounded} (see Figure~\ref{fig3:witmetrics}).
This separation reflects a fundamental asymmetry between factual and alternative scenarios.
The factual scenario represents source-supported conditions that actually occurred or remained relevant during the forecast horizon and therefore has an observed future against which forecast accuracy can be measured.
An alternative scenario did not occur for the same anchor, so constructing a synthetic future for it would require additional assumptions about the unobserved outcome.
WIT instead evaluates alternative scenarios through neutral-relative relations that can be specified without constructing synthetic futures and, where empirical support is available, through comparable real cases in which the same mechanism occurred.
Further evaluation details and a concrete Economy-domain example are provided in Appendix~\ref{app:eval_details} and Figure~\ref{fig:wit_visualization}.


\subsection{Neutral-Relative Scenario Revisions}
\label{sec:neutral}

Let $\widehat{\mathbf y}_i^{\,s}=f_\theta(\mathbf x_i,\mathcal C_D,s)$ denote the forecast conditioned on scenario $s$, and let $\widehat{\mathbf y}_i^{\,0}=f_\theta(\mathbf x_i,\mathcal C_D,\varnothing)$ denote the corresponding \emph{neutral forecast}.
Neutral means scenario-unspecified; it does not represent an event-free counterfactual.
We define the scenario-induced revision relative to the model's own neutral forecast as
\vspace{10pt}
\begin{equation}
\Delta_{i,h}(s)
=
\widehat y_{i,h}^{\,s}
-
\widehat y_{i,h}^{\,0},
\qquad h=1,\ldots,H.
\end{equation}

This neutral-relative formulation focuses evaluation on scenario-induced forecast revisions while reducing sensitivity to differences in baseline forecast levels across models.

\subsection{Factual Comparison}

Factual Comparison asks how accurately the model predicts the observed future $\mathbf y_i$ and whether providing the factual scenario improves that prediction relative to the neutral forecast.
Let $s_i^{*}$ denote the factual scenario for anchor $i$, and define
\vspace{10pt}
\begin{equation}
e_i^q
=
\frac{1}{H}
\sum_{h=1}^{H}
\left(
\widehat y_{i,h}^{\,q}-y_{i,h}
\right)^2,
\qquad
\overline e^{\,q}
=
\frac{1}{N}\sum_{i=1}^{N}e_i^q,
\end{equation}

where $q=0$ denotes the neutral condition and $q=s_i^{*}$ the factual condition.

We characterize factual performance from three complementary perspectives:
\vspace{5pt}
\begin{equation}
\begin{aligned}
\mathrm{FFE}=\overline e^{\,s^{*}},
\quad
\mathrm{FSG}=100\left(
1-\frac{\overline e^{\,s^{*}}}{\overline e^{\,0}}
\right),
\quad
\mathrm{CWR}=
\frac{100}{N}
\sum_{i=1}^{N}
\mathbf 1
\left[
e_i^{s_i^{*}}<e_i^0
\right].
\end{aligned}
\end{equation}

\textbf{Factual Forecast Error (FFE)} measures forecast error when the factual scenario is provided.
\textbf{Factual Scenario Gain (FSG)} measures the aggregate reduction in error relative to the same model's neutral forecast; positive FSG indicates lower mean forecast error under the factual scenario.
FSG is computed from aggregate errors rather than by averaging anchor-specific percentage changes.
\textbf{Context Win Rate (CWR)} measures how consistently the factual scenario reduces forecast error across anchors, with ties not counted as improvements.
FFE, FSG, and CWR therefore capture factual forecast accuracy, aggregate improvement, and the share of anchors that improve, respectively.


\subsection{Relational Comparison}

Relational Comparison evaluates whether neutral-relative forecast revisions satisfy predefined relations among controlled scenarios under $\mathcal C_D$.
We consider four complementary properties: \emph{Direction}, \emph{Contrast}, \emph{Intensity}, and \emph{Restraint}.
\begin{equation}
\begin{aligned}
R_i(s)=\frac{1}{H}\sum_{h=1}^H\Delta_{i,h}(s), \qquad
A_i(s)=\frac{1}{H}\sum_{h=1}^H|\Delta_{i,h}(s)|.
\end{aligned}
\end{equation}
Here, $R_i(s)$ summarizes the signed average revision induced by scenario $s$, while $A_i(s)$ measures its average absolute magnitude.

Let $d_s\in\{-1,+1\}$ denote the expected direction of scenario $s$, let $(s_+,s_-)$ denote a pair of opposing scenarios, and let $(s_k,s_{k+1})$ denote adjacent intensity levels with common direction $d$.
Using domain-specific practical tolerances $\eta_D$ and $\tau_D$, we evaluate
\begin{equation}
\begin{aligned}
\mathrm{Direction}: d_sR_i(s)>\eta_D, \quad
\mathrm{Contrast}: R_i(s_+)-R_i(s_-)>\eta_D, \\[5pt] 
\mathrm{Intensity}: d\left[R_i(s_{k+1})-R_i(s_k)\right]>\eta_D, \quad
\mathrm{Restraint}: A_i(s_{\mathrm{irr}})\le\tau_D.
\end{aligned}
\end{equation}
The current-release values, reusable procedure for setting these tolerances from domain-specific target scales, and sensitivity analyses are detailed in Appendix~\ref{app:relation_details}.

Each relation probes a different aspect of scenario response and is evaluated using \textbf{Scenario Logic Accuracy (SLA)}.
\textbf{Direction} checks for a non-negligible revision in the expected direction.
\textbf{Contrast} asks whether opposing scenarios produce distinguishable revisions.
\textbf{Intensity} captures whether stronger conditions lead to correspondingly stronger revisions.
\textbf{Restraint} measures whether irrelevant information leaves the forecast essentially unchanged.
Together, the four relations assess directional responsiveness, contrast between opposing scenarios, sensitivity to scenario strength, and restraint under irrelevant information.

Let $\mathcal R_{i,c}$ denote the relations for anchor $i$ in category $c$, and let $\mathcal A_c$ denote the anchors eligible for that category.
Relations are first averaged within each anchor, anchors are weighted equally within each category, and the four categories are weighted equally in the Overall SLA.
\begin{equation}
\begin{aligned}
\mathrm{SLA}_c=
\frac{100}{|\mathcal A_c|}
\sum_{i\in\mathcal A_c}
\frac{1}{|\mathcal R_{i,c}|}
\sum_{r\in\mathcal R_{i,c}}\mathbf 1[r],
\qquad
\mathrm{SLA}=
\frac{1}{4}
\sum_{c\in\{\mathrm{Dir},\mathrm{Con},\mathrm{Int},\mathrm{Res}\}}
\mathrm{SLA}_c.
\end{aligned}
\end{equation}

We report Direction, Contrast, Intensity, Restraint, and Overall separately so that different failure modes remain visible.
SLA therefore measures consistency with predefined scenario relations without requiring a target trajectory for an unobserved alternative scenario.

\subsection{Grounded Comparison}

Relational Comparison establishes whether forecast revisions have the expected qualitative structure, but satisfying those relations does not ensure agreement with responses observed in comparable real cases.
For example, a model may correctly order two scenarios while producing revisions whose magnitudes differ substantially from empirical references.
Grounded Comparison therefore evaluates scenario-induced revisions against matched real cases.

For each supported anchor--scenario pair $(i,s)$, let $\mathcal P_i(s)$ denote its empirical matched set, constructed as described in Section~\ref{sec:empirical_matched_set}.
For each matched case $j$, define
\begin{equation}
d_{j,h}
=
y_{j,h}
-
\widehat y_{j,h}^{\,0},
\end{equation}
where $\widehat y_{j,h}^{\,0}$ is the evaluated model's own neutral forecast for matched case $j$.
Thus, $d_{j,h}$ represents the observed neutral-relative deviation in that matched case.

The pair-level mean squared curve discrepancy is
\begin{equation}
g_i(s)
=
\frac{1}{H|\mathcal P_i(s)|}
\sum_{j\in\mathcal P_i(s)}
\sum_{h=1}^{H}
\left(
\Delta_{i,h}(s)-d_{j,h}
\right)^2.
\end{equation}
The corresponding no-revision baseline sets $\Delta_{i,h}(s)=0$:
\begin{equation}
g_i^{0}(s)
=
\frac{1}{H|\mathcal P_i(s)|}
\sum_{j\in\mathcal P_i(s)}
\sum_{h=1}^{H}
(d_{j,h})^2.
\end{equation}

Let $\mathcal Q_D$ denote the supported anchor--scenario pairs in domain $D$.
We first average pair-level discrepancies within each domain:
\begin{equation}
\begin{aligned}
\overline g_D
&=
\frac{1}{|\mathcal Q_D|}
\sum_{(i,s)\in\mathcal Q_D}
g_i(s),
\qquad
\overline g_D^{\,0}
=
\frac{1}{|\mathcal Q_D|}
\sum_{(i,s)\in\mathcal Q_D}
g_i^{0}(s).
\end{aligned}
\end{equation}

We report domain-wise \textbf{Grounded Curve Error (GCE)} and \textbf{Grounded Scenario Gain (GSG)}
\begin{equation}
\mathrm{GCE}_D
=
\sqrt{\overline g_D},
\qquad
\mathrm{GSG}_D
=
100\left(
1-\frac{\overline g_D}{\overline g_D^{\,0}}
\right).
\end{equation}
Overall GSG is reported as the equal-domain macro-average of $\mathrm{GSG}_D$.

GCE measures the discrepancy between the predicted scenario-induced revision and neutral-relative deviations observed in matched real cases across the forecast horizon.
Because the alternative future is unobserved, matched cases serve as empirical references rather than ground truth: lower GCE indicates closer empirical alignment, but zero does not represent a uniquely correct alternative future.
GSG measures the improvement in this alignment relative to making no revision; positive GSG indicates better agreement than the no-revision baseline.

%% file: scripts/5_experiments.tex
\section{Experiments and Findings}
\label{sec:experiments}

We evaluate WIT across three domains with distinct targets and relevant external mechanisms: Energy, Economy, and Politics.
The current release contains 114 forecasting anchors, with 38 anchors per domain, constructed following the common protocol in Section~\ref{sec:wit_construction}.
Each anchor consists of a 24-step numerical history and a 20-step forecast horizon, and is paired with the fixed domain background context, its factual scenario, and the domain’s alternative and irrelevant scenarios.
Supported anchor--scenario pairs are additionally linked to empirical matched sets for Grounded Comparison.

Following the evaluation framework in Section~\ref{sec:eval_details}, our experiments examine whether gains under factual scenarios are accompanied by responses that satisfy predefined scenario relations and align with comparable real cases.
We evaluate this through Factual, Relational, and Grounded Comparisons, testing whether factual forecasting improvement alone is sufficient to characterize scenario-guided forecasting behavior.

\vspace{-5pt}
\paragraph{Models.}
We primarily compare two classes of scenario-conditioned forecasters.
\emph{Directly prompted LLMs}, including Mistral-7B-Instruct v0.3~\citep{jiang2023mistral7b}, Llama-3-70B-Instruct~\citep{grattafiori2024llama}, and GPT-4o~\citep{openai2024gpt4technicalreport}, receive the numerical history, domain background context, and scenario through a standardized prompt and directly generate the forecast trajectory.
This group tests whether general-purpose language models can produce scenario-dependent forecasts from numerical and textual inputs without a dedicated forecasting architecture.

We additionally evaluate two \emph{context-aware forecasting systems}: DoubleCast~\citep{zheng2026overcoming}, using the cutoff-controlled Qwen1.5-14B-Chat text encoder and CAF-7M cross-modal alignment, and MoiraiAgent~\citep{yan2026moiraiagent}, using GPT-4o with Chronos-2 as its time-series forecasting tool.
These systems explicitly couple textual context with time-series forecasting, providing specialized counterparts to direct LLM prompting.
Exact model components, knowledge cutoffs, alignment details, and inference settings are provided in Appendix~\ref{app:exp_details}.

\vspace{-5pt}
\paragraph{Evaluation protocol.}
All scenario-conditioned models are evaluated on the same fixed anchors, scenarios, and empirical matched sets.
Scale-dependent metrics are reported separately for Energy, Economy, and Politics, while dimensionless metrics are computed within each domain and macro-averaged with equal domain weight.
Each configuration is evaluated over three runs, and we report the mean and sample standard deviation.
Additional stability analyses examine decoding variability, sensitivity to neutral forecast replacement, and reproducibility of evaluation outcomes across runs (Appendix~\ref{app:stochastic_decoding_variability}).
Further evaluation details are provided in Appendices~\ref{app:eval_details} and~\ref{app:repeated_runs}.

\input{tables/table-main1}




\paragraph{Cross-comparison findings.}
The results reveal distinct patterns across the three comparisons. Full domain-wise results and additional analyses are provided in Appendix~\ref{app:analysis}.
In Factual Comparison (Table~\ref{tab:wit_factual_results}), factual scenarios improve average forecast accuracy for all models, but not consistently across anchors.
CWR shows that aggregate gains can coexist with many cases in which the factual scenario does not improve the forecast, distinguishing the magnitude of average improvement from the share of anchors that actually improve.

Relational Comparison (Table~\ref{tab:wit_relational_results}) indicates that appropriate responses to relevant scenarios do not necessarily imply restraint under irrelevant information or sensitivity to scenario strength.
Similar Overall SLA can therefore hide important differences across response types.

Grounded Comparison (Table~\ref{tab:wit_grounded_results}) shows a different pattern from Relational Comparison.
The context-aware systems consistently achieve lower GCE across all three domains and higher GSG than the directly prompted LLMs, despite not achieving the highest Overall SLA.
Thus, stronger relational consistency does not necessarily imply closer alignment with neutral-relative deviations observed in matched real cases.

\begin{figure*}[t]
\centering

\begin{minipage}[t]{0.61\textwidth}
\captionsetup{type=table}
\input{tables/table-main3}
\end{minipage}
\hfill
\begin{minipage}[t]{0.37\textwidth}
\vspace{0pt}
\centering

\includegraphics[
    width=\linewidth
]{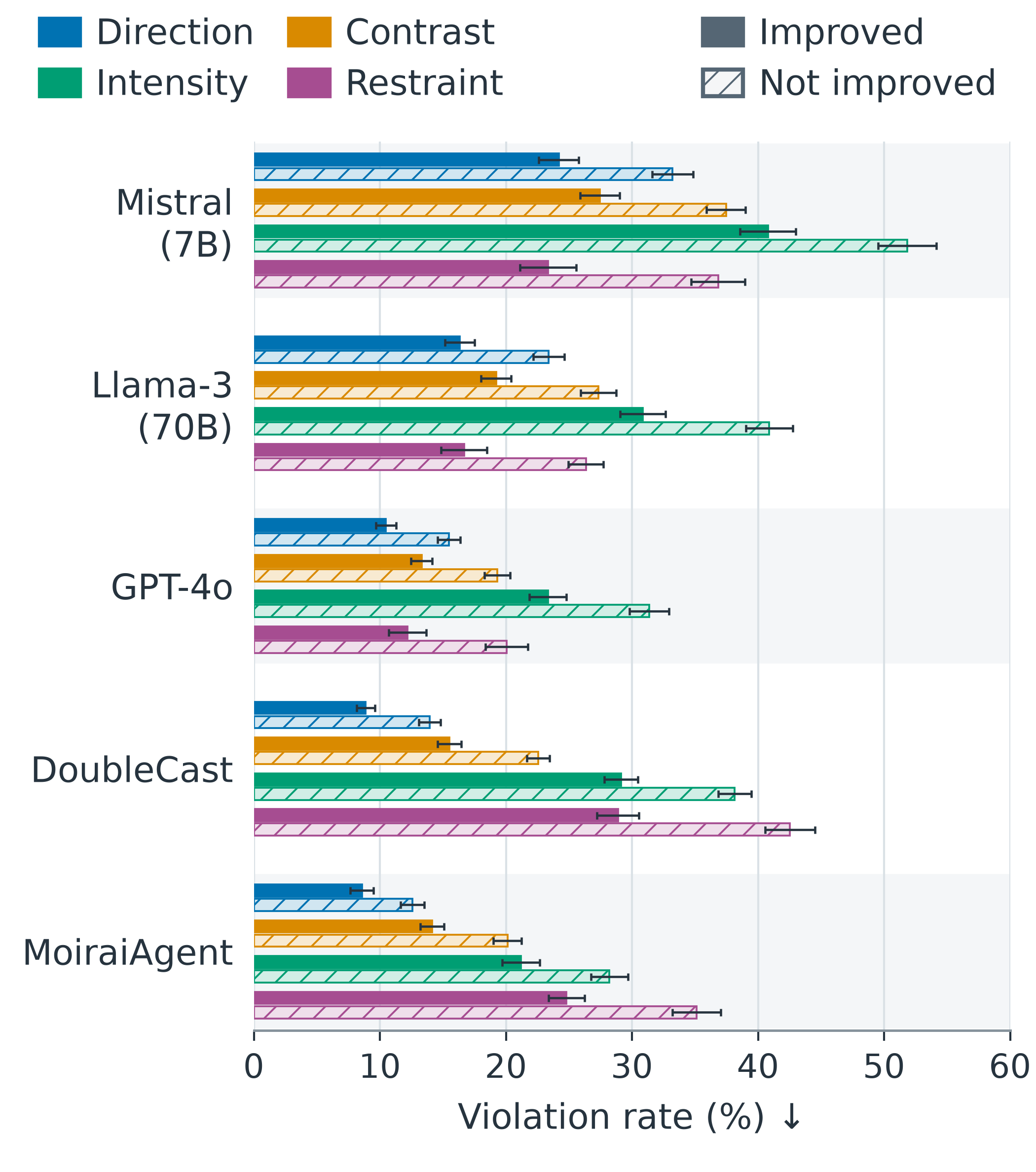}

\vspace{-5pt}
\captionof{figure}{
Relational violation rates for anchors with and without
factual forecasting improvement across three domains.
}
\label{fig:anchor_analysis}

\end{minipage}

\vspace{-5pt}
\end{figure*}

\paragraph{Anchor-level analysis.}
We further examine whether factual improvement is associated with more consistent responses across scenarios (Figure~\ref{fig:anchor_analysis}; Appendix~\ref{app:analysis}).
Anchors whose factual scenario reduces forecast error generally show lower relational violation rates than those without improvement, although violations remain in both groups.
Thus, factual improvement is associated with stronger relational consistency but still does not guarantee appropriate responses across all scenarios.

\input{tables/table-main2}

%% file: tables/table-main1.tex
\begin{table*}[t]
\centering
\small
\vspace{-10pt}
\caption{
Factual Comparison on WIT.
Factual Forecast Error (neutral forecast error) is reported separately for each domain.
FSG and CWR are macro-averaged across the three domains.
}
\vspace{-5pt}
\setlength{\tabcolsep}{5.5pt}
\renewcommand{\arraystretch}{1.15}
\begin{tabular*}{\linewidth}{
@{\extracolsep{\fill}}
l*{5}{c}
@{}
}
\toprule
\multirow{2}{*}{\textbf{Model}}
& \multicolumn{3}{c}{\textbf{Factual Forecast Error} (neutral forecast error)}
& \multirow{2}{*}{\shortstack{\textbf{Factual}\\\textbf{Scenario}\\\textbf{Gain (\%)}}}
& \multirow{2}{*}{\shortstack{\textbf{Context}\\\textbf{Win}\\\textbf{Rate (\%)}}} \\
\cmidrule(lr){2-4}
& \textbf{Energy}
& \textbf{Economy}
& \textbf{Politics}
& & \\
\midrule
\midrule

\multicolumn{6}{l}{\textit{Directly prompted LLMs}} \\
\addlinespace[2pt]

Mistral (7B)
& \factualerr{0.312}{0.013}{0.337}{0.012}
& \factualerr{2.596}{0.110}{2.805}{0.099}
& \factualerr{25.704}{0.924}{27.552}{0.840}
& $7.19 {\scriptstyle \pm 0.63}$
& $55.56 {\scriptstyle \pm 1.83}$ \\

Llama-3 (70B)
& \factualerr{0.276}{0.009}{0.313}{0.008}
& \factualerr{2.184}{0.072}{2.470}{0.066}
& \factualerr{22.792}{0.686}{25.424}{0.616}
& $11.25 {\scriptstyle \pm 0.69}$
& $60.82 {\scriptstyle \pm 2.21}$ \\

GPT-4o
& \factualerr{0.254}{0.007}{0.298}{0.006}
& \factualerr{2.233}{0.061}{2.580}{0.061}
& \factualerr{19.614}{0.518}{23.500}{0.500}
& $\mathbf{14.92} {\scriptstyle \mathbf{\pm 0.64}}$
& $65.50 {\scriptstyle \pm 1.83}$ \\

\midrule
\multicolumn{6}{l}{\textit{Context-aware forecasting systems}} \\
\addlinespace[2pt]

DoubleCast
& \factualerr{0.245}{0.003}{0.274}{0.003}
& \bestfactualerr{2.045}{0.021}{2.300}{0.021}
& \factualerr{19.200}{0.210}{23.550}{0.240}
& $13.38 {\scriptstyle \pm 0.42}$
& $66.08 {\scriptstyle \pm 1.34}$ \\

MoiraiAgent
& \bestfactualerr{0.242}{0.007}{0.272}{0.006}
& \factualerr{2.070}{0.058}{2.290}{0.055}
& \bestfactualerr{18.950}{0.500}{23.450}{0.490}
& $13.28 {\scriptstyle \pm 0.79}$
& $\mathbf{67.25} {\scriptstyle \mathbf{\pm 2.21}}$ \\

\bottomrule
\end{tabular*}
\label{tab:wit_factual_results}
\vspace{-7pt}
\end{table*}

%% file: tables/table-main2.tex
\begin{table*}[t]
\centering
\small
\vspace{-10pt}
\caption{
Relational Comparison on WIT, macro-averaged across three domains.
}
\vspace{-5pt}
\setlength{\tabcolsep}{5pt}
\renewcommand{\arraystretch}{1.12}
\begin{tabular*}{\linewidth}{
@{\extracolsep{\fill}}
l*{5}{c}
@{}
}
\toprule
\multirow{2}{*}{\textbf{Model}}
& \multicolumn{5}{c}{\textbf{Scenario Logic Accuracy (\%)}} \\
\cmidrule(lr){2-6}
& Direction
& Contrast
& Intensity
& Restraint
& \textbf{Overall} \\
\midrule
\midrule

\multicolumn{6}{l}{\textit{Directly prompted LLMs}} \\
\addlinespace[2pt]

Mistral (7B)
& $71.80 {\scriptstyle \pm 1.55}$
& $68.10 {\scriptstyle \pm 1.45}$
& $54.30 {\scriptstyle \pm 2.09}$
& $70.66 {\scriptstyle \pm 2.19}$
& $66.21 {\scriptstyle \pm 1.63}$ \\

Llama-3 (70B)
& $80.90 {\scriptstyle \pm 1.05}$
& $77.60 {\scriptstyle \pm 1.10}$
& $65.21 {\scriptstyle \pm 1.65}$
& $79.53 {\scriptstyle \pm 1.75}$
& $75.81 {\scriptstyle \pm 1.23}$ \\

GPT-4o
& $87.79 {\scriptstyle \pm 0.75}$
& $\mathbf{84.61} {\scriptstyle \mathbf{\pm 0.85}}$
& $73.90 {\scriptstyle \pm 1.36}$
& $\mathbf{85.09} {\scriptstyle \mathbf{\pm 1.46}}$
& $\mathbf{82.85} {\scriptstyle \mathbf{\pm 0.98}}$ \\

\midrule
\multicolumn{6}{l}{\textit{Context-aware forecasting systems}} \\
\addlinespace[2pt]

DoubleCast
& $89.40 {\scriptstyle \pm 0.70}$
& $82.10 {\scriptstyle \pm 0.81}$
& $67.80 {\scriptstyle \pm 1.21}$
& $66.47 {\scriptstyle \pm 1.61}$
& $76.44 {\scriptstyle \pm 1.08}$ \\

MoiraiAgent
& $\mathbf{90.11} {\scriptstyle \mathbf{\pm 0.90}}$
& $83.90 {\scriptstyle \pm 0.99}$
& $\mathbf{76.51} {\scriptstyle \mathbf{\pm 1.36}}$
& $71.83 {\scriptstyle \pm 1.61}$
& $80.59 {\scriptstyle \pm 1.11}$ \\

\bottomrule
\end{tabular*}
\label{tab:wit_relational_results}
\vspace{-7pt}
\end{table*}

%% file: tables/table-main3.tex
\centering
\small
\caption{
Grounded Comparison on WIT.
Grounded Curve Error (GCE) is reported separately for each domain, while Grounded Scenario Gain (GSG) is macro-averaged with equal weight across the three domains.
}
\vspace{-2pt}
\setlength{\tabcolsep}{4pt}
\renewcommand{\arraystretch}{1.15}

\begin{tabular*}{\linewidth}{
@{\extracolsep{\fill}}
l*{4}{c}
@{}
}
\toprule
\multirow{2}{*}{\textbf{Model}}
& \multicolumn{3}{c}{\textbf{Grounded Curve Error}}
& \multirow{2}{*}{\shortstack{\textbf{Grounded}\\\textbf{Scenario}\\\textbf{Gain (\%)}}} \\
\cmidrule(lr){2-4}
& \textbf{Energy}
& \textbf{Economy}
& \textbf{Politics}
& \\
\midrule
\midrule

\multicolumn{5}{l}{\textit{Directly prompted LLMs}} \\
\addlinespace[2pt]

Mistral (7B)
& \resultcell{0.592}{0.009}
& \resultcell{1.740}{0.028}
& \resultcell{6.885}{0.150}
& \resultcell{-4.56}{1.17} \\

\addlinespace[2pt]
Llama-3 (70B)
& \resultcell{0.515}{0.005}
& \resultcell{1.527}{0.019}
& \resultcell{5.949}{0.112}
& \resultcell{8.20}{0.50} \\

\addlinespace[2pt]
GPT-4o
& \resultcell{0.491}{0.009}
& \resultcell{1.449}{0.023}
& \resultcell{5.538}{0.094}
& \resultcell{11.38}{0.86} \\

\midrule
\multicolumn{5}{l}{\textit{Context-aware forecasting systems}} \\
\addlinespace[2pt]

DoubleCast
& \bestresultcell{0.446}{0.004}
& \bestresultcell{1.311}{0.011}
& \resultcell{4.902}{0.050}
& \bestresultcell{17.64}{0.43} \\

\addlinespace[2pt]
MoiraiAgent
& \resultcell{0.449}{0.006}
& \resultcell{1.320}{0.018}
& \bestresultcell{4.827}{0.078}
& \resultcell{17.50}{0.64} \\

\bottomrule
\end{tabular*}
\label{tab:wit_grounded_results}
\vspace{-7pt}

%% file: scripts/6_conclusion.tex
\section{Conclusion}
\label{sec:conclusion}

We introduced What If TSF (WIT), a benchmark for evaluating scenario-guided forecasting beyond accuracy on the observed future.
WIT separates \emph{Factual}, \emph{Relational}, and \emph{Grounded} comparisons to assess predictive gains, consistency with predefined relations across controlled scenario changes, and empirical alignment with matched real cases.
Across three domains, factual gains do not ensure appropriate responses to alternative scenarios, and stronger relational consistency does not necessarily imply closer empirical alignment.
These findings highlight the need to evaluate scenario responsiveness directly, rather than relying only on accuracy for the observed future.

%% file: appendix/a_data.tex
\section{Data Sources and Benchmark Statistics}
\label{app:sources_and_stats}

\subsection{Domain Data Sources}
\label{app:sources}

\subsubsection{Energy}
WIT uses the \textit{Henry Hub natural gas spot price} as the forecasting target for the \textit{Energy} domain.
We combine the numerical price series with textual information from official energy reports and news articles describing market conditions and relevant external events.
The data collection period spans January 2, 2024 to September 9, 2026.

\paragraph{Time-series data.}
The numerical data consist of the Henry Hub Natural Gas Spot Price (NG.RNGWHHD.D), obtained from the U.S.\ Energy Information Administration (EIA) through its Open Data API~\footnote{\url{https://www.eia.gov/opendata/}}.
The original observation calendar is preserved during benchmark construction; window construction and treatment of unavailable observations are described in Appendix~\ref{app:data_windows}.

\paragraph{Textual sources.}
Textual information is collected from official EIA reports and energy news articles indexed through the Global Database of Events, Language, and Tone (GDELT).

\begin{itemize}
    \item \textbf{Daily reports:} U.S.\ EIA, \textit{Today in Energy}, tagged with Natural Gas, from January 4, 2024 to September 9, 2026.\\
    \url{https://www.eia.gov/todayinenergy/index.php?tg=natural%20gas}

    \item \textbf{Weekly reports:} U.S.\ EIA, \textit{Natural Gas Weekly Update} and \textit{Weekly Natural Gas Storage Report Supplement}, from January 11, 2024 to September 9, 2026.\\
    \url{https://www.eia.gov/naturalgas/weekly/}\\
    \url{https://www.eia.gov/naturalgas/weekly/supplement/}

    \item \textbf{News articles:} articles indexed through the GDELT 1.0 Global Knowledge Graph (GKG) from January 2, 2024 to September 9, 2026, retrieved using the keywords \textit{natural gas} and \textit{Henry Hub}.\\
    \url{http://data.gdeltproject.org/gkg/index.html}
\end{itemize}


\subsubsection{Economy}

WIT uses the \textit{Nominal Broad U.S.\ Dollar Index} as the forecasting target for the \textit{Economy} domain.
We combine the numerical index series with textual information from news articles describing exchange rates, dollar conditions, and related macroeconomic developments.
The data collection period spans January 2, 2024 to September 9, 2026.

\paragraph{Time-series data.}
The numerical data consist of the Nominal Broad U.S.\ Dollar Index (DTWEXBGS), obtained from the Federal Reserve Bank of St.\ Louis (FRED)~\footnote{\url{https://fred.stlouisfed.org/series/DTWEXBGS}}.
The original observation calendar is preserved during benchmark construction; window construction and treatment of unavailable observations are described in Appendix~\ref{app:data_windows}.

\paragraph{Textual sources.}
Textual information is collected from domestic and international news articles indexed through the Global Database of Events, Language, and Tone (GDELT).

\begin{itemize}
    \item \textbf{News articles:} articles indexed through the GDELT 1.0 Global Knowledge Graph (GKG) from January 2, 2024 to September 9, 2026, retrieved using the keywords \textit{dollar index}, \textit{USD index}, and \textit{exchange rate}.\\
    \url{http://data.gdeltproject.org/gkg/index.html}
\end{itemize}


\subsubsection{Politics}
WIT uses reported polling support for selected political parties as the forecasting target for the \textit{Politics} domain.
The numerical data cover multiple party-specific series across several countries and are observed at varying intervals.
We combine these series with country-specific news articles describing political and related external conditions relevant to party support.
The data collection period spans January 2, 2024 to August 21, 2026.

\paragraph{Time-series data.}
The numerical data consist of reported polling support for major political parties, compiled by Europe Elects\footnote{\url{https://europeelects.eu/}}.
For each country, we select two to three parties with substantial polling support during the collection period.
The selected parties are:

\begin{itemize}
    \item \textbf{United Kingdom:} Conservative Party (CON), Labour Party (LAB), Reform UK (REFORM).
    
    \item \textbf{Hungary:} FIDESZ--Magyar Polgári Szövetség--Kereszténydemokrata Néppárt (Fidesz+KDNP), Tisztelet és Szabadság Párt (TISZA).
    
    \item \textbf{Germany:} Alternative für Deutschland (AfD), Christlich Demokratische Union Deutschlands--Christlich-Soziale Union in Bayern e.V. (CDU+CSU).
    
    \item \textbf{Greece:} \foreignlanguage{greek}{Νέα Δημοκρατία} (ND), \foreignlanguage{greek}{ΠΑΣΟΚ - Κίνημα Αλλαγής} (PASOK-KINAL), \foreignlanguage{greek}{Συνασπισμός Ριζοσπαστικής Αριστεράς – Προοδευτική Συμμαχία} (SYRIZA).
\end{itemize}

The original observation calendar of each party-specific series is preserved during benchmark construction; window construction for irregularly observed series is described in Appendix~\ref{app:data_windows}.

\paragraph{Textual sources.} 
Textual information is collected from major domestic and international news outlets for each country.
Articles are retrieved according to country and target relevance, with detailed evidence filtering and factual-scenario construction described in Appendix~\ref{app:factual_construction}.
The primary news sources are:

\begin{itemize}
    \item \textbf{United Kingdom:}
    The Guardian\footnote{\url{https://www.theguardian.com/}},
    BBC\footnote{\url{https://www.bbc.com/}},
    and The Telegraph\footnote{\url{https://www.telegraph.co.uk/}}.

    \item \textbf{Hungary:}
    Magyar Nemzet\footnote{\url{https://magyarnemzet.hu/}},
    Népszava\footnote{\url{https://nepszava.hu/}},
    and 24.hu\footnote{\url{https://24.hu/}}.

    \item \textbf{Germany:}
    Der Spiegel\footnote{\url{https://www.spiegel.de/}},
    Die Zeit\footnote{\url{https://www.zeit.de/index}},
    and Frankfurter Allgemeine Zeitung\footnote{\url{https://www.faz.net/aktuell/}}.

    \item \textbf{Greece:}
    Kathimerini\footnote{\url{https://www.ekathimerini.com/}},
    To Vima\footnote{\url{https://www.tovima.com/}},
    and Ta Nea\footnote{\url{https://www.tanea.gr/}}.
\end{itemize}


\subsection{Benchmark Statistics}
\label{app:stats}

The current WIT release contains three domains---Energy, Economy, and Politics---with 38 forecasting anchors per domain and 114 anchors in total.
All domains use a 24-step numerical history and a 20-step forecast horizon.
The forecasting targets are the Henry Hub natural gas spot price in the Energy domain, the Nominal Broad U.S.\ Dollar Index in the Economy domain, and reported polling support for a selected party in the Politics domain.
The scenario inventory differs by domain according to the mechanisms represented in its domain background context.

\subsubsection{Temporal Coverage and Window Statistics}
\label{app:temporal_stats}

A step corresponds to an available numerical observation rather than a fixed calendar duration.
We preserve the original observation calendar without interpolating missing dates; consequently, the same 24-step history and 20-step forecast horizon can span different calendar durations across domains and anchors.
Table~\ref{tab:window_stats} summarizes the observation  periods, the number of forecasting entities, and the calendar spans of the history and forecast windows for each domain.
The observation period covers the full range of valid observations; the candidate period covers forecast origins where complete 24-step histories and 20-step futures can be formed before screening; and the anchor period covers the forecast origins of the 38 anchors retained in the final benchmark.

\input{tables/apdx-window-stats}

\subsubsection{Anchor and Scenario Composition}
\label{app:anchor_scenario_stats}

Table~\ref{tab:data_comp} summarizes the anchor-level composition of WIT.
Each anchor contains one factual scenario, a domain-specific set of controlled alternative scenarios, and three irrelevant scenarios.
Energy uses 30 controlled alternatives per anchor, Economy 24, and Politics 27.
The domain-level alternative and irrelevant scenarios are shared across anchors, while factual scenarios are constructed for individual anchors.
Across all anchors, this yields 3,078 controlled alternative, 342 irrelevant, and 114 factual anchor--scenario pairs.

Including the factual scenario, controlled alternatives, three irrelevant scenarios, and one neutral condition, each Energy, Economy, and Politics anchor requires 35, 29, and 32 forecasts, respectively.
This gives 3,648 anchor-level forecasts per scenario-conditioned model and run; additional neutral forecasts for empirical matched cases are excluded from this count.

\input{tables/apdx-data-composition-domain}

\subsubsection{Controlled Scenario and Relation Inventory}
\label{app:relational_inventory}

The controlled scenario sets generate substantially more relational tests than the number of anchors alone would suggest.
Table~\ref{tab:relational_inventory} reports the number of relations in each category.
Each controlled alternative contributes a \emph{Direction} relation; \emph{Contrast} compares predefined opposing scenario pairs; \emph{Intensity} compares adjacent levels within ordered scenario-strength sequences; and each irrelevant scenario contributes a \emph{Restraint} relation.

Across the selected anchors, the predefined relations yield 3,078 Direction, 1,482 Contrast, 2,052 Intensity, and 342 Restraint tests, for a total of 6,954 relational tests.
The counts below refer to anchor-level relation applications involving controlled alternatives and irrelevant scenarios.
Because the domain-level scenario sets differ in size, each Energy, Economy, and Politics anchor contributes 68, 55, and 60 relations, respectively.

\input{tables/apdx-data-relational-inventory}

\subsubsection{Empirical Matched-Set Coverage}
\label{app:grounded_coverage}

Grounded Comparison uses a subset of anchor--scenario pairs from the fixed benchmark anchors, each supported by at least three empirical matched cases.
These matched cases are drawn from the full candidate-window pool described in Appendix~\ref{app:matched_sets}.

Grounded Comparison requires sufficient empirical matches for each evaluated anchor--scenario pair, whereas Relational Comparison does not require matched real cases.
Grounded Curve Error is therefore reported separately by domain, while Grounded Scenario Gain is macro-averaged with equal domain weight so that Economy's larger number of supported pairs does not dominate the aggregate result.

\input{tables/apdx-data-grounded-support}

%% file: tables/apdx-window-stats.tex
\begin{table*}[t]
\centering
\small
\caption{
Temporal coverage and window statistics by domain.
History and horizon spans are measured in calendar days, while all windows contain 24 and 20 available observations, respectively.
}
\label{tab:window_stats}
\vspace{-4pt}
\setlength{\tabcolsep}{5pt}
\renewcommand{\arraystretch}{1.12}
\begin{tabular*}{\linewidth}{
@{\extracolsep{\fill}}
lccc
@{}
}
\toprule
\textbf{Domain}
& \shortstack{\textbf{Observation} \textbf{period}}
& \shortstack{\textbf{\# Forecasting} \textbf{entities}}
& \shortstack{\textbf{Calendar span (days)}} \\ 
\midrule
\midrule
Energy
& \daterange{Jan 2, 2024}{Sep 9, 2026}
& 1
& 31--35 / 28--33 \\

Economy
& \daterange{Jan 2, 2024}{Sep 9, 2026}
& 1
& 31--36 / 28--32 \\

Politics
& \daterange{Jan 2, 2024}{Aug 21, 2026}
& 10
& 32--42 / 30--40 \\

\bottomrule
\end{tabular*}
\vspace{-5pt}
\end{table*}

%% file: tables/apdx-data-composition-domain.tex
\begin{table*}[t]
\centering
\small
\caption{
WIT benchmark composition by domain.
Scenario counts are reported per anchor and in total; forecast-condition counts additionally include the scenario-free neutral condition.
}
\label{tab:data_comp}
\vspace{-5pt}
\setlength{\tabcolsep}{4.5pt}
\renewcommand{\arraystretch}{1.12}
\begin{tabular*}{\linewidth}{
@{\extracolsep{\fill}}
lcccccc
@{}
}
\toprule
\textbf{Domain}
& \textbf{Anchors}
& \shortstack{\textbf{Factual}\\\textbf{/ anchor}}
& \shortstack{\textbf{Alternative}\\\textbf{/ anchor}}
& \shortstack{\textbf{Irrelevant}\\\textbf{/ anchor}}
& \shortstack{\textbf{Scenario}\\\textbf{conditions}}
& \shortstack{\textbf{All forecast}\\\textbf{conditions}} \\
\midrule
\midrule
Energy
& 38
& 1 & 30 & 3
& 1,292 & 1,330 \\

Economy
& 38
& 1 & 24 & 3
& 1,064 & 1,102 \\

Politics
& 38
& 1 & 27 & 3
& 1,178 & 1,216 \\

\midrule
\textbf{Total}
& \textbf{114}
& \textbf{114 factual}
& \textbf{3,078 alternatives}
& \textbf{342 irrelevant}
& \textbf{3,534}
& \textbf{3,648} \\

\bottomrule
\end{tabular*}
\vspace{-7pt}
\end{table*}

%% file: tables/apdx-data-relational-inventory.tex
\begin{table*}[t]
\centering
\small
\caption{
Relational Comparison inventory by domain.
Counts aggregate predefined relations over all 38 anchors; values in parentheses show relations per anchor.
}
\label{tab:relational_inventory}
\vspace{-5pt}
\setlength{\tabcolsep}{7pt}
\renewcommand{\arraystretch}{1.12}
\begin{tabular*}{\linewidth}{
@{\extracolsep{\fill}}
lcccccc
@{}
}
\toprule
\textbf{Domain}
& \textbf{Direction}
& \textbf{Contrast}
& \textbf{Intensity}
& \textbf{Restraint}
& \shortstack{\textbf{Total}\\\textbf{relations}}
& \shortstack{\textbf{Relations}\\\textbf{/ anchor}} \\
\midrule
\midrule
Energy
& 1,140 (30)
& 570 (15)
& 760 (20)
& 114 (3)
& 2,584
& 68 \\

Economy
& 912 (24)
& 456 (12)
& 608 (16)
& 114 (3)
& 2,090
& 55 \\

Politics
& 1,026 (27)
& 456 (12)
& 684 (18)
& 114 (3)
& 2,280
& 60 \\

\midrule
\textbf{Total}
& \textbf{3,078}
& \textbf{1,482}
& \textbf{2,052}
& \textbf{342}
& \textbf{6,954}
& -- \\

\bottomrule
\end{tabular*}
\vspace{-7pt}
\end{table*}

%% file: tables/apdx-data-grounded-support.tex
\begin{table}[t]
\centering
\small
\caption{
Empirical matched-set support for Grounded Comparison.
Supported pairs have at least three empirical matched cases.
Total matched cases sum pool sizes, counting reuse across pairs.
}
\label{tab:grounded_support}
\vspace{-4pt}
\setlength{\tabcolsep}{6pt}
\renewcommand{\arraystretch}{1.12}
\begin{tabular}{lccc}
\toprule
\textbf{Domain}
& \shortstack{\textbf{Supported}\\\textbf{pairs}}
& \shortstack{\textbf{Total matched}\\\textbf{cases}}
& \shortstack{\textbf{Matched cases / pair}\\\textbf{(min / median / max)}} \\
\midrule
\midrule
Energy
& 3
& 10
& 3 / 3 / 4 \\

Economy
& 11
& 33
& 3 / 3 / 3 \\

Politics
& 6
& 19
& 3 / 3 / 4 \\

\midrule
\textbf{Total}
& \textbf{20}
& \textbf{62}
& -- \\
\bottomrule
\end{tabular}
\vspace{-5pt}
\end{table}

%% file: appendix/b_construction.tex
\section{Benchmark Construction Protocol}
\label{app:construction_protocol}


This section details the operational steps of the common construction framework in Section~\ref{sec:wit_construction}.
The same procedure is applied across domains, with domain-specific numerical and textual sources used to construct each benchmark component.
The protocol first defines the domain background and candidate forecasting states, then screens candidates using frozen history-only references.
Source-grounded factual scenarios and final anchors are established next.
Alternative and irrelevant scenarios are constructed and validated separately, followed by empirical matched-set construction.
The final subsection describes controls shared across these stages.

\subsection{Domain Background Context and Candidate States}

\subsubsection{Domain Background Context Construction}
\label{app:background_context}

For each domain $D$, the domain background context $\mathcal{C}_D$ is constructed from reference material using the same procedure, with all sources published no later than December 31, 2023.
It defines the forecasting target and a compact mechanism map describing general external drivers, their supported qualitative effects, applicability conditions, and limitations.
Information tied to specific benchmark-period events, including timestamped outlooks and anchor-specific events or conditions, is excluded.

\subsubsection{Candidate Windows and Forecasting States}
\label{app:data_windows}

Candidate windows are generated with stride one over consecutive available observations on each target's original observation calendar, without interpolation.
Each candidate $i$ contains a 24-step numerical history $\mathbf{x}_i$ and a 20-step observed future $\mathbf{y}_i$, and is retained only when both windows can be fully formed from available observations.
The corresponding forecasting state is $z_i=(\mathbf{x}_i,\mathcal{C}_D)$, where $\mathcal{C}_D$ is the fixed domain background context.
Thus, $L=24$ and $H=20$ refer to numbers of available observations rather than fixed calendar durations.

Candidate histories and futures are drawn only from observations from January 2024 onward.
These temporal restrictions serve distinct purposes: the background cutoff separates general domain knowledge from benchmark-period events, while the observation cutoff reduces overlap with the reported knowledge cutoffs of the evaluated language models.

\subsection{Reference-Guided Candidate Screening}
\label{app:reference_screening}

We use Chronos-2~\citep{ansari2025chronos2}, TimesFM-3~\citep{jain2026timesfm3}, and TiRex-2~\citep{podest2026tirex2} as frozen history-only construction references.
Screening identifies candidates whose observed futures show consistent and substantial deviations from the calibrated numerical expectations of all three references, without using target-unit-specific thresholds.

\subsubsection{Construction-Only Calibration Pool}


For each domain $D$, let $\mathcal K_D$ denote a construction-only calibration pool fixed before candidate screening and kept separate from released anchors and empirical matched cases.
Each calibration window contains a numerical history and its observed future and is used to estimate the references' typical forecasting errors.
Because stride-one windows can share much of the same history and future, calibration for each candidate excludes windows whose history--future intervals overlap that of the candidate.

For candidate $i$, let $I_i$ denote the calendar interval spanning its history and future.
Its admissible calibration pool is
\begin{equation}
\mathcal K_D^{(-i)}=\left\{c\in\mathcal K_D: I_c\cap I_i=\varnothing\right\}.
\label{eq:crossfit_calibration_pool}
\end{equation}
Thus, no calibration window whose history--future interval overlaps that of candidate $i$ contributes to its calibration statistics.

\subsubsection{Cross-Fitted Reference-Relative Deviations}
\label{app:crossfit_departures}

For reference model $m$ and forecast lead $h$, define the calibration residual
\begin{equation}
e_{c,h}^{(m)}=y_{c,h}-b_{c,h}^{(m)},
\end{equation}
where $b_{c,h}^{(m)}$ is the frozen history-only forecast for calibration window $c$.
Using only $c\in\mathcal K_D^{(-i)}$, we estimate a robust center separately for each forecast step $h$
\begin{equation}
\mu_{D,m,h}^{(-i)}=\operatorname{median}_{c\in\mathcal K_D^{(-i)}} e_{c,h}^{(m)}
\end{equation}
and the corresponding raw robust scale
\begin{equation}
\widetilde{\sigma}_{D,m,h}^{(-i)}=1.4826\,\operatorname{MAD}_{c\in\mathcal K_D^{(-i)}} e_{c,h}^{(m)},
\label{eq:crossfit_reference_scale}
\end{equation}
where MAD denotes the median absolute deviation from the median, providing a robust estimate of residual dispersion that is less sensitive to extreme forecast errors than the standard deviation; the factor \(1.4826\) rescales it to the standard-deviation scale under a Gaussian distribution.

\paragraph{Zero-scale safeguard.}

The MAD at a given forecast step can be zero when the admissible residuals show insufficient variation.
Rather than introducing an arbitrary numerical constant, we first borrow scale information from other forecast leads of the same reference model and domain.
We define
\begin{equation}
\sigma_{D,m,\mathrm{pool}}^{(-i)}=1.4826\,\operatorname{MAD}\left(\left\{e_{c,h}^{(m)}-\mu_{D,m,h}^{(-i)}:c\in\mathcal K_D^{(-i)},\;h=1,\ldots,H\right\}\right).
\end{equation}
The effective scale is then
\begin{equation}
\sigma_{D,m,h}^{(-i)}=
\begin{cases}
\widetilde{\sigma}_{D,m,h}^{(-i)},
& \widetilde{\sigma}_{D,m,h}^{(-i)}>0,\\[2mm]
\sigma_{D,m,\mathrm{pool}}^{(-i)},
& \widetilde{\sigma}_{D,m,h}^{(-i)}=0
\,\,\, \text{and} \,\,\,
\sigma_{D,m,\mathrm{pool}}^{(-i)}>0,\\[2mm]
\delta_D,
& \text{otherwise},
\end{cases}
\end{equation}
where $\delta_D$ is the documented numerical resolution of the target.
This fallback preserves the center estimated for each forecast step while preventing division by zero in zero-MAD cases.

\paragraph{Candidate standardization.}
The candidate residual is standardized as
\begin{equation}
z_{i,h}^{(m)}=\frac{y_{i,h}-b_{i,h}^{(m)}-\mu_{D,m,h}^{(-i)}}{\sigma_{D,m,h}^{(-i)}}.
\label{eq:standardized_departure}
\end{equation}
This expresses each deviation relative to the typical forecasting error of the corresponding reference model at the same forecast step, rather than in the native units of the target.

\paragraph{Cross-model consensus and persistence.}
We determine the dominant direction of the candidate deviation by
\begin{equation}
p_i=\operatorname{sign}\left(\operatorname{median}_{m,h}z_{i,h}^{(m)}\right),
\label{eq:departure_direction}
\end{equation}
and define the lead-wise cross-model consensus as
\begin{equation}
\gamma_{i,h}=\min_mp_i z_{i,h}^{(m)}.
\label{eq:consensus_departure}
\end{equation}
Because the minimum is taken across reference models, $\gamma_{i,h}>0$ only when all three standardized reference-relative deviations have sign $p_i$ at lead $h$.

We summarize the persistence and magnitude of these consensus deviations using
\begin{equation}
S_i=Q_{0.25}\left(\gamma_{i,1},\ldots,\gamma_{i,H}\right).
\label{eq:persistent_departure_score}
\end{equation}
The lower quartile prevents a small number of extreme forecast errors from determining candidate eligibility.
A positive $S_i$ requires same-direction cross-model consensus over most of the forecast horizon, while larger values indicate departures that are large relative to the references' usual forecasting errors.

\paragraph{Cross-fitted percentile calibration.}
Calibration windows are scored using the same leave-overlap-out procedure.
For each $c\in\mathcal K_D$, its score $S_c$ is computed using calibration statistics that exclude all windows whose intervals overlap $I_c$.
For candidate $i$, only scores from calibration windows that do not overlap $I_i$ are used to form its reference distribution:
\begin{equation}
\widehat F_i(s)=\frac{1}{|\mathcal K_D^{(-i)}|}\sum_{c\in\mathcal K_D^{(-i)}}\mathbf 1[S_c\le s].
\end{equation}
The calibrated departure percentile is then
\begin{equation}
\pi_i=\widehat F_i(S_i).
\label{eq:departure_percentile}
\end{equation}
Thus, $\pi_i$ measures how unusually large the candidate's persistent cross-model deviation is relative to admissible construction-only calibration windows.

\subsubsection{Candidate Eligibility}

A candidate is numerically eligible only if $S_i>0$ and $\pi_i>\kappa$.
The current release uses the common threshold $\kappa=0.90$, so an eligible candidate must have a calibrated empirical percentile above 0.90 in addition to exhibiting a positive persistent consensus departure.
The value of $\kappa$ controls screening stringency rather than statistical significance and is fixed before anchor selection and applied identically across domains.

Together, the two conditions require both persistent same-direction agreement across the construction references and an unusually large deviation relative to the domain's calibrated reference-error distribution.
Because the procedure operates first in standardized residual space and then in percentile space, the same screening rule can be applied to forecasting targets with different units and numerical scales.

\subsection{Factual Scenario Construction and Anchor Selection}
\label{app:factual_construction}

LLM assistance is used for evidence extraction and factual scenario composition, followed by the source and semantic review described below.

\subsubsection{Source-Supported Factual Evidence}
\label{app:factual_evidence}

For each numerically eligible candidate, we retrieve time-stamped source evidence relevant to the forecast interval and to mechanisms in the fixed domain background context $\mathcal C_D$.
Each evidence record retains the source, timing, affected entity or scope, and supporting passages.
The source establishes what condition occurred and under what circumstances, while $\mathcal C_D$ determines whether the condition maps to a supported forecasting mechanism.

\subsubsection{Outcome-Blind Event Extraction}
\label{app:event_extraction}

Here, outcome-blind extraction means that the observed numerical target future $\mathbf y_i$, construction-reference forecasts, reference-relative deviations, calibrated departure scores, and preferred response directions are not supplied to the extractor.
Its inputs are limited to the target definition and fixed domain background, the forecast interval, relevant entity and scope information, and dated source passages.
Only source-supported conditions are retained, together with their timing, scope, mechanism, and applicability requirements.
Unsupported, conflicting, or ambiguous claims remain unresolved rather than being inferred from the numerical trajectory.

\subsubsection{Outcome-Blind Factual Scenario Composition}
\label{app:scenario_composition}

Verified conditions are selected using only source support, target relevance, temporal relevance, and mechanism applicability.
The selected conditions are condensed into a compact factual scenario that preserves information required to interpret the mechanism, including relevant timing, scope, roles, and uncertainty, while removing unnecessary source-specific or identifying details.

The model-facing scenario contains no realized target information, reference forecasts, calibrated residuals, quantified target effects, or explicit forecasting instructions.
Any qualitative alignment between the fixed scenario and the realized reference-relative deviation is assessed only afterward and does not affect scenario wording or sample inclusion.

\subsubsection{Source and Semantic Review}
\label{app:scenario_review}

A separate review checks the fixed scenario against the original source evidence and $\mathcal C_D$.
Each substantive claim must be source-supported, consistent with the reported timing, scope, status, and uncertainty, and satisfy the applicability requirements of its assigned mechanism.
Unsupported claims are not repaired using outside knowledge or realized target information; scenarios requiring substantive correction are revised and reviewed again.

\subsubsection{Sample Acceptance}
\label{app:sample_acceptance}

A candidate proceeds to anchor formation only after passing both numerical screening and factual-evidence review.
Acceptance requires at least one supported, forecast-relevant condition from which a valid factual scenario can be composed; agreement with the realized reference-relative deviation is not required.
Candidates with unresolved occurrence, timing, scope, or mechanism applicability do not proceed to anchor formation.

\subsection{Final Anchor Selection}
\label{app:final_anchor_selection}

Only candidates that pass both numerical screening and factual-evidence review are eligible for final anchor selection.
Once eligibility is established, candidates are not further ranked by realized future magnitude, reference-relative deviation strength, $S_i$, or $\pi_i$.
Final selection instead aims to provide broad temporal coverage while limiting near-duplicate forecasting states.

\paragraph{Temporal coverage and redundancy control.}
Redundancy is controlled within each forecasting entity using the distance between forecast origins in the original sequence of available observations rather than calendar time.
We define the minimum separation as $\Delta=\left\lceil \rho_{\mathrm{sep}}L\right\rceil$ with $\rho_{\mathrm{sep}}=0.5$, where $L$ is the history length.
For the current release, $L=24$, giving $\Delta=12$ available observation steps across all domains.
This ensures that selected anchors from the same forecasting entity share at most half of their numerical history, reducing highly redundant forecasting states while retaining broad temporal coverage.
Because separation is measured in available observations, the same rule applies to regularly and irregularly observed target series without introducing domain-specific calendar-time thresholds.

For Politics, the separation rule is applied independently within each party-specific series; candidates from different parties are not treated as redundant solely because their forecast origins are temporally close.

\paragraph{Deterministic selection.}
To make selection deterministic, the evaluation period is divided into equally spaced temporal strata, and eligible candidates closest to the center of each stratum are selected subject to the separation rule, with ties resolved by the earlier forecast origin.

\paragraph{Independence from empirical matching.}
Final anchor selection is independent of empirical matched-set availability.
The Factual and Relational anchor population is therefore determined without considering whether an anchor--scenario pair is subsequently supported for Grounded Comparison.


\subsection{Controlled Alternative and Irrelevant Scenarios}
\label{app:controlled_scenarios}

Alternative and irrelevant scenarios were constructed separately from the source-grounded factual scenarios, using the fixed domain background $\mathcal C_D$ as the semantic reference.
Both types of scenarios were constructed once per domain and reused across its finalized anchors.

\subsubsection{Mechanism-Grounded Alternative Scenario Construction}
\label{app:mechanism_grounded_construction}

For each domain included in WIT, one domain expert developed the candidate alternative
scenarios and proposed Direction and Intensity relations with LLM
assistance, using the fixed domain background context $\mathcal C_D$
as the reference.
For target-relevant alternatives, construction proceeded from
supported domain mechanisms to explicit target-relevant conditions,
which were then translated into candidate scenario texts and relations.
The domain expert was responsible for specifying the relevant conditions,
identifying the attributes to vary or hold fixed, and proposing
candidate Direction labels and Intensity comparisons.

The candidate Direction and Intensity relations were subsequently
reviewed by two additional domain experts per domain, neither of
whom participated in their construction.
For each domain, the scenario-construction expert and the two independent reviewers each held at least a bachelor's degree in a field relevant to that domain.
The independent validation procedure is detailed in Appendix~\ref{app:independent_relation_validation}.

\paragraph{From mechanisms to explicit conditions.}
A mechanism in $\mathcal C_D$ identifies how an external condition can be
relevant to the forecasting target, but a broad event description is not
sufficient to determine a candidate relation.
We therefore specify the conditions needed to connect the event to the
target, including the affected entities, the relevant transmission pathway,
the direction and scope of the change, and its timing or duration where
required.
Conditions needed to interpret a scenario must be stated in the scenario
text or supplied by $\mathcal C_D$, rather than inferred from unstated
anchor-specific circumstances.
For example, pipeline scenarios concern changes in gas actually delivered
to Henry Hub, rather than equipment constraints without a specified
connection to the hub.
Similarly, capital-flow scenarios specify the assets being reallocated
and the associated currency transactions, rather than referring only
to generic capital inflows or outflows.

\paragraph{From conditions to candidate scenarios.}
We translate these specifications into related hypothetical conditions
that differ in designated attributes.
When constructing directionally opposing candidate scenarios, the target-relevant change is reversed while other comparison conditions remain comparable.
For intensity variants, the direction and mechanism are preserved,
and one strength attribute is varied while the other specified
attributes are held fixed.
Holding attributes fixed does not mean preventing the consequences
of the stated condition from unfolding.
For example, economic activity, yields, capital flows, and exchange
rates may respond to an economic scenario; these downstream responses
are not fixed merely to keep the scenario comparison controlled.
Where specified by the construction conditions, separate offsetting
shocks are excluded.
The resulting texts describe the hypothetical conditions, while
their expected revision directions and Intensity orderings are
recorded as candidate relations.

\paragraph{Candidate Direction and Intensity relations.}
\emph{Direction} associates an individual scenario with an expected
upward or downward revision relative to the same model's neutral
forecast, which contains no scenario.
The labels P and M denote these expected positive and negative
revisions, respectively; they do not indicate whether the forecast
must rise or fall relative to the last historical observation.
\emph{Intensity} links weaker and stronger conditions within the
same directional sequence.
A stronger P condition is expected to produce a larger upward
revision, whereas a stronger M condition is expected to produce
a larger downward revision.
These relations specify an ordering, not a proportional response
or an exact numerical effect size.

\paragraph{Example: Economy---capital flows.}
The \texttt{M-CF} group shows how the domain mechanism
linking international portfolio reallocations to currency demand
is translated into explicit scenario conditions and candidate Direction and Intensity relations.

\begin{itemize}
    \item \textbf{Scenario conditions.}
    The P scenarios specify reallocations from foreign-currency
    sovereign bonds to U.S. Treasuries, involving sales of foreign
    currencies and purchases of U.S. dollars.
    The M scenarios reverse these transactions.
    These conditions explicitly connect portfolio movements
    to currency demand and the U.S. dollar index.

    \item \textbf{Direction.}
    \texttt{M-CF-P1} supports an upward revision relative to
    the neutral forecast because investors sell foreign currencies
    and purchase U.S. dollars.
    Conversely, \texttt{M-CF-M1} supports a downward revision
    through the reverse transactions.

    \item \textbf{Intensity.}
    Levels 1, 2, and 3 specify small, moderate, and large
    net portfolio reallocations.
    Only the magnitude of the reallocation varies, while the
    investor population, source and destination asset classes,
    currency-basket composition, and period remain fixed.
    The predefined comparisons use adjacent levels:
    \texttt{P1}$\rightarrow$\texttt{P2} and
    \texttt{P2}$\rightarrow$\texttt{P3}, with corresponding
    comparisons for M.
    Direct level-1-to-level-3 comparisons are excluded.
\end{itemize}

\paragraph{Restricting Intensity comparisons.}
The single-attribute principle applies to each Intensity comparison,
even when a scenario family varies multiple attributes.
For example, \texttt{E-FS} varies industrial output and the
gas-supplied share of process heat, while \texttt{P-EC} varies changes in workers' wages relative to essential living costs and the extent of layoffs.
Candidate Intensity relations are proposed only between scenarios
that differ in one attribute while holding the other fixed.
This yields separate attribute-specific comparisons rather than
a single intensity ordering across all scenario variants.

\subsubsection{Irrelevant Scenario Construction}
\emph{Restraint} scenarios were constructed separately from
target-relevant alternatives.
The domain expert used LLM assistance to generate candidate irrelevant scenarios.
During construction, the expert assessed each candidate at two levels: whether its general topic was unrelated to the target and its underlying mechanisms, and whether its specific content introduced any target-relevant conditions that could justify a meaningful forecast revision.
Only candidates satisfying both criteria were retained, yielding three irrelevant scenarios per domain.
These scenarios provide no stated basis for either direction and are therefore expected to leave the forecast close to the same model's neutral forecast.


\subsubsection{Independent Relation-Level Validation}
\label{app:independent_relation_validation}

The candidate Direction and Intensity relations were reviewed by two
additional domain experts per domain, neither of whom participated in
their construction.
Validation separated an initial independent assessment from subsequent
discussion, revision, and finalization.

\paragraph{Review units and information provided.}
The unit of review was a candidate relation, rather than the linguistic
quality or plausibility of an isolated scenario sentence.
A Direction review concerned the expected neutral-relative revision
associated with one scenario.
An Intensity review concerned the expected ordering of revisions
associated with two scenarios.
For each candidate, the reviewers received the relevant scenario
text or texts, the fixed domain background $\mathcal C_D$, and the
definition of the relation type to be assessed.
The proposed revision directions and intensity orderings, as well as scenario identifiers such as P/M prefixes, were withheld during the initial assessment.
Thus, the reviewers were asked to derive the expected relation from the
stated conditions, rather than to endorse a supplied label.

\paragraph{Independent judgments.}
Before discussing the candidates with each other or consulting the
construction-stage labels, the two reviewers independently determined
the expected relation and assessed whether it was justified by the
scenario text and $\mathcal C_D$.

\begin{itemize}
    \item \textbf{Direction.}
    Each reviewer assessed whether the stated conditions justified an
    upward or downward revision relative to the neutral forecast.
    The judgment concerned the direction of the scenario-induced
    revision, not whether the future trajectory should rise or fall
    relative to the last historical observation.
    A direction was not inferred from a generic event description
    when the conditions connecting that event to the target were
    missing or ambiguous.

    \item \textbf{Intensity.}
    Each reviewer assessed whether the two scenarios supported a
    common directional implication and a justified ordering of
    revision strength.
    This required identifying the strength attribute that changed
    and checking that other relevant comparison conditions remained
    fixed.
    The reviewer then determined which condition supported a larger
    revision in that direction.
    For an upward implication, the stronger condition should induce
    a larger upward revision; for a downward implication, it should
    induce a larger downward revision.
    More emphatic wording or greater event severity was not sufficient
    unless the stated difference justified a stronger target-relevant
    implication.
\end{itemize}

\paragraph{Discussion, revision, and exclusion.}
After the initial independent judgments were completed, the reviewers
compared their assessments.
The construction-stage proposals were then made available and considered
alongside the reviewers' judgments during discussion.
The proposed labels served as candidates to be examined, not as
authoritative answers that the reviewers were expected to recover.
Discussion focused on whether the stated conditions and $\mathcal C_D$
provided a sufficient justification for the relation.
Candidates requiring clarification of the target connection, relevant
conditions, or the attribute being compared were revised and reviewed
again by both experts.
Relations that remained insufficiently justified were excluded.
The following Energy pipeline cases illustrate the distinction between retaining a relation after clarification and excluding an unsupported ordering:

\begin{itemize}
    \item \textbf{Retained after clarification (Direction).}
    The initial wording did not clarify how equipment constraints
    affected gas deliveries to Henry Hub.
    The revised scenario explicitly connected the constraints to
    a reduction in gas reaching the hub, supporting an upward
    neutral-relative revision.
    \begin{itemize}
        \item \emph{Initial}:
        ``Output constraints on compression equipment along pipelines reduce \ldots''
        \item \emph{Revised} (\texttt{E-PI}):
        ``Output \ldots{} pipelines \emph{feeding Henry Hub}
        reduce the amount of natural gas
        \emph{that actually reaches the hub}.''
    \end{itemize}

    \item \textbf{Excluded (Intensity).}
    Greater equipment-damage severity alone did not justify
    a larger reduction in gas delivered to Henry Hub when
    both conditions already specified a complete shutdown.
    The proposed ordering was therefore excluded.
    \begin{itemize}
        \item \emph{Proposed pair}:
        ``\ldots{} completely shut down due to
        \emph{equipment damage}''
        $\rightarrow$
        ``\ldots{} completely shut down due to
        \emph{more severe equipment damage}''
    \end{itemize}
\end{itemize}

\paragraph{Contrast pair construction.}
For Contrast, a domain expert formed pairs from scenarios with independently validated opposing Direction relations.
For example, the Economy capital-flow scenarios were paired at
corresponding strength levels:
\texttt{M-CF-P1}$\leftrightarrow$\texttt{M-CF-M1},
\texttt{M-CF-P2}$\leftrightarrow$\texttt{M-CF-M2}, and
\texttt{M-CF-P3}$\leftrightarrow$\texttt{M-CF-M3}.


\subsubsection{Cross-Anchor Applicability Checks}
\label{app:cross_anchor_applicability}

Relation-level validation assessed whether the scenario
text and fixed domain background $\mathcal C_D$ justified an expected
direction or ordering.
We separately checked whether each retained Direction and Intensity relation remained
applicable at the selected forecasting anchors.
Even when a relation was justified by the scenario and background,
an anchor's forecasting state could introduce ambiguity or conflicting
conditions, or make its application depend on additional unstated
assumptions.
The cross-anchor checks therefore examined whether such issues
undermined the application of the retained relation at each anchor.

\paragraph{Review unit and coverage.}
The unit of review was an \emph{anchor--relation
application}: one retained relation applied to one selected
forecasting anchor.
The same two domain experts who conducted the relation-level
validation examined every application in their respective domain,
using the checklist in
Table~\ref{tab:cross_anchor_applicability_checklist}.
Each application was reviewed by both experts.
The applicability of each Contrast pair follows from that of both corresponding Direction relations at the same anchor.
Each retained Direction and Intensity relation was reviewed across all 38 anchors in its respective domain, yielding 5,130 anchor--relation applications in total.

\paragraph{Information provided to the reviewers.}
For each application, the reviewers received the validated scenario text with the retained relation specification, the fixed domain background $\mathcal C_D$, and a line plot of the anchor's numerical history.
They also received source-grounded textual information describing conditions during the history window.
This contextual information was used to check whether circumstances at the anchor introduced ambiguities or conflicts that could compromise
the applicability of the already validated relation.

\input{tables/apdx-valid-checklist}

\paragraph{Scope of the applicability assessment.}
For each review item in Table~\ref{tab:cross_anchor_applicability_checklist}, the Direction and Intensity questions were assessed using the following criteria:
\begin{itemize}
    \item \textbf{Target applicability.}
    Whether the relevant entities, roles, relationships, and scope
    were consistent with the forecasting target and the stated
    scenario conditions.
    \item \textbf{State compatibility.}
    Whether the anchor's numerical history and source-grounded
    contextual information introduced ambiguities or conflicting
    conditions affecting the expected relation.
    \item \textbf{Assumption sufficiency.}
    Whether applying the expected relation required additional
    unstated or anchor-specific assumptions.
\end{itemize}

\paragraph{Cross-anchor validation example: Capital flows under contrasting historical states.}

In the Economy domain, experts examined a capital-flow scenario involving a
reallocation from foreign-currency sovereign bonds to U.S. Treasuries:
\emph{a net reallocation involving foreign-currency sales and
U.S. dollar purchases.}
The specified transactions were expected to increase demand for
U.S. dollars against the relevant foreign currencies, supporting
an upward neutral-relative revision of the Broad U.S. Dollar Index.
The experts assessed whether this Direction relation remained
applicable across two historical anchors with contrasting
market conditions.

\begin{itemize}
    \item 
    \textbf{Anchor A:}
    The dollar index rose from 119.9 to 120.2, while the expected
    interest-rate gap widened in favor of the U.S.

    \item 
    \textbf{Anchor B:}
    The dollar index fell from 120.5 to 120.2, while the expected
    interest-rate gap narrowed, reducing the U.S. advantage.
\end{itemize}

The anchors exhibited opposing dollar-index trends and
macroeconomic contexts despite sharing the same last observation
of 120.2.
The interest-rate gap refers to the difference between U.S.
interest rates and those of major trading partners.
The experts applied the three validation criteria as follows:

\begin{itemize}
    \item 
    \textbf{Target applicability assessment.}
    The experts examined whether the specified foreign-currency
    sales and U.S. dollar purchases provided a relevant mechanism
    for upward pressure on the trade-weighted dollar index.
    They confirmed that the transactions involved relevant
    currencies and that the dollar-buying channel remained
    applicable at both anchors.

    \item 
    \textbf{State compatibility assessment.}
    The experts examined whether the contrasting historical
    trends and interest-rate conditions undermined the expected
    upward revision.
    No anchor-specific conditions were identified that invalidated the expected Direction relation at either anchor.
    In particular, Anchor B's historical dollar decline and
    narrowing interest-rate gap did not negate the additional
    dollar-demand pressure from the specified future transactions.
    An upward neutral-relative revision did not require the
    scenario-conditioned forecast to exceed the last observed
    value of 120.2, but only to exceed its own neutral forecast.

    \item 
    \textbf{Assumption sufficiency assessment.}
    The scenario explicitly specified the foreign-currency sales
    and U.S. dollar purchases underlying the expected revision.
    The experts confirmed that the Direction relation could be
    justified without assuming additional future interest-rate
    changes or other unstated anchor-specific conditions.
\end{itemize}

Both anchors satisfied all three criteria, and the experts
retained the same upward Direction relation despite their
contrasting historical states.
The source-grounded contextual information used for these
applicability checks was not provided as additional model input.
The retained relation did not require identical neutral forecasts
or scenario-induced revision magnitudes across anchors.

\paragraph{Review outcome.}
Under the reviewed information and stated scenario conditions, no anchor-specific ambiguities, conflicting conditions, or additional unstated assumptions were found to invalidate the retained relations. No anchors were excluded or replaced on this basis.


\subsubsection{Validation Outcomes}
\label{app:validation_outcomes}

Table~\ref{tab:validation_outcomes} summarizes the construction-stage
validation outcomes for candidate Direction and Intensity relations
and the final relation inventory across domains.
Initial inter-reviewer agreement reflects the independent judgments made before discussion and disclosure of the proposed labels.
Clarifications were not limited to initially disputed relations; some relations with initial agreement were also revised to make their conditions or wording more precise.
The alternative and irrelevant scenario texts, relation labels, comparison pairs,
attributes held fixed or varied, and applicability across the
selected anchors were finalized before model evaluation.

\input{tables/apdx-valid-outcomes}

For cross-anchor applicability, both reviewers in each domain examined every retained Direction and Intensity relation at all 38 selected anchors in their respective domain.
The 50 relations in Energy, 40 in Economy, and 45 in Politics yielded 5,130 anchor--relation applications
($38 \times (50+40+45)$). All applications were retained after review.

For each anchor, its factual scenario and the finalized alternative and irrelevant scenarios together form its controlled scenario set.

\subsection{Empirical Matched Sets}
\label{app:matched_sets}

Empirical matched sets are constructed separately after the anchors and scenario inventories are fixed.
Matched cases are sought in the full candidate-window pool available for empirical matching, rather than only among the finalized anchors or candidates that passed anchor screening.
For each anchor--scenario pair $(i,s)$, empirical matching considers cases with (i) a similar numerical history, (ii) similar history-only forecasts from the frozen construction references, and (iii) source evidence that the same scenario condition actually occurred in comparable forecasting settings.
Numerical and semantic matching use separate criteria, and observed future target trajectories are not used during matching.

\subsubsection{Numerical Matching}
To reduce dependence on absolute target levels, histories and construction-reference forecasts are expressed relative to each case's last observed value and scaled by a construction-only robust domain scale $\omega_D$.
For a single-series domain, $\omega_D$ is the median non-zero change between consecutive available observations; for a multi-entity domain, this scale is first computed separately for each entity and then aggregated by the median across entities.
For anchor $i$, we define
\begin{equation}
\widetilde{x}_{i,\ell}
=
\frac{x_{i,\ell}-x_{i,L}}{\omega_D},
\qquad
\widetilde{b}_{i,h}^{(m)}
=
\frac{b_{i,h}^{(m)}-x_{i,L}}{\omega_D},
\end{equation}
where $b_{i,h}^{(m)}$ denotes the history-only forecast from construction reference $m$.

For a candidate matched case $j$ drawn from this search pool, the history distance from anchor $i$ is
\begin{equation}
d_x(i,j)
=
\sqrt{
\frac{1}{L}
\sum_{\ell=1}^{L}
\left(
\widetilde{x}_{i,\ell}
-
\widetilde{x}_{j,\ell}
\right)^2
},
\end{equation}
and the corresponding reference-forecast distance is computed separately for each construction reference:
\begin{equation}
d_b^{(m)}(i,j)
=
\sqrt{
\frac{1}{H}
\sum_{h=1}^{H}
\left(
\widetilde{b}_{i,h}^{(m)}
-
\widetilde{b}_{j,h}^{(m)}
\right)^2
}.
\end{equation}

\paragraph{Percentile-based similarity criteria.}
To avoid domain-specific raw-distance thresholds, these distances are converted to percentile ranks.
For each anchor $i$, candidate $j$ is ranked relative to the corresponding distances between $i$ and the same-domain construction windows in $\mathcal K_D^{(-i)}$, whose history--future intervals do not overlap that of $i$.
Let $r_x(i,j)$ and $r_b^{(m)}(i,j)$ denote the resulting history and reference-forecast distance percentiles, with lower values indicating greater similarity.
The current release uses a common threshold $\alpha=0.10$, and a candidate passes numerical matching only if
\begin{equation}
r_x(i,j)\le\alpha,
\qquad
\max_m r_b^{(m)}(i,j)\le\alpha.
\label{eq:matched_numerical_gate}
\end{equation}
Thus, a matched case must satisfy the lower-decile distance threshold defined by the admissible construction windows for both numerical history and every construction-reference forecast.

\subsubsection{Semantic Matching and Matched-Set Selection}
Numerically admissible cases are retained only when source evidence independently verifies that the same scenario condition actually occurred and was relevant to the target under comparable timing, scope, and roles.
Self-matches and repeated instances of the same underlying event are removed, and temporal redundancy within each forecasting entity is controlled using the observation-step separation rule in Appendix~\ref{app:final_anchor_selection}.

\paragraph{Matched-set selection and support.}
Among cases satisfying all numerical and semantic criteria, we define
\begin{equation}
r_{\mathrm{num}}(i,j)
=
\frac{1}{2}r_x(i,j)
+
\frac{1}{2M}
\sum_{m=1}^{M}r_b^{(m)}(i,j),
\end{equation}
where $M$ is the number of construction references.
Admissible cases are ordered by $r_{\mathrm{num}}(i,j)$, with ties resolved by earlier forecast origin.
Up to four cases are retained for each anchor--scenario pair, and pairs with fewer than three admissible cases are marked unsupported.

\paragraph{Empirical role and coverage.}
The persistent-deviation criterion used for anchor screening is not imposed on candidate matched cases; matched-set selection follows the numerical and semantic criteria and ranking rules above.
These cases provide empirical references for neutral-relative revision trajectories, rather than counterfactual outcomes for an anchor or estimates of causal effects.
Grounded Comparison provides complementary empirical evidence when supported pairs are available.
Coverage could be constrained by the difficulty of finding cases with both similar forecasting states and compatible scenario occurrences within the limited observation period permitted by strict knowledge-cutoff requirements.

\subsection{Leakage Controls}
\label{app:freeze_audit}

The domain background context uses only reference material published by December 31, 2023, while candidate windows are drawn from post-2023 observations.
Factual evidence extraction and factual scenario composition are outcome-blind.
Empirical matching does not use observed future target trajectories.
Construction-reference forecasts are used for numerical screening and empirical matching, while outputs and results generated during the subsequent model evaluation are not used to select anchors or define benchmark components.





%% file: tables/apdx-valid-checklist.tex
\begin{table*}[t]
    \centering
    \small
    \setlength{\tabcolsep}{4pt}
    \caption{Checklist for cross-anchor applicability of Direction and Intensity relations.}
    \label{tab:cross_anchor_applicability_checklist}
    \begin{tabular}{@{}p{0.24\textwidth}p{0.35\textwidth}p{0.35\textwidth}@{}}
        \toprule
        \textbf{Review item / question}
        & \textbf{Direction}
        & \textbf{Intensity} \\
        \midrule
        \midrule

        \textbf{Target applicability}
        & Is the scenario applicable to the forecasting target
        at this anchor?
        & Are both scenarios applicable to the forecasting target
        at this anchor? \\[0.6em]

        \midrule

        \textbf{State compatibility}
        & Does the anchor's state introduce ambiguity
        or conflicts with the expected direction?
        & Does the anchor's state make the expected intensity ordering between the two scenarios ambiguous or introduce conditions that conflict with it? \\[0.6em]

        \midrule

        \textbf{Assumption sufficiency}
        & Does the expected direction require additional unstated
        or anchor-specific assumptions?
        & Does the expected intensity ordering require additional
        unstated or anchor-specific assumptions? \\

        \bottomrule
    \end{tabular}
\end{table*}

%% file: tables/apdx-valid-outcomes.tex
\begin{table*}[t]
    \centering
    \small
    \setlength{\tabcolsep}{5pt}
    \renewcommand{\arraystretch}{1.12}
    \caption{Domain-wise outcomes of relation-level validation
    and final relation inventory.}
    \label{tab:validation_outcomes}
    \begin{tabular}{@{}p{0.35\textwidth}rrrr@{}}
        \toprule
        \textbf{Statistic / Domain}
        & \textbf{Energy}
        & \textbf{Economy}
        & \textbf{Politics}
        & \textbf{Total} \\
        \midrule
        \midrule

        \multicolumn{5}{@{}l}{\textit{Initial independent review}} \\
        Candidate Direction
        & 31 & 24 & 28 & 83 \\
        Candidate Intensity
        & 21 & 17 & 20 & 58 \\
        Total candidates
        & 52 & 41 & 48 & 141 \\[0.3em]

        Initial agreement (Direction)
        & 29/31 (93.5\%)
        & 23/24 (95.8\%)
        & 26/28 (92.9\%)
        & 78/83 (94.0\%) \\

        Initial agreement (Intensity)
        & 18/21 (85.7\%)
        & 15/17 (88.2\%)
        & 16/20 (80.0\%)
        & 49/58 (84.5\%) \\

        Total initial agreement
        & 47/52 (90.4\%)
        & 38/41 (92.7\%)
        & 42/48 (87.5\%)
        & 127/141 (90.1\%) \\

        \midrule
        \multicolumn{5}{@{}l}{\textit{Discussion and reassessment}} \\
        Excluded Direction
        & 1 & 0 & 1 & 2 \\
        Excluded Intensity
        & 1 & 1 & 2 & 4 \\
        Total excluded
        & 2 & 1 & 3 & 6 \\[0.3em]

        Clarified and retained Direction
        & 3 & 2 & 4 & 9 \\
        Clarified and retained Intensity
        & 5 & 3 & 6 & 14 \\
        Total clarified and retained
        & 8 & 5 & 10 & 23 \\

        \midrule
        \multicolumn{5}{@{}l}{\textit{Final relation inventory}} \\
        Direction
        & 30 & 24 & 27 & 81 \\

        Intensity
        & 20 & 16 & 18 & 54 \\

        Contrast pairs from validated Direction
        & 15 & 12 & 12 & 39 \\

        Restraint scenarios
        & 3 & 3 & 3 & 9 \\

        \bottomrule
    \end{tabular}
\end{table*}

%% file: appendix/c_prompt.tex
\section{Construction Prompts and Model Inputs}
\label{app:prompts_inputs}

\subsection{Construction Prompt Templates}

The templates below cover domain-background construction and factual-scenario grounding.
Alternative and irrelevant scenario construction and relation validation are described separately in Appendix~\ref{app:controlled_scenarios}.

\subsubsection{Domain Background Context Construction}
\label{app:background_prompt}

The same prompt structure is used across domains to construct the fixed domain background context $\mathcal C_D$ from pre-cutoff reference material.
The prompt separates general, reusable domain mechanisms from benchmark-period events and requires each mechanism to retain its applicability conditions and uncertainty (Table~\ref{tab:domain_background_construction}).

\subsubsection{Event Extraction}
\label{app:event_extraction_prompt}

The \emph{event extraction} prompt identifies source-supported conditions that are relevant to the forecast interval and correspond to mechanisms in $\mathcal C_D$.
It records their timing, scope, status, mechanism, and applicability, and instructs the extractor not to infer conditions from numerical target trajectories, model forecasts, or other construction-derived signals.
Unsupported claims and unresolved information are recorded as unresolved items rather than treated as verified conditions (Table~\ref{tab:event_extraction}).

\subsubsection{Factual Scenario Composition}
\label{app:scenario_composition_prompt}

The \emph{scenario composition} prompt selects up to two verified conditions based only on source support, target and temporal relevance, and mechanism applicability, and compresses them into a concise, mechanism-preserving factual scenario.
It preserves necessary qualifications while excluding realized target information, other construction-derived signals, unsupported assumptions, and explicit forecasting guidance (Table~\ref{tab:scenario_composition}).

\subsubsection{Source and Semantic Review}
\label{app:scenario_review_prompt}

The \emph{source and semantic review} prompt audits the proposed factual scenario against the original evidence and $\mathcal C_D$.
It checks source support, timing and scope, mechanism applicability, preservation of uncertainty, and whether the scenario contains realized target outcomes or unsupported causal interpretations (Table~\ref{tab:review}).
Scenarios requiring substantive correction must be revised and reviewed again before acceptance.

\subsection{Model-Facing Input Examples}

\subsubsection{Domain Background Context}
\label{app:input_domain_background}

The domain background context provides fixed, domain-level knowledge shared across all anchors within each domain.
It defines the forecasting target and summarizes general mechanisms through which external conditions may affect it.
For Energy, these mechanisms cover weather-driven demand, production, pipelines, storage, LNG and trade, and industry or fuel substitution.
For Economy, they cover monetary policy and relative interest rates, relative economic conditions, global risk sentiment, and foreign developments and capital flows.
For Politics, they cover economic and public-service conditions, government performance, leadership and party unity, policy positions and issue salience, and political competition.
Each background also states relevant applicability conditions and qualifications rather than assuming that the same directional effect holds in every setting.

The background contains no anchor-specific events, future scenarios, or other time-varying information from the benchmark period.
The full domain background contexts $\mathcal{C}_D$ for all three domains are provided in Tables~\ref{tab:domain_background_energy}--\ref{tab:domain_background_economy}.

\subsubsection{Factual Scenario}
\label{app:input_factual_scenario}

The factual scenario summarizes source-supported external conditions relevant to the forecast horizon, preserving their reported status and uncertainty.
Across Energy, Economy, and Politics, it preserves the timing, scope, roles, and other information needed to interpret the corresponding mechanism in $\mathcal C_D$, while excluding realized target outcomes and explicit forecasting guidance.
The scenario is constructed from the verified textual evidence and provided as the factual condition for forecasting.
Examples from each domain are provided in Table~\ref{tab:factual}.

\subsubsection{Controlled Alternative Scenarios}
\label{app:input_controlled_scenarios}

Controlled alternative scenarios describe plausible future conditions under the same forecasting state, varying only the scenario condition.
They are constructed from mechanisms in the corresponding domain background context and reused across anchors to support controlled comparisons of scenario responses.
Domain-specific examples are provided in Tables~\ref{tab:alternatives_energy}--\ref{tab:alternatives_politics}.

\subsubsection{Irrelevant Scenario}
\label{app:input_irrelevant_scenario}

Irrelevant scenarios describe conditions unrelated to the forecasting target under the corresponding domain background context.
They test whether the model avoids unnecessary forecast revisions when the provided scenario is not relevant to the target.
An example is provided in Table~\ref{tab:irrelevant}.

\subsection{Evaluation Prompt Templates}

All scenario-conditioned systems receive the same benchmark content, while model-specific input formats and native interfaces are preserved.
In the neutral condition, the numerical history and domain background remain unchanged, but no future scenario information is supplied.
The same no-scenario condition is applied to each matched real case using its own forecasting state.
For directly prompted LLMs, the numerical history and textual inputs are combined in a single prompt.
For context-aware forecasting systems, the same textual content is provided through their designated context inputs while preserving the original model interface.
The following templates specify these input formats.

\subsubsection{Directly Prompted LLMs}

Directly prompted LLMs receive the numerical history, domain background context, and scenario in a single prompt.
They are instructed to use scenario information when relevant and to rely on the numerical history and domain background when the scenario is unrelated to the target.
The output is restricted to a JSON array containing exactly the required number of forecast values, with no additional explanation.
The prompt template is provided in Table~\ref{tab:llm_template}.

\subsubsection{Context-aware Forecasting Systems}

For context-aware forecasting systems, the numerical history is provided through each model's native time-series input, while the domain background context and scenario are passed through the designated textual input using a minimal template.
For DoubleCast~\citep{zheng2026overcoming}, this follows the simple context format used in CAF-7M, while MoiraiAgent~\citep{yan2026moiraiagent} processes the same textual content through its GPT-4o backbone.
We do not add an explicit instruction to ignore irrelevant scenarios, so that relevance handling remains part of each system's native context-processing behavior.
The textual input template is provided in Table~\ref{tab:multimodal_template}.

\input{tables/apdx-prompt-domain-background-construction}
\input{tables/apdx-prompt-event-extraction}
\input{tables/apdx-prompt-scenario-composition}
\input{tables/apdx-prompt-review}

\input{tables/apdx-prompt-domain-background-energy}
\input{tables/apdx-prompt-domain-background-politics}
\input{tables/apdx-prompt-domain-background-economy}

\input{tables/apdx-prompt-factual}

\input{tables/apdx-prompt-alternatives-energy}
\input{tables/apdx-prompt-alternatives-economy}

\input{tables/apdx-prompt-alternatives-politics}

\input{tables/apdx-prompt-irrelevant}

\input{tables/apdx-llm-template}
\input{tables/apdx-multimodal-template}


\subsection{Case Visualization}

\begin{figure*}[t]
    \centering
    \includegraphics[width=\linewidth]{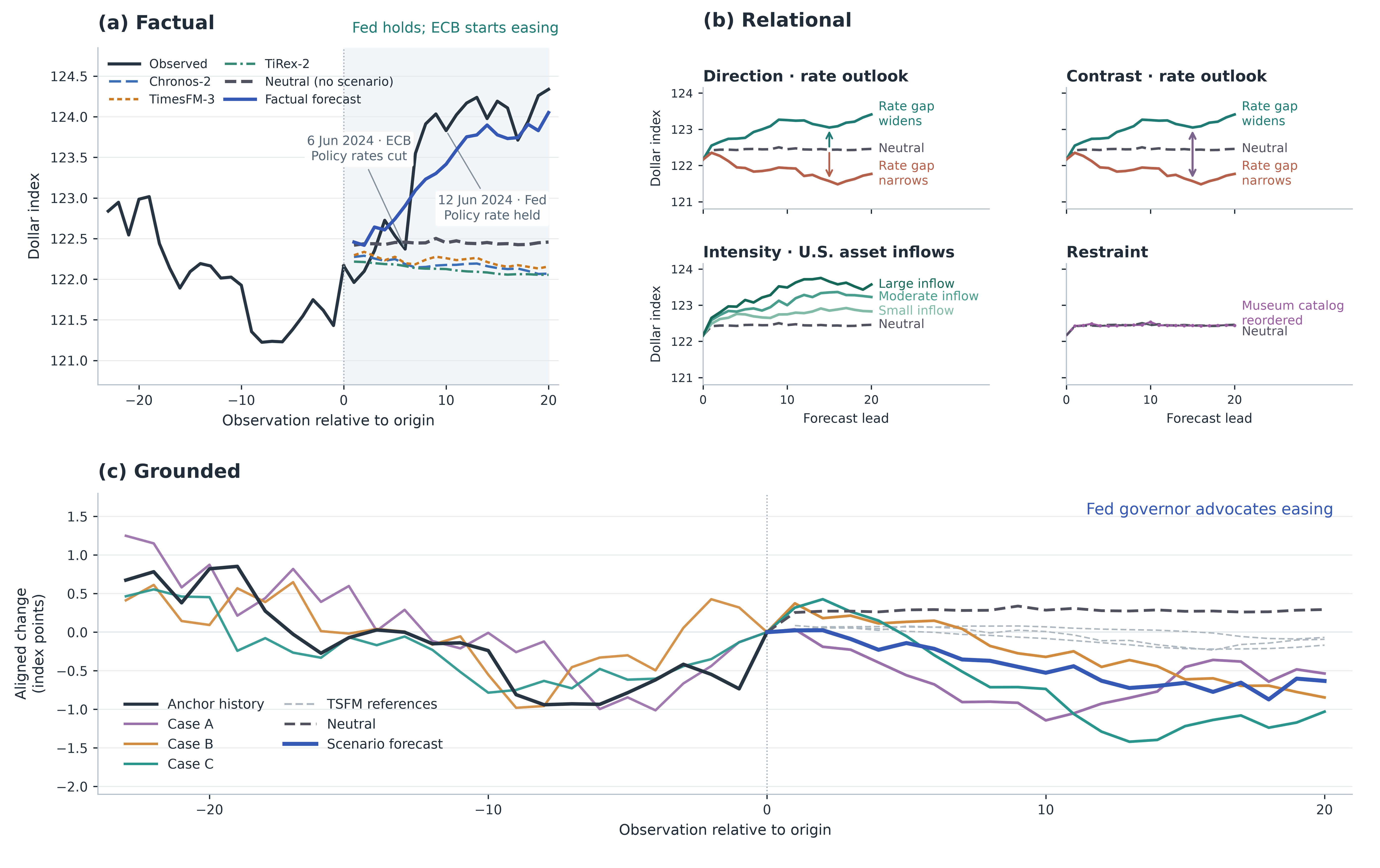}
    \caption{Example of WIT's three comparisons in the Economy domain.
    (a) \textbf{Factual}: the numerical history and observed future, timestamped supporting evidence, factual and neutral forecasts, and numerical TSFM references.
    (b) \textbf{Relational}: alternative and irrelevant scenarios and their forecast responses, illustrating Direction, Contrast, Intensity, and Restraint.
    (c) \textbf{Grounded}: the neutral-relative revision trajectory under an alternative scenario, compared with empirical references from matched real cases.}
    \label{fig:wit_visualization}
\end{figure*}

Figure~\ref{fig:wit_visualization} visualizes an Economy-domain case, including the observed time series, forecasts under different scenario conditions, and empirical references from matched real cases.

%% file: tables/apdx-prompt-domain-background-construction.tex
\begin{table}[H]
\caption{Domain background context construction prompt.}
\vspace{-8pt}
\label{tab:domain_background_construction}

\begin{tcolorbox}[
    width=\textwidth,
    colback=white,
    colframe=black,
    boxsep=5pt,
    top=0pt,
    bottom=0pt,
    arc=2pt
]
\begin{lstlisting}[
    basicstyle=\ttfamily\scriptsize,
    breaklines=true,
    breakatwhitespace=false,
    columns=fullflexible,
    keepspaces=true,
    showstringspaces=false
]
ROLE
Construct a fixed domain background for a scenario-conditioned forecasting benchmark.

INPUT
- Target definition and units
- Background cutoff date
- Pre-cutoff reference passages

TASK
Write:
1. a concise definition of the forecasting target, and
2. a compact set of general driver mechanisms that can affect it.

For each mechanism, describe:
- the relevant condition or driver,
- its conditional upward or downward pressure where supported,
- assumptions, applicability conditions, and limitations required for that interpretation.

RULES
- Use only the supplied pre-cutoff reference material.
- Include only general mechanisms that can be reused across benchmark episodes.
- Exclude benchmark episodes, anchor-specific entities or dates, post-cutoff events,
  dated forecasts, contemporary settings, and episode-specific expectations.
- Do not predict the target or assign numerical effect sizes.
- Do not infer that any mechanism actually occurs in a benchmark episode.
- Preserve uncertainty and conditionality; do not turn context-dependent relations
  into universal directional rules.
- Omit mechanisms not supported by the supplied sources.

OUTPUT
background_text,
mechanism_map: mechanism, supported directions, required assumptions,
               limitations, supporting source IDs, unresolved_items
\end{lstlisting}
\end{tcolorbox}

\end{table}

%% file: tables/apdx-prompt-event-extraction.tex
\begin{table}[H]
\vspace{-20pt}
\caption{Event extraction prompt.}
\label{tab:event_extraction}
\vspace{-8pt}
\begin{tcolorbox}[
    width=\textwidth,
    colback=white,
    colframe=black,
    boxsep=5pt,
    top=0pt,
    bottom=0pt,
    arc=2pt
]
\begin{lstlisting}[
    basicstyle=\ttfamily\scriptsize,
    breaklines=true,
    breakatwhitespace=false,
    columns=fullflexible,
    keepspaces=true,
    showstringspaces=false
]
ROLE
Extract source-supported conditions relevant to scenario annotation.

INPUT
- Target and fixed domain background
- Forecast interval
- Relevant entity, geography or scope, and role or relation if applicable
- Dated source passages

TASK
Identify distinct events or conditions that:
1. are explicitly supported by the supplied sources,
2. overlap or remain relevant during the forecast interval, and
3. correspond to a mechanism in the fixed domain background.

For each condition, record:
- factual condition,
- timing and scope,
- event status (observed, expected, announced, alleged, resolved, etc.),
- applicable mechanism and required assumptions,
- conflicting evidence or unresolved information.

RULES
- Use only the supplied sources.
- Do not use outside knowledge.
- Do not infer events from the target trajectory, model forecasts, reference-relative
  direction, or calibrated departure score.
- Do not infer that a condition caused, dominated, or explains the target movement.
- Preserve uncertainty and distinguish occurrence from expectation, announcement,
  allegation, and resolution.
- Mark missing applicability conditions explicitly rather than filling them in.
- If no usable condition is supported, return INSUFFICIENT.

OUTPUT
evidence_status, extracted_events, unresolved_items
\end{lstlisting}
\end{tcolorbox}

\end{table}

%% file: tables/apdx-prompt-scenario-composition.tex
\begin{table}[H]
\vspace{-20pt}
\caption{Scenario composition prompt.}
\label{tab:scenario_composition}
\vspace{-8pt}

\begin{tcolorbox}[
    width=\textwidth,
    colback=white,
    colframe=black,
    boxsep=5pt,
    top=0pt,
    bottom=0pt,
    arc=2pt
]
\begin{lstlisting}[
    basicstyle=\ttfamily\scriptsize,
    breaklines=true,
    breakatwhitespace=false,
    columns=fullflexible,
    keepspaces=true,
    showstringspaces=false
]
ROLE
Convert source-supported event records into a concise factual scenario.

INPUT
- Target and fixed domain background
- Source-supported event records
- Forecast interval

TASK
Select at most two conditions using only evidence-based criteria:
1. strength of source support,
2. relevance to the forecasting target,
3. temporal relevance to the forecast interval, and
4. applicability of the corresponding domain mechanism.

Express the selected conditions as one or two concise, mechanism-preserving sentences.

RULES
- Use only the supplied event records.
- Do not use the realized target trajectory, construction forecasts,
  reference-relative direction, or calibrated departure score.
- Preserve qualifications required for timing, scope, status, uncertainty,
  and mechanism validity.
- Remove identifying details that are unnecessary for interpretation while
  preserving roles, relationships, scope, and other attributes required for
  the mechanism to remain meaningful.
- Do not reveal target values, realized target movement, numerical effect sizes,
  or instructions about how the forecast should change.
- Do not introduce unsupported assumptions or claim that a selected condition
  caused, dominated, or fully explained the realized future.
- If no sufficiently supported and forecast-relevant condition is available,
  return HOLD_SAMPLE.

OUTPUT
status, scenario_text, selected_event_ids, mechanism_ids,
claim_map, retained_qualifications, hold_reason
\end{lstlisting}
\end{tcolorbox}

\end{table}

%% file: tables/apdx-prompt-review.tex
\begin{table}[H]
\caption{Source and semantic review prompt.}
\vspace{-8pt}
\label{tab:review}
\begin{tcolorbox}[
    width=\textwidth,
    colback=white,
    colframe=black,
    boxsep=5pt,
    top=0pt,
    bottom=0pt,
    arc=2pt
]
\begin{lstlisting}[
    basicstyle=\ttfamily\scriptsize,
    breaklines=true,
    breakatwhitespace=false,
    columns=fullflexible,
    keepspaces=true,
    showstringspaces=false
]
ROLE
Audit a proposed factual scenario against its original evidence and the fixed
domain specification.

INPUT
- Fixed domain background and mechanism map
- Proposed scenario and claim-to-source mapping
- Source-supported event records and original source passages
- Forecast interval
- Relevant entity, geography or scope, and role or relation if applicable

CHECK
For every substantive claim, verify:
1. direct support from the cited source,
2. correct timing, scope, entity, and event status,
3. applicability of the assigned mechanism and its required assumptions,
4. preservation of uncertainty and necessary qualifications,
5. fidelity after compression or de-identification,
6. absence of realized target information, forecasting instructions,
   or unsupported causal interpretation.

RULES
- Use only the supplied evidence and fixed domain specification.
- Do not use outside knowledge.
- Do not use or infer validity from the realized target trajectory,
  construction forecasts, reference-relative direction, or calibrated departure score.
- Do not repair unsupported claims with plausible assumptions.
- Any unsupported claim necessary for the scenario's interpretation requires HOLD_SAMPLE.
- A revised scenario must be reviewed again before acceptance.

OUTPUT
verdict: PASS or HOLD_SAMPLE
claim_verdicts, required_corrections, unresolved_items
\end{lstlisting}
\end{tcolorbox}

\end{table}

%% file: tables/apdx-prompt-domain-background-energy.tex
\begin{table}[H]
\caption{Domain background context $\mathcal{C}$ in the Energy domain.}
\vspace{-8pt}
\label{tab:domain_background_energy}

\begin{tcolorbox}[
    width=\textwidth,
    colback=white,
    colframe=black,
    boxsep=5pt,
    top=0pt,
    bottom=0pt,
    arc=2pt
]
\begin{lstlisting}[
    basicstyle=\ttfamily\scriptsize,
    breaklines=true,
    breakatwhitespace=false,
    columns=fullflexible,
    keepspaces=true,
    showstringspaces=false
]
The target is the Henry Hub natural gas spot price (USD/MMBtu). Higher demand or less available gas generally pushes prices up; lower demand or more available gas generally pushes prices down.

- Weather demand: Cold winter weather increases gas use for heating. Hot summer weather increases electricity use for air conditioning, which can increase gas use at power plants. Warmer winters or cooler summers reduce these demands.

- Production: Freezing weather and equipment shutdowns can reduce gas production. Production restarts or growth add supply. Storms can also reduce gas use by disrupting businesses and electricity demand, partly offsetting production losses.

- Pipelines: A blocked pipeline can leave consumers short of gas while trapping excess gas near producers. Prices may rise where gas cannot arrive and fall where gas cannot leave. The effect on Henry Hub depends on how the affected pipeline connects to the hub.

- Storage: Less stored gas than usual for the season leaves a smaller reserve to meet demand; more stored gas provides a larger reserve. Withdrawals above expectations or injections below expectations suggest a tighter supply-demand balance; the reverse suggests a looser balance.

- LNG and trade: More LNG exports leave less U.S. gas available for domestic use. Export-terminal shutdowns reduce exports and leave more gas at home; restarts reverse this effect. Overseas events matter when they affect U.S. gas imports or exports, within the limits of available infrastructure.

- Industry and fuel substitution: Higher industrial activity increases gas use. When nuclear or hydropower output falls, gas-fired plants may burn more gas to replace it. More renewable generation or switching to other fuels can reduce gas use.
\end{lstlisting}
\end{tcolorbox}

\end{table}

%% file: tables/apdx-prompt-domain-background-politics.tex
\begin{table}[H]
\caption{Domain background context $\mathcal{C}$ in the Politics domain.}
\label{tab:domain_background_politics}
\vspace{-12pt}
\begin{tcolorbox}[
    width=\textwidth,
    colback=white,
    colframe=black,
    boxsep=5pt,
    top=0pt,
    bottom=0pt,
    arc=2pt
]
\begin{lstlisting}[
    basicstyle=\ttfamily\scriptsize,
    breaklines=true,
    breakatwhitespace=false,
    columns=fullflexible,
    keepspaces=true,
    showstringspaces=false
]
The target is the selected party's reported polling support in its country, measured as a percentage. A higher value means greater support for that party; a lower value means less support. Effects depend on the party's role and on which parties voters credit, blame, or view as credible alternatives.

- Economic conditions and public services: Improving living standards, employment, housing, or public services can benefit parties credited with those outcomes, while deteriorating conditions can hurt parties held responsible.

- Government performance and policy delivery: Effective administration and delivery of valued policies can strengthen support, while policy failures, unpopular decisions, or broken commitments can weaken it. Responsibility may be shared unevenly in coalition governments.

- Leadership, integrity, and party unity: Favorable evaluations of party leadership, competence, and cohesion can strengthen support, while leadership disputes, defections, scandals, or perceived integrity failures can weaken it.

- Policy positions and issue salience: Parties can gain when issues on which they are viewed as credible become more important and their positions align with voter priorities. They can lose when salient positions conflict with those priorities or are viewed as ineffective.

- Political competition and alternatives: Stronger campaigns, alliances, or coordination can consolidate support, while credible rivals, fragmentation, or shifts toward competing parties can reduce it. Dissatisfaction with one party does not necessarily benefit any particular alternative.
\end{lstlisting}
\end{tcolorbox}

\end{table}

%% file: tables/apdx-prompt-domain-background-economy.tex
\begin{table}[H]
\vspace{-15pt}
\caption{Domain background context $\mathcal{C}$ in the Economy domain.}
\label{tab:domain_background_economy}
\vspace{-12pt}
\begin{tcolorbox}[
    width=\textwidth,
    colback=white,
    colframe=black,
    boxsep=5pt,
    top=0pt,
    bottom=0pt,
    arc=2pt
]
\begin{lstlisting}[
    basicstyle=\ttfamily\scriptsize,
    breaklines=true,
    breakatwhitespace=false,
    columns=fullflexible,
    keepspaces=true,
    showstringspaces=false
]
The target is the Nominal Broad U.S. Dollar Index (DTWEXBGS), a trade-weighted measure of the U.S. dollar against major U.S. trading partners. A higher index means a stronger dollar; a lower index means a weaker dollar.

- Monetary policy and relative interest rates: Higher expected U.S. interest rates relative to other economies generally support the dollar, while lower relative rates or faster expected U.S. easing tend to weaken it.

- Relative economic conditions: Stronger U.S. growth or economic data relative to major trading partners generally support the dollar, while weaker relative conditions tend to weigh on it. Inflation matters mainly through its effects on expected monetary policy and real returns.

- Global risk sentiment: Financial stress, geopolitical uncertainty, or weaker risk appetite often increase demand for liquid U.S. dollar assets and strengthen the dollar. Improving global risk sentiment can reduce this safe-haven demand.

- Foreign developments and capital flows: Stronger demand for U.S. assets or weaker monetary and economic conditions abroad generally support the dollar, while capital shifting toward foreign assets can weaken it. Effects are larger for currencies with greater weights in the broad dollar index.
\end{lstlisting}
\end{tcolorbox}

\end{table}

%% file: tables/apdx-prompt-factual.tex
\begin{table}[H]
\caption{Example factual scenario $s^{*}$ for each domain.}
\label{tab:factual}
\vspace{-12pt}
\begin{tcolorbox}[
    width=\textwidth,
    colback=white,
    colframe=black,
    boxsep=5pt,
    top=0pt,
    bottom=0pt,
    arc=2pt
]
\begin{lstlisting}[
    basicstyle=\ttfamily\scriptsize,
    breaklines=true,
    breakatwhitespace=false,
    columns=fullflexible,
    keepspaces=true,
    showstringspaces=false
]
(Energy) -> Weather-driven demand mechanism
Hotter-than-normal weather across several U.S. regions raises cooling demand and natural gas use for power generation.

(Economy) -> Monetary policy and relative interest rates mechanism
The European Central Bank cuts its policy interest rates, while the Federal Reserve leaves its policy rate unchanged.

(Politics) -> Leadership disputes and party cohesion mechanism
Internal leadership disputes culminate in the removal of the party's leader, deepening uncertainty over the party's direction.
\end{lstlisting}
\end{tcolorbox}

\end{table}





%% file: tables/apdx-prompt-alternatives-energy.tex
\begin{table}[H]
\vspace{-10pt}
\caption{Examples of alternative scenarios in the Energy domain.}
\vspace{-12pt}
\label{tab:alternatives_energy}

\begin{tcolorbox}[
    width=\textwidth,
    colback=white,
    colframe=black,
    boxsep=5pt,
    top=0pt,
    bottom=0pt,
    arc=2pt
]
\begin{lstlisting}[
    basicstyle=\ttfamily\scriptsize,
    breaklines=true,
    breakatwhitespace=false,
    columns=fullflexible,
    keepspaces=true,
    showstringspaces=false
]
(Weather Demand) -> E-W-M2
- Moderate temperature changes relative to seasonal norms reduce heating or cooling needs, reducing natural gas consumption for heating or power generation.

(Production) -> E-PR-Pb
- Major operating constraints on compression and gas processing equipment at some production facilities substantially reduce natural gas output at each facility.

(Pipelines) -> E-PI-P2
- Major output constraints on compression equipment along pipelines feeding Henry Hub substantially reduce the amount of natural gas that actually reaches the hub.

(Storage) -> E-ST-P2
- Natural gas inventories available for withdrawal and supply from underground storage facilities are substantially below typical seasonal levels.

(LNG and trade) -> E-LN-P2
- Higher operating loads on liquefaction equipment at export terminals substantially increase the amount of gas withdrawn from the domestic pipeline network to produce LNG for export.

(Industry and fuel substitution) -> E-FS-M4
- Industrial facilities substantially reduce production and shift a large share of process-heat supply from natural gas boilers to fuel-oil boilers.
\end{lstlisting}
\end{tcolorbox}

\end{table}







%% file: tables/apdx-prompt-alternatives-economy.tex
\begin{table}[H]
\vspace{-15pt}
\caption{Examples of alternative scenarios in the Economy domain.}
\vspace{-12pt}
\label{tab:alternatives_economy}

\begin{tcolorbox}[
    width=\textwidth,
    colback=white,
    colframe=black,
    boxsep=5pt,
    top=0pt,
    bottom=0pt,
    arc=2pt
]
\begin{lstlisting}[
    basicstyle=\ttfamily\scriptsize,
    breaklines=true,
    breakatwhitespace=false,
    columns=fullflexible,
    keepspaces=true,
    showstringspaces=false
]
(Monetary policy and relative interest rates) -> M-IR-P2
- U.S. monetary policy shifts in a more restrictive direction relative to that of major trading partners, moderately raising the expected U.S. interest-rate path relative to theirs.

(Relative economic conditions) -> M-EC-M2
- U.S. real household consumption growth in goods and services weakens moderately relative to that of major trading partners, worsening the outlook for U.S. economic activity relative to theirs.

(Global risk sentiment) -> M-RI-P2
- Funding stress in global financial markets heightens risk aversion, moderately increasing international investors' total demand for liquid short-term U.S. Treasury securities.

(Foreign developments and capital flows) -> M-CF-M2
- International bond investors make a moderate net shift of investment funds from U.S. Treasuries to sovereign bonds denominated in major trading partners' currencies, selling U.S. dollars and buying foreign currencies.
\end{lstlisting}
\end{tcolorbox}

\end{table}





%% file: tables/apdx-prompt-alternatives-politics.tex
\begin{table}[H]
\vspace{-15pt}
\caption{Examples of alternative scenarios in the Politics domain.}
\vspace{-12pt}
\label{tab:alternatives_politics}

\begin{tcolorbox}[
    width=\textwidth,
    colback=white,
    colframe=black,
    boxsep=5pt,
    top=0pt,
    bottom=0pt,
    arc=2pt
]
\begin{lstlisting}[
    basicstyle=\ttfamily\scriptsize,
    breaklines=true,
    breakatwhitespace=false,
    columns=fullflexible,
    keepspaces=true,
    showstringspaces=false
]
(Economic conditions and public services) -> P-EC-Md
- Under economic policies backed by the target party, workers' pay falls substantially relative to essential living costs and layoffs increase substantially.

(Government performance and policy delivery) -> P-GV-Pd
- Implemented policies backed by the target party substantially cut income tax on wages and substantially raise public pension payments.

(Leadership, integrity, and party unity) -> P-LD-MU
- Senior figures in the target party publicly challenge its leader and threaten to leave over its economic programme.

(Policy positions and issue salience) -> P-PO-P2
- Amid broad public demands for tighter controls on unauthorized entry, the target party proposes a substantial tightening of border controls, stopping short of the requested level.

(Political competition and alternatives) -> P-CO-M1
- Rival parties form a political alliance to compete directly for the target party's established electoral base.
\end{lstlisting}
\end{tcolorbox}

\end{table}






%% file: tables/apdx-prompt-irrelevant.tex
\begin{table}[H]
\caption{Example of an irrelevant scenario.}
\vspace{-7pt}
\label{tab:irrelevant}

\begin{tcolorbox}[
    width=\textwidth,
    colback=white,
    colframe=black,
    boxsep=5pt,
    top=0pt,
    bottom=0pt,
    arc=2pt
]
\begin{lstlisting}[
    basicstyle=\ttfamily\scriptsize,
    breaklines=true,
    breakatwhitespace=false,
    columns=fullflexible,
    keepspaces=true,
    showstringspaces=false
]
A university library updates its online manuscript catalog, correcting publication dates and adding subject tags to help researchers find historical documents more easily.
\end{lstlisting}
\end{tcolorbox}

\end{table}


%% file: tables/apdx-llm-template.tex
\begin{table}[H]
\caption{Prompt template for directly prompted LLMs.}
\vspace{-7pt}
\label{tab:llm_template}
\begin{tcolorbox}[
    width=\textwidth,
    colback=white,
    colframe=black,
    boxsep=5pt,
    top=0pt,
    bottom=0pt,
    arc=2pt
]
\begin{lstlisting}[
    basicstyle=\ttfamily\scriptsize,
    breaklines=true,
    breakatwhitespace=false,
    columns=fullflexible,
    keepspaces=true,
    showstringspaces=false
]
Forecast the target time series using its history, domain background, and relevant scenario information. If the scenario is unrelated to the target, rely on the historical time series and domain background.

Domain background: {domain_background}

Historical time series: {history_timeseries}

Scenario: {scenario_text}

Forecast the next {prediction_horizon} observations.
Output only a JSON array of exactly {prediction_horizon} numbers, ordered from earliest to latest, in the same units and scale as the history. Do not include explanations or Markdown.
\end{lstlisting}
\end{tcolorbox}

\end{table}

  
  
  
  

%% file: tables/apdx-multimodal-template.tex
\begin{table}[H]
\caption{Prompt template for context-aware forecasting systems.}
\vspace{-7pt}
\label{tab:multimodal_template}

\begin{tcolorbox}[
    width=\textwidth,
    colback=white,
    colframe=black,
    boxsep=5pt,
    top=0pt,
    bottom=0pt,
    arc=2pt
]
\begin{lstlisting}[
    basicstyle=\ttfamily\scriptsize,
    breaklines=true,
    breakatwhitespace=false,
    columns=fullflexible,
    keepspaces=true,
    showstringspaces=false
]
Domain background: {domain_background}
Scenario: {scenario_text}
\end{lstlisting}
\end{tcolorbox}

\end{table}


%% file: appendix/d_eval_detail.tex
\section{Additional Evaluation Details}
\label{app:eval_details}

\subsection{Evaluation Populations and Aggregation}
\label{app:eval_aggregation}

For Relational Comparison, relation outcomes are first averaged within each anchor, anchors are weighted equally within each relation category, and the four category scores are weighted equally in Overall SLA.
This prevents anchors or categories with more predefined relations from having greater influence on the aggregate score.

For Grounded Comparison, supported anchor--scenario pairs are weighted equally within each domain.
GCE is reported separately by domain, while GSG is macro-averaged with equal domain weight so that domains with more supported pairs do not dominate the overall score.

\subsection{Factual Comparison Details}
\label{app:factual_details}

Within each domain, FSG is computed from the ratio of aggregate factual and neutral errors rather than by averaging anchor-specific percentage changes.
Overall FSG is then macro-averaged with equal domain weight.
CWR is the strict share of anchors satisfying $e_i^{s_i^{*}}<e_i^0$, with ties not counted as improvements, and is likewise macro-averaged across domains.


\subsection{Relational Comparison Details}
\label{app:relation_details}

\subsubsection{Relation Eligibility and Scoring}
\label{app:relation_scoring}

The predefined relations are applied to all anchors within
their respective domains using the fixed controlled scenario
sets, following the cross-anchor applicability checks
described in Appendix~\ref{app:cross_anchor_applicability}.
Direction is evaluated for individual controlled scenarios, Contrast for predefined opposing pairs, Intensity for adjacent levels of an ordered scenario sequence, and Restraint for irrelevant scenarios.
Each relation produces a binary outcome according to the criteria in Section~\ref{sec:eval_details} and the practical tolerances defined below.
Direction, Contrast, and Intensity assess horizon-level
relations between forecast revisions without separately
scoring effect onset or duration.

\subsubsection{Practical Tolerances}
\label{app:tolerances}

The relational criteria use practical tolerances to avoid treating numerically negligible forecast changes as meaningful responses.
For Direction, Contrast, and Intensity, $\eta_D$ specifies the minimum meaningful revision; for Restraint, $\tau_D$ specifies the largest revision treated as negligible.
These are practical-effect thresholds rather than statistical significance thresholds.

\paragraph{Common scale-based rule.}
Because target units and numerical variation differ across domains, we derive the tolerances from a common scale-based rule.
The scale statistics are computed from historical observations available before the evaluation period and fixed for all anchors and evaluated models.
For forecasting entity $e$ in domain $D$, let
\begin{equation}
\mathcal U_{D,e}
=
\left\{
|x_{e,t_k}-x_{e,t_{k-1}}|
:
k=2,\ldots,T,\;
x_{e,t_k}\neq x_{e,t_{k-1}}
\right\},
\end{equation}
and define
\begin{equation}
v_{D,e}
=
\operatorname{median}\mathcal U_{D,e}.
\end{equation}
Thus, $v_{D,e}$ represents a typical non-zero change between consecutive available observations.
For a single-entity domain, $v_D=v_{D,e}$; for a multi-entity domain,
\begin{equation}
v_D=\operatorname{median}_{e}v_{D,e}.
\end{equation}

We define
\begin{equation}
\epsilon_D=\rho v_D,
\qquad \rho=0.1,
\end{equation}
using the same pre-specified coefficient for every domain.
The current release uses the same scale for both practical thresholds,
\begin{equation}
\eta_D=\tau_D=\epsilon_D.
\end{equation}
Applying this rule gives $\eta_D=\tau_D=0.01$, $0.02$, and $0.10$ for Energy, Economy, and Politics, respectively.
If no non-zero change is available, the documented numerical resolution of the target is used as the fallback scale.

\subsubsection{Tolerance Sensitivity}
\label{app:tolerance_sensitivity}

We repeat Relational Comparison with
$\rho\in\{0.05,0.10,0.20\}$, corresponding to $0.5\times$, $1\times$, and $2\times$ the default tolerance.
The same multiplier is applied to both $\eta_D$ and $\tau_D$ in every domain.

\input{tables/apdx-tolerance-sensitivity}

As expected, larger tolerances make Direction, Contrast, and Intensity stricter and Restraint more permissive.
Although individual category scores change, the overall
model rankings and Overall SLA remain relatively stable
across the tested settings.

\paragraph{Sign-only sensitivity.}
We additionally evaluate Direction, Contrast, and Intensity
with $\eta_D=0$, retaining only the expected direction
or ordering without a minimum magnitude requirement.
Restraint is omitted because it is inherently defined
by revision magnitude.

\input{tables/apdx-sign-only-sensitivity}

The broad model-level response patterns persist without
a nonzero magnitude requirement, although absolute scores change.


\subsection{Grounded Comparison Details}

Recall that $\Delta_{i,h}(s)=\widehat y_{i,h}^{\,s}-\widehat y_{i,h}^{\,0}$ denotes the scenario-induced revision, and $g_i(s)$ its mean squared discrepancy from the neutral-relative deviations of the empirical matched cases.

\subsubsection{Empirical References for Evaluation}
\label{app:grounded_references}

Grounded Comparison uses the fixed empirical matched set $\mathcal P_i(s)$ for each supported anchor--scenario pair $(i,s)$.
For matched case $j$, the evaluated model generates its own neutral forecast and defines
$d_{j,h}=y_{j,h}-\widehat y_{j,h}^{\,0}$.
Thus, both anchor revisions and matched-case deviations are measured relative to the evaluated model's own neutral baseline; the construction-reference forecasts used for matching are not used as evaluation references.

\subsubsection{Pair and Matched-Case Weighting}
\label{app:grounded_weighting}

Forecast steps and matched cases receive equal weight within each pair, and supported pairs receive equal weight within each domain.
A matched case reused across multiple empirical sets
contributes to the discrepancy of each corresponding pair,
without changing the pair-level weighting.
Domain-level GSG is computed from the ratio of aggregated
discrepancies rather than by averaging pair-specific
percentage gains.
GSG is macro-averaged with equal domain weight, while GCE is reported separately by domain.

\subsubsection{No-Revision Baseline}
\label{app:grounded_no_revision}

The no-revision baseline uses the same pairs, matched sets, and weighting, but sets $\Delta_{i,h}(s)=0$.
GSG therefore measures improvement in empirical alignment relative to retaining the neutral forecast.

\subsubsection{Interpretation of GCE and GSG}
\label{app:grounded_interpretation}

Matched real cases provide empirical references rather
than counterfactual ground truth.
GCE aggregates timestep-wise discrepancies between
forecast revisions and neutral-relative deviations
observed in matched cases, allowing temporal mismatches
to affect the score.
However, these references do not establish the correct
effect onset or duration, and zero GCE does not imply
a uniquely correct alternative future.
GSG measures whether scenario conditioning improves
this empirical alignment relative to making no revision.


\subsection{Stochastic Decoding Variability and Evaluation Stability}
\label{app:stochastic_decoding_variability}

\subsubsection{Analysis protocol}
\label{app:decoding_analysis_protocol}

We analyze saved forecasts from the three original runs
(Appendix~\ref{app:repeated_runs}) for all five
scenario-conditioned systems, without additional inference.
Model inputs, decoding settings, domain-specific practical
tolerances, relation labels, matched sets, and aggregation
rules remain unchanged.

The analysis covers all 114 anchors (38 per domain) and their
neutral, factual, controlled-alternative, and irrelevant-scenario
forecasts, along with the 20 supported anchor--scenario pairs
and 62 matched-case assignments used in Grounded Comparison.
Neutral replacements are applied consistently across all
comparisons sharing the same neutral output.
Only valid 20-step numerical forecasts are included.

\subsubsection{Neutral–neutral variability}
\label{app:neutral_neutral_variability}

\paragraph{Comparison unit.}
We compare the neutral forecasts stored from the three original runs, keeping the decoding settings and practical tolerances unchanged.
For each model and anchor, we evaluate the three unordered run pairs $(1,2)$, $(1,3)$, and $(2,3)$.
A pair is included only when both neutral forecasts are valid under the original output-validity criteria.
With valid outputs in all three runs, the 38 anchors in each domain yield $38 \times \binom{3}{2} = 114$ pairwise comparisons per model--domain group.
The three comparisons within an anchor reuse the same three neutral forecasts; they are therefore not independent decoding repetitions.

\paragraph{Pairwise differences.}
For a fixed model, let $\hat y^0_{i,h,r}$ denote the neutral forecast for anchor $i$, forecast step $h$, and run $r$.
For each pair $r<r'$, we compute
\begin{equation}
\label{eq:neutral_pairwise_differences}
\begin{aligned}
d^{\mathrm{Dir}}_{i,r,r'}
=
\left|
\frac{1}{H}\sum_{h=1}^{H}
\left(
\hat y^0_{i,h,r}-\hat y^0_{i,h,r'}
\right)
\right|, \qquad
d^{\mathrm{Res}}_{i,r,r'}
=
\frac{1}{H}\sum_{h=1}^{H}
\left|
\hat y^0_{i,h,r}-\hat y^0_{i,h,r'}
\right|.
\end{aligned}
\end{equation}
The first quantity measures the absolute difference between horizon-averaged forecasts, matching the summary used by Direction.
The second measures the mean absolute pointwise difference, matching the quantity used by Restraint.
It captures trajectory differences that can cancel when averaged over the horizon.

\paragraph{Aggregation and reporting.}
Within each model--domain group, we pool the valid pairwise differences across anchors.
We normalize $d^{\mathrm{Dir}}$ by the domain's Direction tolerance $\eta_D$ and $d^{\mathrm{Res}}$ by its Restraint tolerance $\tau_D$.
For each normalized diagnostic, the empirical 95th percentile is the $\lceil 0.95N \rceil$th smallest value, where $N$ is the number of valid pairs.
For $N=114$, this is the 109th ordered value.

We also count comparisons satisfying
$d^{\mathrm{Dir}}_{i,r,r'}>\eta_D$ or
$d^{\mathrm{Res}}_{i,r,r'}>\tau_D$ and report each count together
with its denominator and percentage.
Both diagnostics use the same valid pair set.
Table~\ref{tab:neutral_neutral_variability} reports these
descriptive statistics for all five systems and three domains.
Under the evaluated decoding settings, most neutral–neutral differences remained within the practical tolerances, although occasional exceedances indicate non-negligible variability in some cases.

\input{tables/apdx-neutral-variability}

\subsubsection{Neutral-replacement sensitivity}
\label{app:neutral_replacement_sensitivity}


\paragraph{Replacement protocol.}
We assess sensitivity to the choice of neutral forecast by holding
scenario-conditioned forecasts fixed and replacing each run's
neutral forecasts with those from either of the other two runs.
For each model, this yields six ordered replacements
$r \leftarrow r'$ with $r \neq r'$, where
\begin{equation}
\Delta_{i,h}^{r \leftarrow r'}(s)
=
\hat y^s_{i,h,r}-\hat y^0_{i,h,r'}.
\end{equation}
Each replacement is evaluated against the original results of run $r$,
using unchanged scenario inputs, practical tolerances, and evaluation
populations.
No additional forecasts are generated.

\paragraph{Recomputation across comparisons.}
For Factual Comparison, we recompute FSG and CWR using the replacement
neutral forecasts while keeping factual forecasts fixed.
For Relational Comparison, we recompute Direction and Restraint
decisions and count those that change relative to the original run.
Contrast and Intensity remain unchanged because the shared neutral
forecast cancels in their pairwise scenario comparisons.
For Grounded Comparison, we replace both anchor and matched-case
neutral forecasts, preserving the matched-case assignments and
the original sharing of neutral outputs.
We then recompute GCE, GSG, and the corresponding no-revision
baseline using the replacement neutral forecasts.

\paragraph{Aggregation and reporting.}
For each model, we report changes under all six neutral replacements
relative to their corresponding original runs, with domain-wise values
ordered as Energy / Economy / Politics.
FSG, CWR, and GSG are reported as signed percentage-point changes,
GCE as signed relative percentage changes, and Direction and Restraint
as counts of flipped decisions.
The results are reported in
Table~\ref{tab:neutral_replacement_domain}.
Across the six neutral replacements, most Direction and Restraint decisions remained unchanged, and GCE and GSG showed limited variation, although FSG was more sensitive to the choice of neutral forecast.

\input{tables/apdx-neutral-replacement}

\subsubsection{Decision reproducibility}
\label{app:decision_reproducibility}

\paragraph{Comparison unit and scope.}
We compare decisions for the same anchor--relation applications
across the three original complete runs, allowing both neutral
and scenario-conditioned forecasts to vary.
Using the original evaluation criteria, we assess Direction
and Restraint separately.
This analysis examines whether individual applications receive
consistent pass/fail decisions, beyond the reported variation
in category-level scores.

\paragraph{Agreement and persistent violations.}
For a given model, domain, and category, let $N$ denote the
number of evaluated anchor--relation applications.
Let $n_{\mathrm{same}}$ count applications receiving identical
decisions in all three runs, $n_{\mathrm{all}}$ count applications
violated in all three runs, and $n_{\mathrm{any}}$ count applications
violated in at least one run.
We define three-run unanimous agreement and the
persistent-violation rate as
\begin{equation}
\mathrm{Agreement}
=
100\,\frac{n_{\mathrm{same}}}{N},
\qquad
\mathrm{PersistentViolation}
=
100\,\frac{n_{\mathrm{all}}}{n_{\mathrm{any}}}.
\end{equation}
Unanimous agreement includes both applications that consistently
pass and those that consistently violate the relation.
The persistent-violation rate instead measures how often an
observed violation recurs in all three runs.
When $n_{\mathrm{any}}=0$, we report the persistent-violation
rate as \texttt{N/A}, since no violations were observed.
Such entries are excluded when reporting the minimum
persistent-violation rate.

\paragraph{Aggregation and reporting.}
Table~\ref{tab:decision_reproducibility} reports both diagnostics
for every model--domain combination, separately for Direction
and Restraint.
Agreement is reported as $n_{\mathrm{same}}/N$, and persistent
violations as $n_{\mathrm{all}}/n_{\mathrm{any}}$, each accompanied
by its percentage.
With complete decision coverage, Direction comprises
1,140, 912, and 1,026 applications in Energy, Economy, and Politics,
respectively, while Restraint comprises 114 applications
in each domain.
These denominators count distinct anchor--relation applications,
rather than individual run-specific decisions.
Most Direction and Restraint decisions remained consistent across the three runs, and observed violations frequently recurred on the same applications, suggesting that the identified failures were not solely explained by stochastic decoding.

\input{tables/apdx-decision-reproduce}

\subsubsection{Interpretation and limitations}
\label{app:decoding_interpretation}

Most neutral--neutral differences remained within the practical
tolerances, with occasional exceedances.
Under the six neutral replacements, most Direction and Restraint
decisions remained unchanged, and GCE and GSG showed limited
variation, whereas FSG was more sensitive.
Across the three original runs, most decisions were consistent,
and violations frequently recurred on the same applications,
suggesting that these violations were not solely explained by
run-to-run decoding variability.
Together, these diagnostics alleviate concerns about dependence
on a particular neutral forecast while identifying some
decoding-sensitive cases.

These findings are limited to the evaluated models, anchors,
decoding settings, and practical tolerances.
The three neutral--neutral pairs and six ordered replacements
reuse outputs from only three original runs and do not constitute
additional independent decoding repetitions.
Shared neutral forecasts across scenarios and matched cases
also introduce dependence among evaluation outcomes.
Thus, the reported counts should not be interpreted as
independent samples, and the analysis cannot characterize
rare decoding outcomes or isolate the causes of persistent
violations.

%% file: tables/apdx-tolerance-sensitivity.tex
\begin{table*}[t]
\centering
\small
\caption{
Sensitivity of Relational Comparison to the practical tolerance.
Scores are macro-averaged across the three domains.
$0.5\times$, $1\times$, and $2\times$ scale both $\eta_D$ and $\tau_D$ relative to the default setting.
}
\label{tab:tolerance_sensitivity}
\setlength{\tabcolsep}{4.5pt}
\renewcommand{\arraystretch}{1.08}
\begin{tabular}{llccccc}
\toprule
\textbf{Tolerance} & \textbf{Model}
& \textbf{Direction}
& \textbf{Contrast}
& \textbf{Intensity}
& \textbf{Restraint}
& \textbf{Overall} \\
\midrule
\midrule

\multirow{5}{*}{$0.5\times$}
& Mistral (7B)  & 73.45 & 70.95 & 57.85 & 66.96 & 67.30 \\
& Llama-3 (70B) & 82.35 & 80.05 & 68.65 & 75.44 & 76.62 \\
& GPT-4o        & 89.10 & 86.65 & 77.10 & 81.58 & 83.61 \\
& DoubleCast    & 90.70 & 84.45 & 71.20 & 61.99 & 77.09 \\
& MoiraiAgent   & 91.35 & 86.20 & 79.80 & 67.25 & 81.15 \\
\midrule

\multirow{5}{*}{$1\times$}
& Mistral (7B)  & 71.80 & 68.10 & 54.30 & 70.66 & 66.21 \\
& Llama-3 (70B) & 80.90 & 77.60 & 65.21 & 79.53 & 75.81 \\
& GPT-4o        & 87.79 & 84.61 & 73.90 & 85.09 & 82.85 \\
& DoubleCast    & 89.40 & 82.10 & 67.80 & 66.47 & 76.44 \\
& MoiraiAgent   & 90.11 & 83.90 & 76.51 & 71.83 & 80.59 \\
\midrule

\multirow{5}{*}{$2\times$}
& Mistral (7B)  & 68.80 & 64.65 & 49.25 & 76.90 & 64.90 \\
& Llama-3 (70B) & 78.10 & 73.95 & 59.80 & 84.80 & 74.16 \\
& GPT-4o        & 85.45 & 81.05 & 68.80 & 89.77 & 81.27 \\
& DoubleCast    & 87.10 & 78.00 & 62.15 & 72.81 & 75.02 \\
& MoiraiAgent   & 87.90 & 79.80 & 71.30 & 77.19 & 79.05 \\
\bottomrule
\end{tabular}
\end{table*}

%% file: tables/apdx-sign-only-sensitivity.tex
\begin{table}[t]
\centering
\small
\caption{
Sign-only sensitivity of relational criteria, with $\eta_D=0$.
}
\label{tab:sign_sensitivity}
\setlength{\tabcolsep}{5pt}
\renewcommand{\arraystretch}{1.05}
\begin{tabular}{lccc}
\toprule
\textbf{Model}
& \textbf{Direction}
& \textbf{Contrast}
& \textbf{Intensity} \\
\midrule
\midrule
Mistral (7B)   & 75.62 & 73.18 & 61.84 \\
Llama-3 (70B)  & 84.06 & 82.17 & 72.48 \\
GPT-4o         & 90.53 & 88.61 & 81.24 \\
DoubleCast     & 92.01 & 86.82 & 75.93 \\
MoiraiAgent    & 92.64 & 88.47 & 83.52 \\
\bottomrule
\end{tabular}
\end{table}

%% file: tables/apdx-neutral-variability.tex
\begin{table*}[t]
    \centering
    \small
    \setlength{\tabcolsep}{4pt}
    \renewcommand{\arraystretch}{1.12}
    \caption{
    Neutral--neutral variability across three runs.
    P95 denotes the 95th percentile of pairwise differences
    normalized by $\eta_D$ for Direction and $\tau_D$ for Restraint.
    Exceedances are reported as count/valid pairs (percentage).
    Each model--domain group includes 114 pairwise comparisons
    from three valid neutral forecasts at each of 38 anchors.
    }
    \label{tab:neutral_neutral_variability}
    \begin{tabular}{@{}llrrrr@{}}
        \toprule
        \textbf{System}
        & \textbf{Domain}
        & \multicolumn{2}{c}{\textbf{Direction}}
        & \multicolumn{2}{c}{\textbf{Restraint}} \\
        \cmidrule(lr){3-4}
        \cmidrule(l){5-6}
        &
        & \textbf{P95}
        & \textbf{Exceedances}
        & \textbf{P95}
        & \textbf{Exceedances} \\
        \midrule
        \midrule

        Mistral (7B)
        & Energy
        & 1.12 & 7/114 (6.1\%)
        & 1.34 & 10/114 (8.8\%) \\

        & Economy
        & 0.74 & 2/114 (1.8\%)
        & 0.98 & 5/114 (4.4\%) \\

        & Politics
        & 0.96 & 5/114 (4.4\%)
        & 1.17 & 8/114 (7.0\%) \\

        \midrule

        Llama-3 (70B)
        & Energy
        & 0.88 & 3/114 (2.6\%)
        & 1.06 & 7/114 (6.1\%) \\

        & Economy
        & 0.61 & 1/114 (0.9\%)
        & 0.79 & 2/114 (1.8\%) \\

        & Politics
        & 0.83 & 4/114 (3.5\%)
        & 1.09 & 7/114 (6.1\%) \\

        \midrule

        GPT-4o
        & Energy
        & 0.63 & 2/114 (1.8\%)
        & 0.86 & 4/114 (3.5\%) \\

        & Economy
        & 1.06 & 6/114 (5.3\%)
        & 1.27 & 9/114 (7.9\%) \\

        & Politics
        & 0.71 & 2/114 (1.8\%)
        & 0.95 & 5/114 (4.4\%) \\

        \midrule

        DoubleCast
        & Energy
        & 1.04 & 6/114 (5.3\%)
        & 1.25 & 9/114 (7.9\%) \\

        & Economy
        & 0.58 & 1/114 (0.9\%)
        & 0.83 & 3/114 (2.6\%) \\

        & Politics
        & 0.90 & 4/114 (3.5\%)
        & 1.12 & 7/114 (6.1\%) \\

        \midrule

        MoiraiAgent
        & Energy
        & 0.94 & 4/114 (3.5\%)
        & 1.21 & 8/114 (7.0\%) \\

        & Economy
        & 0.67 & 2/114 (1.8\%)
        & 0.88 & 4/114 (3.5\%) \\

        & Politics
        & 0.79 & 3/114 (2.6\%)
        & 1.03 & 6/114 (5.3\%) \\

        \bottomrule
    \end{tabular}
\end{table*}

%% file: tables/apdx-neutral-replacement.tex
\begin{table*}[t]
    \centering
    \scriptsize
    \setlength{\tabcolsep}{3pt}
    \renewcommand{\arraystretch}{1.08}

    \caption{
    Neutral-replacement sensitivity across domains.
    Values are reported as Energy / Economy / Politics.
    FSG, CWR, and GSG are signed changes in percentage points;
    GCE is a signed relative percentage change.
    Direction and Restraint flips are counts, with respective
    denominators of 1,140 / 912 / 1,026 and 114 / 114 / 114.
    Each replacement holds scenario-conditioned forecasts fixed
    and reports changes relative to the original run.
    }
    \label{tab:neutral_replacement_domain}

    \resizebox{\textwidth}{!}{%
    \begin{tabular}{@{}llrrrrrr@{}}
        \toprule
        \textbf{Model}
        & \textbf{Swap}
        & \textbf{$\Delta$FSG (pp)}
        & \textbf{$\Delta$CWR (pp)}
        & \textbf{Dir. flips}
        & \textbf{Res. flips}
        & \textbf{$\Delta$GCE (\%)}
        & \textbf{$\Delta$GSG (pp)} \\
        \midrule
        \midrule

        Mistral (7B)
        & $1\leftarrow2$
        & -1.26/+1.11/-1.75
        & -2.6/+5.3/-5.3
        & 18/8/14
        & 3/1/2
        & +0.21/+0.26/+0.52
        & +1.44/+1.85/+2.11 \\

        & $1\leftarrow3$
        & +4.81/+5.86/+3.69
        & +2.6/+5.3/0.0
        & 39/18/30
        & 5/2/3
        & +1.33/+2.24/+3.20
        & +1.16/+0.35/-0.32 \\

        & $2\leftarrow1$
        & +1.25/-1.13/+1.73
        & 0.0/0.0/0.0
        & 24/13/21
        & 2/1/2
        & -1.02/-1.20/-1.78
        & +0.22/+0.06/+0.56 \\

        & $2\leftarrow3$
        & +6.03/+4.84/+5.37
        & +2.6/+5.3/0.0
        & 31/14/25
        & 4/2/4
        & +0.93/+1.36/+1.76
        & +0.09/-0.27/-0.59 \\

        & $3\leftarrow1$
        & -5.14/-6.34/-3.90
        & -2.6/0.0/-5.3
        & 37/9/18
        & 6/1/3
        & -1.92/-2.54/-2.99
        & +0.10/+0.38/+0.09 \\

        & $3\leftarrow2$
        & -6.49/-5.15/-5.75
        & -2.6/0.0/-5.3
        & 21/11/16
        & 3/2/1
        & -0.90/-1.02/-1.95
        & -0.13/-0.40/+1.08 \\

        \midrule

        Llama-3 (70B)
        & $1\leftarrow2$
        & -3.91/+0.98/-4.42
        & -2.6/0.0/-2.6
        & 15/6/10
        & 2/1/1
        & -1.11/-0.83/-0.50
        & +0.18/-0.09/+0.86 \\

        & $1\leftarrow3$
        & -4.10/-3.64/-2.79
        & -2.6/-2.6/-2.6
        & 24/10/21
        & 3/1/2
        & -1.75/-2.52/-2.80
        & +0.93/+0.96/+0.24 \\

        & $2\leftarrow1$
        & +3.72/-0.99/+4.18
        & 0.0/-7.9/+2.6
        & 18/8/15
        & 3/0/2
        & +1.11/+0.86/-0.59
        & -0.16/+0.04/+1.18 \\

        & $2\leftarrow3$
        & -0.18/-4.64/+1.55
        & -2.6/-5.3/+2.6
        & 21/7/19
        & 4/2/2
        & -0.73/-1.53/-2.45
        & +0.89/+0.71/-0.36 \\

        & $3\leftarrow1$
        & +3.86/+3.44/+2.66
        & 0.0/0.0/+5.3
        & 23/11/17
        & 2/1/3
        & +0.98/+2.14/+2.61
        & +0.50/-0.17/+0.26 \\

        & $3\leftarrow2$
        & +0.18/+4.37/-1.56
        & 0.0/0.0/0.0
        & 17/6/13
        & 3/1/2
        & -0.03/+1.10/+2.16
        & +0.48/+0.10/+0.99 \\

        \midrule

        GPT-4o
        & $1\leftarrow2$
        & +0.78/+3.97/+3.38
        & 0.0/+7.9/0.0
        & 5/2/6
        & 1/0/1
        & -1.49/-2.11/-2.02
        & -0.11/+0.40/-0.31 \\

        & $1\leftarrow3$
        & -2.61/+1.59/+0.99
        & -5.3/+5.3/0.0
        & 11/4/10
        & 2/1/2
        & +1.49/-0.14/-0.39
        & -1.06/+0.78/+0.18 \\

        & $2\leftarrow1$
        & -0.79/-4.17/-3.53
        & -2.6/0.0/-5.3
        & 6/3/7
        & 1/1/1
        & +1.78/+1.29/+2.05
        & -0.34/+1.09/+0.33 \\

        & $2\leftarrow3$
        & -3.43/-2.51/-2.49
        & -5.3/0.0/-2.6
        & 7/3/8
        & 1/1/2
        & +2.63/+2.08/+1.93
        & -0.26/+0.26/+0.02 \\

        & $3\leftarrow1$
        & +2.51/-1.60/-0.99
        & 0.0/-2.6/-5.3
        & 10/5/12
        & 2/1/2
        & -0.89/-0.44/+0.37
        & +0.03/+0.26/-0.15 \\

        & $3\leftarrow2$
        & +3.26/+2.40/+2.38
        & 0.0/+2.6/0.0
        & 6/2/7
        & 1/0/1
        & -2.30/-2.30/-2.00
        & -0.20/+0.20/+0.18 \\

        \midrule

        DoubleCast
        & $1\leftarrow2$
        & -1.98/-1.64/-1.63
        & -2.6/-2.6/-2.6
        & 16/4/11
        & 2/0/1
        & +1.19/+1.06/+1.46
        & -0.66/+0.15/-0.11 \\

        & $1\leftarrow3$
        & -1.11/-0.73/-0.44
        & -2.6/-2.6/0.0
        & 28/9/20
        & 5/1/2
        & -1.06/+0.49/+0.06
        & +1.08/-0.62/+1.00 \\

        & $2\leftarrow1$
        & +1.93/+1.61/+1.60
        & 0.0/+2.6/0.0
        & 18/5/12
        & 3/1/2
        & -0.71/-1.84/-1.95
        & -0.13/+1.16/+0.98 \\

        & $2\leftarrow3$
        & +0.85/+0.89/+1.17
        & 0.0/+2.6/0.0
        & 21/6/23
        & 3/1/2
        & -1.43/-1.05/-0.77
        & +0.42/+0.05/+0.10 \\

        & $3\leftarrow1$
        & +1.09/+0.72/+0.43
        & 0.0/+5.3/0.0
        & 33/8/20
        & 4/1/3
        & +0.14/+0.20/-0.67
        & +0.42/-0.50/+0.02 \\

        & $3\leftarrow2$
        & -0.85/-0.89/-1.17
        & -2.6/0.0/-5.3
        & 18/5/13
        & 2/0/2
        & +1.48/+1.12/+0.08
        & -0.46/-0.14/+1.05 \\

        \midrule

        MoiraiAgent
        & $1\leftarrow2$
        & +3.03/+0.54/+1.48
        & 0.0/0.0/+2.6
        & 13/5/10
        & 2/1/1
        & -0.61/-0.22/+1.43
        & +0.06/+0.56/-0.72 \\

        & $1\leftarrow3$
        & +3.56/+3.86/-1.86
        & 0.0/+2.6/-2.6
        & 27/9/17
        & 3/2/2
        & -1.58/-1.38/-1.95
        & -0.30/-0.60/+1.11 \\

        & $2\leftarrow1$
        & -3.18/-0.55/-1.54
        & -5.3/-2.6/0.0
        & 15/6/11
        & 2/1/2
        & +0.49/+0.51/-0.34
        & +0.15/-1.04/-1.09 \\

        & $2\leftarrow3$
        & +0.56/+3.39/-3.48
        & 0.0/+2.6/0.0
        & 18/7/13
        & 3/1/2
        & -1.57/-1.76/-2.25
        & +0.63/-0.18/-0.02 \\

        & $3\leftarrow1$
        & -3.75/-4.08/+1.85
        & -5.3/-5.3/+5.3
        & 23/11/20
        & 4/2/3
        & +2.48/+1.94/+1.36
        & -1.11/-0.27/-0.08 \\

        & $3\leftarrow2$
        & -0.56/-3.51/+3.31
        & 0.0/-5.3/+5.3
        & 15/6/11
        & 2/1/2
        & +1.46/+1.65/+1.85
        & -0.42/+0.39/+0.74 \\

        \bottomrule
    \end{tabular}%
    }
\end{table*}

%% file: tables/apdx-decision-reproduce.tex
\begin{table*}[t]
    \centering
    \small
    \setlength{\tabcolsep}{4pt}
    \renewcommand{\arraystretch}{1.12}

    \caption{
    Decision reproducibility across three runs.
    Agreement measures the proportion of anchor--relation
    applications receiving identical pass/fail decisions in all
    three runs.
    Persistent violations measure the proportion of applications
    violated in all three runs among those violated at least once.
    Each entry reports the count ratio and percentage.
    \texttt{N/A} indicates that no violations were observed.
    }
    \label{tab:decision_reproducibility}

    \begin{tabular}{@{}llrrrr@{}}
        \toprule
        & & \multicolumn{2}{c}{\textbf{Direction}}
        & \multicolumn{2}{c}{\textbf{Restraint}} \\
        \cmidrule(lr){3-4}
        \cmidrule(l){5-6}

        \textbf{Model}
        & \textbf{Domain}
        & \textbf{Agreement}
        & \textbf{Persistent violations}
        & \textbf{Agreement}
        & \textbf{Persistent violations} \\
        \midrule
        \midrule

        Mistral (7B)
        & Energy
        & 1062/1140 (93.2\%)
        & 266/344 (77.3\%)
        & 106/114 (93.0\%)
        & 27/35 (77.1\%) \\

        & Economy
        & 860/912 (94.3\%)
        & 229/281 (81.5\%)
        & 105/114 (92.1\%)
        & 29/38 (76.3\%) \\

        & Politics
        & 960/1026 (93.6\%)
        & 274/340 (80.6\%)
        & 106/114 (93.0\%)
        & 31/39 (79.5\%) \\

        \midrule

        Llama-3 (70B)
        & Energy
        & 1100/1140 (96.5\%)
        & 184/224 (82.1\%)
        & 109/114 (95.6\%)
        & 20/25 (80.0\%) \\

        & Economy
        & 881/912 (96.6\%)
        & 157/188 (83.5\%)
        & 110/114 (96.5\%)
        & 21/25 (84.0\%) \\

        & Politics
        & 980/1026 (95.5\%)
        & 187/233 (80.3\%)
        & 108/114 (94.7\%)
        & 22/28 (78.6\%) \\

        \midrule

        GPT-4o
        & Energy
        & 1118/1140 (98.1\%)
        & 120/142 (84.5\%)
        & 110/114 (96.5\%)
        & 14/18 (77.8\%) \\

        & Economy
        & 891/912 (97.7\%)
        & 99/120 (82.5\%)
        & 111/114 (97.4\%)
        & 15/18 (83.3\%) \\

        & Politics
        & 1001/1026 (97.6\%)
        & 122/147 (83.0\%)
        & 110/114 (96.5\%)
        & 16/20 (80.0\%) \\

        \midrule

        DoubleCast
        & Energy
        & 1117/1140 (98.0\%)
        & 104/127 (81.9\%)
        & 104/114 (91.2\%)
        & 31/41 (75.6\%) \\

        & Economy
        & 895/912 (98.1\%)
        & 80/97 (82.5\%)
        & 106/114 (93.0\%)
        & 34/42 (81.0\%) \\

        & Politics
        & 1000/1026 (97.5\%)
        & 110/136 (80.9\%)
        & 105/114 (92.1\%)
        & 35/44 (79.5\%) \\

        \midrule

        MoiraiAgent
        & Energy
        & 1115/1140 (97.8\%)
        & 94/119 (79.0\%)
        & 108/114 (94.7\%)
        & 28/34 (82.4\%) \\

        & Economy
        & 892/912 (97.8\%)
        & 79/99 (79.8\%)
        & 107/114 (93.9\%)
        & 28/35 (80.0\%) \\

        & Politics
        & 999/1026 (97.4\%)
        & 95/122 (77.9\%)
        & 106/114 (93.0\%)
        & 29/37 (78.4\%) \\

        \bottomrule
    \end{tabular}
\end{table*}

%% file: appendix/e_experiment.tex
\section{Experimental Setup}
\label{app:exp_details}

\subsection{Evaluated Models}

We evaluate three groups of models: directly prompted LLMs, context-aware forecasting systems, and time-series foundation models (TSFMs).

The directly prompted LLMs include Mistral-7B-Instruct v0.3~\citep{jiang2023mistral7b}, Llama-3-70B-Instruct~\citep{grattafiori2024llama}, and GPT-4o~\citep{openai2024gpt4technicalreport}.

For context-aware forecasting, we evaluate DoubleCast~\citep{zheng2026overcoming} and MoiraiAgent~\citep{yan2026moiraiagent}.
For DoubleCast, we retain the released architecture and Chronos-T5-Large numerical backbone, replace its original Qwen3-14B text encoder with Qwen1.5-14B-Chat, and perform cross-modal alignment on CAF-7M before evaluation, as detailed below.
For MoiraiAgent, we use the released GIFT-CTX
context-aware forecasting framework with Chronos-2 as its
forecasting tool, replacing its default GPT-5.1 LLM component
with GPT-4o to satisfy our knowledge-cutoff criterion.

The numerical TSFM baselines include Chronos-2~\citep{ansari2025chronos2}, TimesFM-3~\citep{jain2026timesfm3}, and TiRex-2~\citep{podest2026tirex2}.

\subsection{Model Versions and Knowledge Cutoffs}
\label{app:model_versions}

Table~\ref{tab:model_details} reports the forecasting and textual components used in our experiments.
For LLM-based components, we restrict model selection to models with explicitly documented knowledge cutoffs that precede the WIT evaluation period.
Because textual knowledge cutoffs do not characterize numerical time-series pretraining, possible overlap with WIT is audited separately below.

\input{tables/apdx-model-details}

\paragraph{DoubleCast text encoder.}
The released DoubleCast configuration uses Chronos-T5-Large as its numerical backbone and Qwen3-14B as its text encoder.
Because Qwen3-14B does not satisfy our requirement for an explicitly documented pre-WIT knowledge cutoff, we replace it with Qwen1.5-14B-Chat, whose documented knowledge cutoff is September 2023.
Qwen1.5-14B-Chat retains the 5,120-dimensional hidden representation expected by the released DoubleCast cross-attention interface, allowing the model architecture to remain unchanged.
Because replacing the encoder changes its semantic representation despite this dimensional compatibility, we re-align the text--time-series fusion components on CAF-7M before evaluation.
Following the released training procedure, we freeze
the pretrained numerical backbone and text encoder
and update the context-fusion components during alignment.
This alignment uses the same training source as DoubleCast and does not use any WIT numerical series, scenarios, factual evidence, or empirical matched cases.
We refer to this aligned configuration as DoubleCast throughout the experiments.

\paragraph{Time-series pretraining overlap audit.}
We separately inspect the disclosed pretraining sources of
the numerical forecasting components, since LLM knowledge
cutoffs do not address possible numerical-data contamination.
We compare the available source descriptions against the WIT
evaluation datasets and periods.
We found no documented overlap with the WIT evaluation series
in the disclosed training sources.
However, incomplete disclosure of pretraining corpora
precludes ruling out all potential overlap.

\subsection{Inference and Implementation Settings}
\label{app:inference_settings}

The directly prompted GPT-4o baseline and the GPT-4o component of MoiraiAgent are evaluated through API inference.
Mistral-7B-Instruct and Llama-3-70B-Instruct are served locally using vLLM on NVIDIA RTX PRO 6000 Blackwell GPUs with 96\,GB of GPU memory.
DoubleCast uses the released implementation with the
CAF-7M-aligned configuration described above.
Its point forecasts are obtained as the step-wise median
of generated forecast samples.
The numerical TSFMs use their released inference
implementations and checkpoints.

For LLM inference, we use temperature $1.0$ and top-$p=1.0$ where these parameters are exposed by the inference interface, corresponding to no additional temperature scaling or nucleus truncation.
Frequency and presence penalties are set to zero where applicable.
For directly prompted LLMs, the maximum output length is
512 tokens, sufficient for the required 20-step numerical
trajectory.
Decoding settings are held fixed across domains and scenario conditions.

Directly prompted LLM outputs must parse as a JSON array containing exactly $H=20$ finite numerical values.
If parsing fails or the output does not contain exactly $H=20$ finite numerical values,
we retry up to twice, preserving the numerical history
and scenario text while adding a formatting-only reminder.
No numerical values are manually corrected or imputed.

For Chronos-2, TimesFM-3, and TiRex-2, we use the median forecast ($q=0.5$) with their released inference configurations.
These models receive only the numerical history and serve as numerical-only baselines for factual forecasting performance.

\subsection{Repeated-Run Protocol}
\label{app:repeated_runs}

Each scenario-conditioned system is evaluated over three runs using identical benchmark inputs and inference settings.
Where deterministic seed control is available, we use seeds $42$, $43$, and $44$ for the three runs.
For API components without reliable deterministic seed control, the runs consist of independent calls under otherwise identical settings.

Within each run, one neutral forecast is generated per anchor and reused across all scenario comparisons associated with that anchor.
For Grounded Comparison, one neutral forecast is likewise generated for each matched real case and reused wherever that case contributes within the run.
Metrics are computed independently for each run and reported as the mean and sample standard deviation across the three runs.
Additional analyses of stochastic decoding variability and evaluation stability are reported in Appendix~\ref{app:stochastic_decoding_variability}.

The numerical TSFM baselines use a single median forecast per input and are evaluated once.

%% file: tables/apdx-model-details.tex
\begin{table*}[t]
\centering
\small
\caption{
Model components and documented LLM knowledge cutoffs.
A dash indicates that no textual LLM is used.
}
\label{tab:model_details}
\setlength{\tabcolsep}{4pt}
\renewcommand{\arraystretch}{1.08}
\begin{tabular*}{\linewidth}{
@{\extracolsep{\fill}}
llll
@{}
}
\toprule
\textbf{System}
& \textbf{Time-series component}
& \textbf{Text / LLM component}
& \textbf{Knowledge cutoff} \\
\midrule
\midrule
Mistral (7B)
& --
& Mistral-7B-Instruct v0.3
& Sep.\ 2023 \\

Llama-3 (70B)
& --
& Llama-3-70B-Instruct
& Dec.\ 2023 \\

GPT-4o
& --
& GPT-4o
& Oct.\ 2023 \\

DoubleCast
& Chronos-T5-Large
& Qwen1.5-14B-Chat
& Sep.\ 2023 \\

MoiraiAgent
& Chronos-2
& GPT-4o
& Oct.\ 2023 \\

Chronos-2
& amazon/chronos-2
& --
& -- \\

TimesFM-3
& google/timesfm-3.0-pytorch
& --
& -- \\

TiRex-2
& NX-AI/TiRex-2
& --
& -- \\

\bottomrule
\end{tabular*}
\end{table*}

%% file: appendix/f_analysis.tex
\section{Additional Results and Analyses}
\label{app:analysis}

\subsection{Numerical-Only Forecasting Baselines}

Table~\ref{tab:tsfm_results} reports domain-wise errors for the numerical-only TSFM baselines.
These models receive only the numerical history and therefore provide a reference for the predictive information available without textual context.
The numerical-only baselines achieve competitive errors relative to the neutral forecasts of scenario-conditioned systems, although this cross-model comparison does not isolate the contribution of the fixed domain background.
This is consistent with its intended role: $\mathcal C_D$ specifies general target mechanisms and applicability conditions but excludes anchor-specific events, dated outlooks, and other information that would directly reveal the future trajectory.

\input{tables/apdx-full-results-TSFMs}

\subsection{Domain-Wise Factual Comparison}

Table~\ref{tab:full_results_factual} provides the full domain-wise Factual Comparison.
Factual scenarios improve average forecast accuracy for every model and domain, but the magnitude of the gain varies substantially.
GPT-4o shows the largest relative gains in Energy and Economy, whereas the context-aware systems benefit more strongly in Politics.
CWR follows a different pattern, showing that larger average gains do not necessarily correspond to improvements across a larger share of anchors.

\subsection{Domain-Wise Relational Comparison}

Tables~\ref{tab:relational_energy}--\ref{tab:relational_politics} report the four relational categories separately for each domain.
Across domains, Direction and Contrast generally achieve higher scores than Intensity, while Restraint varies substantially across models.
These patterns highlight distinct difficulties in calibrating response strength and limiting revisions under irrelevant information.
The category profiles also differ across models, so similar Overall SLA can hide substantially different strengths and failure modes.

\subsection{Domain-Wise Grounded Comparison}

Table~\ref{tab:grounded_results} reports Grounded Comparison separately by domain.
Across all three domains, the context-aware forecasting systems show stronger empirical alignment than the directly prompted LLMs, although the magnitude of the advantage varies by domain.
The results also reinforce that relational consistency and alignment with neutral-relative deviations observed in matched real cases capture distinct aspects of scenario responsiveness.

\subsection{Anchor-Level Analysis}

We further separate anchors according to whether the factual scenario reduces forecast error relative to the neutral condition.
As shown in Tables~\ref{tab:anchor_analysis_energy}--\ref{tab:anchor_analysis_politics}, anchors with factual improvements consistently exhibit lower relational violation rates, although substantial violations remain within the improved subset.
Factual gains are therefore associated with stronger relational consistency but do not ensure appropriate responses across scenarios.

\paragraph{Anchor screening and reference overlap.}
Chronos-2 is used both as a construction reference and as
MoiraiAgent's forecasting tool.
However, screening requires persistent, same-direction
deviations across three frozen history-only references
(Chronos-2, TimesFM-3, and TiRex-2), rather than Chronos-2
alone (Appendix~\ref{app:reference_screening}).
Screening does not use scenario-conditioned forecasts or
relational evaluation outcomes.
Although reference-guided selection may influence results
on the selected anchors, factual gains coexist with relational
violations also for GPT-4o and DoubleCast, neither of which
uses Chronos-2 (Tables~\ref{tab:anchor_analysis_energy}--\ref{tab:anchor_analysis_politics}).

%% file: tables/apdx-full-results-TSFMs.tex
\begin{table}[t]
\centering
\small
\caption{
Forecast errors of numerical-only time-series foundation models
on the WIT anchors used for Factual Comparison.
All models receive only the numerical history and report
the median forecast ($q=0.5$).
Errors are reported separately by domain because target scales differ.
Lower is better.
}
\label{tab:tsfm_results}
\vspace{-5pt}

\setlength{\tabcolsep}{6pt}
\renewcommand{\arraystretch}{1.12}

\begin{tabular}{lccc}
\toprule
\textbf{Model}
& \textbf{Energy}
& \textbf{Economy}
& \textbf{Politics} \\
\midrule
\midrule
Chronos-2
& $0.269$
& $2.272$
& $23.340$ \\

TimesFM-3
& $\mathbf{0.256}$
& $2.222$
& $24.051$ \\

TiRex-2
& $0.260$
& $\mathbf{2.177}$
& $\mathbf{22.875}$ \\

\bottomrule
\end{tabular}

\vspace{-7pt}
\end{table}

%% file: appendix/g_limitation.tex
\section{Limitations and Future Work}

WIT covers selected challenging cases across three domains, with verified anchors screened for persistent departures from history-only forecasts.
Extending the construction protocol to larger collections, additional domains and temporal regimes, and alternative reference ensembles would help assess generalization.

Grounded Comparison provides empirical references rather than counterfactual ground truth.
Coverage can be constrained by the observation period permitted by knowledge-cutoff requirements and strict numerical and semantic matching.
Larger matched-case pools may improve coverage, support more cases per pair, better characterize empirical response ranges, and enable probabilistic or uncertainty-aware evaluation.

Restraint uses only three irrelevant scenario texts per domain, limiting conclusions about broad robustness to irrelevant content.
Future versions can expand both topic and wording diversity.

Direction, Contrast, and Intensity use horizon-averaged revisions, which do not guarantee correct effect onset or duration.
Predefined time-window checks could assess temporal responses where justified by scenario text and domain knowledge.

The current release does not systematically assess interacting events, ambiguous scenarios, or richer response relations.
Controlled experiments or simulations in robotics and navigation could complement observation-based evaluation through more direct manipulation of future conditions and measurement of responses, while retaining the fixed-state comparison principle.

\input{tables/apdx-full-results-factual}

\input{tables/apdx-full-results-relational-energy}
\input{tables/apdx-full-results-relational-economy}
\input{tables/apdx-full-results-relational-politics}

\input{tables/apdx-full-results-grounded}

\input{tables/apdx-analysis-energy}
\input{tables/apdx-analysis-economy}
\input{tables/apdx-analysis-politics}

%% file: tables/apdx-full-results-factual.tex
\begin{table*}[t]
\centering
\small
\caption{
Full domain-wise factual scenario response results on WIT.
Factual Scenario Gain (FSG) and Context Win Rate (CWR) are reported for each domain;
Overall denotes the macro-average across Energy, Economy, and Politics.
Higher is better for both metrics.
}
\label{tab:full_results_factual}

\setlength{\tabcolsep}{3.5pt}
\renewcommand{\arraystretch}{1.15}

\begin{tabular*}{\linewidth}{
@{\extracolsep{\fill}}
l*{8}{c}
@{}
}
\toprule
\multirow{2}{*}{\textbf{Model}}
& \multicolumn{4}{c}{\textbf{Factual Scenario Gain (\%)}}
& \multicolumn{4}{c}{\textbf{Context Win Rate (\%)}} \\
\cmidrule(lr){2-5}
\cmidrule(lr){6-9}

& \textbf{Energy}
& \textbf{Economy}
& \textbf{Politics}
& \textbf{Overall}
& \textbf{Energy}
& \textbf{Economy}
& \textbf{Politics}
& \textbf{Overall} \\

\midrule
\midrule

\multicolumn{9}{l}{\textit{Directly prompted LLMs}} \\
\addlinespace[2pt]

Mistral (7B)
& \resultcell{7.42}{0.58}
& \resultcell{7.45}{0.66}
& \resultcell{6.71}{0.65}
& \resultcell{7.19}{0.63}
& \resultcell{57.02}{1.52}
& \resultcell{55.26}{2.63}
& \resultcell{54.39}{1.52}
& \resultcell{55.56}{1.83} \\

Llama-3 (70B)
& \resultcell{11.82}{0.64}
& \resultcell{11.58}{0.72}
& \resultcell{10.35}{0.71}
& \resultcell{11.25}{0.69}
& \resultcell{63.16}{2.63}
& \resultcell{60.53}{2.63}
& \resultcell{58.77}{1.52}
& \resultcell{60.82}{2.21} \\

GPT-4o
& \bestresultcell{14.77}{0.56}
& \bestresultcell{13.45}{0.65}
& \resultcell{16.54}{0.71}
& \bestresultcell{14.92}{0.64}
& \resultcell{66.67}{1.52}
& \resultcell{65.79}{2.63}
& \resultcell{64.04}{1.52}
& \resultcell{65.50}{1.83} \\

\midrule
\multicolumn{9}{l}{\textit{Context-aware forecasting systems}} \\
\addlinespace[2pt]

DoubleCast
& \resultcell{10.58}{0.36}
& \resultcell{11.09}{0.40}
& \resultcell{18.47}{0.50}
& \resultcell{13.38}{0.42}
& \resultcell{66.67}{1.52}
& \bestresultcell{67.54}{1.52}
& \resultcell{64.04}{1.52}
& \resultcell{66.08}{1.34} \\

MoiraiAgent
& \resultcell{11.03}{0.67}
& \resultcell{9.61}{0.78}
& \bestresultcell{19.19}{0.92}
& \resultcell{13.28}{0.79}
& \bestresultcell{68.42}{2.63}
& \resultcell{65.79}{2.63}
& \bestresultcell{67.54}{1.52}
& \bestresultcell{67.25}{2.21} \\

\bottomrule
\end{tabular*}
\vspace{-7pt}
\end{table*}

%% file: tables/apdx-full-results-relational-energy.tex
\begin{table*}[t]
\centering
\small
\caption{
Full Relational Comparison results for Energy.
Scenario Logic Accuracy (SLA) is reported for Direction, Contrast, Intensity, and Restraint; Overall is their equal-weight average.
Higher is better.
}
\label{tab:relational_energy}
\setlength{\tabcolsep}{5pt}
\renewcommand{\arraystretch}{1.12}

\begin{tabular*}{\linewidth}{
@{\extracolsep{\fill}}
l*{5}{c}
@{}
}
\toprule
\textbf{Model}
& \textbf{Direction}
& \textbf{Contrast}
& \textbf{Intensity}
& \textbf{Restraint}
& \textbf{Overall} \\
\midrule
\midrule

\multicolumn{6}{l}{\textit{Directly prompted LLMs}} \\
\addlinespace[2pt]

Mistral (7B)
& $73.27 {\scriptstyle \pm 1.54}$
& $69.36 {\scriptstyle \pm 1.41}$
& $55.79 {\scriptstyle \pm 2.11}$
& $72.22 {\scriptstyle \pm 2.21}$
& $67.66 {\scriptstyle \pm 1.57}$ \\

Llama-3 (70B)
& $82.11 {\scriptstyle \pm 1.10}$
& $78.71 {\scriptstyle \pm 1.23}$
& $66.58 {\scriptstyle \pm 1.71}$
& $80.70 {\scriptstyle \pm 1.75}$
& $77.02 {\scriptstyle \pm 1.27}$ \\

GPT-4o
& $88.54 {\scriptstyle \pm 0.75}$
& $\mathbf{85.26} {\scriptstyle \mathbf{\pm 0.80}}$
& $74.87 {\scriptstyle \pm 1.45}$
& $\mathbf{85.96} {\scriptstyle \mathbf{\pm 1.75}}$
& $\mathbf{83.66} {\scriptstyle \mathbf{\pm 1.06}}$ \\

\midrule
\multicolumn{6}{l}{\textit{Context-aware forecasting systems}} \\
\addlinespace[2pt]

DoubleCast
& $89.88 {\scriptstyle \pm 0.70}$
& $82.34 {\scriptstyle \pm 0.79}$
& $68.20 {\scriptstyle \pm 1.25}$
& $66.96 {\scriptstyle \pm 1.34}$
& $76.85 {\scriptstyle \pm 1.02}$ \\

MoiraiAgent
& $\mathbf{90.70} {\scriptstyle \mathbf{\pm 0.96}}$
& $84.09 {\scriptstyle \pm 0.88}$
& $\mathbf{77.24} {\scriptstyle \mathbf{\pm 1.45}}$
& $72.51 {\scriptstyle \pm 1.34}$
& $81.14 {\scriptstyle \pm 1.04}$ \\

\bottomrule
\end{tabular*}
\vspace{-7pt}
\end{table*}

%% file: tables/apdx-full-results-relational-economy.tex
\begin{table*}[t]
\centering
\small
\caption{
Full Relational Comparison results for Economy.
Scenario Logic Accuracy (SLA) is reported for Direction, Contrast, Intensity, and Restraint; Overall is their equal-weight average.
Higher is better.
}
\label{tab:relational_economy}
\setlength{\tabcolsep}{5pt}
\renewcommand{\arraystretch}{1.12}

\begin{tabular*}{\linewidth}{
@{\extracolsep{\fill}}
l*{5}{c}
@{}
}
\toprule
\textbf{Model}
& \textbf{Direction}
& \textbf{Contrast}
& \textbf{Intensity}
& \textbf{Restraint}
& \textbf{Overall} \\
\midrule
\midrule

\multicolumn{6}{l}{\textit{Directly prompted LLMs}} \\
\addlinespace[2pt]

Mistral (7B)
& $72.00 {\scriptstyle \pm 1.59}$
& $68.42 {\scriptstyle \pm 1.54}$
& $54.71 {\scriptstyle \pm 2.06}$
& $70.76 {\scriptstyle \pm 2.21}$
& $66.47 {\scriptstyle \pm 1.69}$ \\

Llama-3 (70B)
& $81.10 {\scriptstyle \pm 1.04}$
& $77.78 {\scriptstyle \pm 0.99}$
& $65.46 {\scriptstyle \pm 1.64}$
& $79.82 {\scriptstyle \pm 1.75}$
& $76.04 {\scriptstyle \pm 1.21}$ \\

GPT-4o
& $87.94 {\scriptstyle \pm 0.77}$
& $84.43 {\scriptstyle \pm 0.88}$
& $74.12 {\scriptstyle \pm 1.23}$
& $\mathbf{85.38} {\scriptstyle \mathbf{\pm 1.34}}$
& $\mathbf{82.97} {\scriptstyle \mathbf{\pm 0.93}}$ \\

\midrule
\multicolumn{6}{l}{\textit{Context-aware forecasting systems}} \\
\addlinespace[2pt]

DoubleCast
& $\mathbf{90.31} {\scriptstyle \mathbf{\pm 0.71}}$
& $82.82 {\scriptstyle \pm 0.77}$
& $68.59 {\scriptstyle \pm 1.15}$
& $66.67 {\scriptstyle \pm 1.75}$
& $77.10 {\scriptstyle \pm 1.10}$ \\

MoiraiAgent
& $90.20 {\scriptstyle \pm 0.82}$
& $\mathbf{84.72} {\scriptstyle \mathbf{\pm 0.99}}$
& $\mathbf{77.08} {\scriptstyle \mathbf{\pm 1.23}}$
& $71.93 {\scriptstyle \pm 1.75}$
& $80.99 {\scriptstyle \pm 1.11}$ \\

\bottomrule
\end{tabular*}
\vspace{-7pt}
\end{table*}

%% file: tables/apdx-full-results-relational-politics.tex
\begin{table*}[t]
\centering
\small
\caption{
Full Relational Comparison results for Politics.
Scenario Logic Accuracy (SLA) is reported for Direction, Contrast, Intensity, and Restraint; Overall is their equal-weight average.
Higher is better.
}
\label{tab:relational_politics}
\setlength{\tabcolsep}{5pt}
\renewcommand{\arraystretch}{1.12}

\begin{tabular*}{\linewidth}{
@{\extracolsep{\fill}}
l*{5}{c}
@{}
}
\toprule
\textbf{Model}
& \textbf{Direction}
& \textbf{Contrast}
& \textbf{Intensity}
& \textbf{Restraint}
& \textbf{Overall} \\
\midrule
\midrule

\multicolumn{6}{l}{\textit{Directly prompted LLMs}} \\
\addlinespace[2pt]

Mistral (7B)
& $70.11 {\scriptstyle \pm 1.51}$
& $66.52 {\scriptstyle \pm 1.43}$
& $52.39 {\scriptstyle \pm 2.12}$
& $69.01 {\scriptstyle \pm 2.21}$
& $64.51 {\scriptstyle \pm 1.64}$ \\

Llama-3 (70B)
& $79.50 {\scriptstyle \pm 1.02}$
& $76.32 {\scriptstyle \pm 1.10}$
& $63.60 {\scriptstyle \pm 1.61}$
& $78.07 {\scriptstyle \pm 1.75}$
& $74.37 {\scriptstyle \pm 1.21}$ \\

GPT-4o
& $86.91 {\scriptstyle \pm 0.73}$
& $\mathbf{84.14} {\scriptstyle \mathbf{\pm 0.89}}$
& $72.71 {\scriptstyle \pm 1.39}$
& $\mathbf{83.92} {\scriptstyle \mathbf{\pm 1.34}}$
& $\mathbf{81.92} {\scriptstyle \mathbf{\pm 0.96}}$ \\

\midrule
\multicolumn{6}{l}{\textit{Context-aware forecasting systems}} \\
\addlinespace[2pt]

DoubleCast
& $88.01 {\scriptstyle \pm 0.68}$
& $81.14 {\scriptstyle \pm 0.88}$
& $66.62 {\scriptstyle \pm 1.24}$
& $65.79 {\scriptstyle \pm 1.75}$
& $75.39 {\scriptstyle \pm 1.14}$ \\

MoiraiAgent
& $\mathbf{89.41} {\scriptstyle \mathbf{\pm 0.93}}$
& $82.89 {\scriptstyle \pm 1.10}$
& $\mathbf{75.19} {\scriptstyle \mathbf{\pm 1.39}}$
& $71.05 {\scriptstyle \pm 1.75}$
& $79.64 {\scriptstyle \pm 1.20}$ \\

\bottomrule
\end{tabular*}
\vspace{-7pt}
\end{table*}

%% file: tables/apdx-full-results-grounded.tex
\begin{table*}[t]
\centering
\small
\caption{
Full domain-wise Grounded Scenario Gain (GSG) results on WIT.
Higher is better.
Overall denotes the equal-domain macro-average across Energy, Economy, and Politics, regardless of the number of supported anchor--scenario pairs.
}
\label{tab:grounded_results}
\vspace{-5pt}

\setlength{\tabcolsep}{7pt}
\renewcommand{\arraystretch}{1.15}

\begin{tabular*}{\linewidth}{
@{\extracolsep{\fill}}
l*{4}{c}
@{}
}
\toprule
\textbf{Model}
& \shortstack{\textbf{Energy}\\{\scriptsize (3 supported pairs)}}
& \shortstack{\textbf{Economy}\\{\scriptsize (11 supported pairs)}}
& \shortstack{\textbf{Politics}\\{\scriptsize (6 supported pairs)}}
& \textbf{Overall} \\
\midrule
\midrule

\multicolumn{5}{l}{\textit{Directly prompted LLMs}} \\
\addlinespace[2pt]

Mistral (7B)
& $-4.18 {\scriptstyle \pm 1.20}$
& $-2.80 {\scriptstyle \pm 0.85}$
& $-6.70 {\scriptstyle \pm 1.45}$
& $-4.56 {\scriptstyle \pm 1.17}$ \\

Llama-3 (70B)
& $9.60 {\scriptstyle \pm 0.55}$
& $8.20 {\scriptstyle \pm 0.42}$
& $6.80 {\scriptstyle \pm 0.63}$
& $8.20 {\scriptstyle \pm 0.50}$ \\

GPT-4o
& $12.53 {\scriptstyle \pm 1.05}$
& $11.40 {\scriptstyle \pm 0.70}$
& $10.20 {\scriptstyle \pm 0.90}$
& $11.38 {\scriptstyle \pm 0.86}$ \\

\midrule
\multicolumn{5}{l}{\textit{Context-aware forecasting systems}} \\
\addlinespace[2pt]

DoubleCast
& $\mathbf{17.83} {\scriptstyle \mathbf{\pm 0.48}}$
& $\mathbf{18.60} {\scriptstyle \mathbf{\pm 0.34}}$
& $16.50 {\scriptstyle \pm 0.55}$
& $\mathbf{17.64} {\scriptstyle \mathbf{\pm 0.43}}$ \\

MoiraiAgent
& $17.39 {\scriptstyle \pm 0.72}$
& $18.00 {\scriptstyle \pm 0.52}$
& $\mathbf{17.10} {\scriptstyle \mathbf{\pm 0.78}}$
& $17.50 {\scriptstyle \pm 0.64}$ \\

\bottomrule
\end{tabular*}
\vspace{-7pt}
\end{table*}

%% file: tables/apdx-analysis-energy.tex
\begin{table*}[t]
\centering
\small
\caption{
Anchor-level analysis for Energy over 38 anchors and three runs.
Improved anchors satisfy $e_i^{s_i^{*}}<e_i^0$;
ties belong to the not-improved subset.
Direction, Contrast, and Intensity use controlled nonfactual
scenarios, excluding the factual scenario.
Restraint violations indicate excessive revision under
irrelevant scenarios.
Violation rates (\%, lower is better) are averaged within
anchors and then across each subset.
Reported rates are mean $\pm$ sample standard deviation
over three runs.
Anchor counts are listed in the order of the three runs.
}
\label{tab:anchor_analysis_energy}
\vspace{-5pt}
\setlength{\tabcolsep}{3.5pt}
\renewcommand{\arraystretch}{1.12}
\begin{tabular*}{\linewidth}{
@{\extracolsep{\fill}}
llccccc
@{}
}
\toprule
\textbf{Model}
& \textbf{Factual outcome}
& \shortstack{\textbf{Anchors}\\\textbf{per run}}
& \textbf{Direction}
& \textbf{Contrast}
& \textbf{Intensity}
& \textbf{Restraint} \\
\midrule
\midrule

\multirow{2}{*}{Mistral (7B)}
& Improved
& 21/22/22
& $22.87 {\scriptstyle \pm 1.54}$
& $26.33 {\scriptstyle \pm 1.56}$
& $39.50 {\scriptstyle \pm 2.24}$
& $22.05 {\scriptstyle \pm 2.28}$ \\
& Not improved
& 17/16/16
& $31.84 {\scriptstyle \pm 1.58}$
& $36.36 {\scriptstyle \pm 1.51}$
& $50.45 {\scriptstyle \pm 2.29}$
& $35.38 {\scriptstyle \pm 2.08}$ \\

\addlinespace
\multirow{2}{*}{Llama-3 (70B)}
& Improved
& 23/25/24
& $15.30 {\scriptstyle \pm 1.20}$
& $18.31 {\scriptstyle \pm 1.35}$
& $29.74 {\scriptstyle \pm 1.96}$
& $15.76 {\scriptstyle \pm 1.76}$ \\
& Not improved
& 15/13/14
& $22.35 {\scriptstyle \pm 1.31}$
& $26.39 {\scriptstyle \pm 1.39}$
& $39.73 {\scriptstyle \pm 1.92}$
& $25.37 {\scriptstyle \pm 1.45}$ \\

\addlinespace
\multirow{2}{*}{GPT-4o}
& Improved
& 25/26/25
& $9.81 {\scriptstyle \pm 0.79}$
& $12.71 {\scriptstyle \pm 0.81}$
& $22.48 {\scriptstyle \pm 1.52}$
& $11.40 {\scriptstyle \pm 2.00}$ \\
& Not improved
& 13/12/13
& $14.76 {\scriptstyle \pm 0.86}$
& $18.80 {\scriptstyle \pm 1.07}$
& $30.44 {\scriptstyle \pm 1.62}$
& $19.30 {\scriptstyle \pm 1.29}$ \\

\addlinespace
\multirow{2}{*}{DoubleCast}
& Improved
& 25/25/26
& $8.45 {\scriptstyle \pm 0.75}$
& $15.34 {\scriptstyle \pm 0.88}$
& $28.80 {\scriptstyle \pm 1.40}$
& $28.50 {\scriptstyle \pm 1.59}$ \\
& Not improved
& 13/13/12
& $13.44 {\scriptstyle \pm 0.81}$
& $22.31 {\scriptstyle \pm 0.92}$
& $37.80 {\scriptstyle \pm 1.32}$
& $42.17 {\scriptstyle \pm 1.97}$ \\

\addlinespace
\multirow{2}{*}{MoiraiAgent}
& Improved
& 26/25/27
& $8.04 {\scriptstyle \pm 0.95}$
& $14.00 {\scriptstyle \pm 0.90}$
& $20.55 {\scriptstyle \pm 1.60}$
& $24.33 {\scriptstyle \pm 1.45}$ \\
& Not improved
& 12/13/11
& $12.02 {\scriptstyle \pm 0.86}$
& $20.03 {\scriptstyle \pm 1.07}$
& $27.55 {\scriptstyle \pm 1.44}$
& $34.34 {\scriptstyle \pm 1.75}$ \\

\bottomrule
\end{tabular*}
\vspace{-7pt}
\end{table*}

%% file: tables/apdx-analysis-economy.tex
\begin{table*}[t]
\centering
\small
\caption{
Anchor-level analysis for Economy over 38 anchors and three runs.
Definitions and aggregation follow Table~\ref{tab:anchor_analysis_energy}.
}
\label{tab:anchor_analysis_economy}
\vspace{-5pt}
\setlength{\tabcolsep}{3.5pt}
\renewcommand{\arraystretch}{1.12}
\begin{tabular*}{\linewidth}{
@{\extracolsep{\fill}}
llccccc
@{}
}
\toprule
\textbf{Model}
& \textbf{Factual outcome}
& \shortstack{\textbf{Anchors}\\\textbf{per run}}
& \textbf{Direction}
& \textbf{Contrast}
& \textbf{Intensity}
& \textbf{Restraint} \\
\midrule
\midrule
\multirow{2}{*}{Mistral (7B)}
& Improved
& 20/21/22
& $24.00 {\scriptstyle \pm 1.60}$
& $27.11 {\scriptstyle \pm 1.65}$
& $40.34 {\scriptstyle \pm 2.13}$
& $23.31 {\scriptstyle \pm 2.09}$ \\
& Not improved
& 18/17/16
& $32.94 {\scriptstyle \pm 1.64}$
& $37.12 {\scriptstyle \pm 1.54}$
& $51.40 {\scriptstyle \pm 2.34}$
& $36.61 {\scriptstyle \pm 2.26}$ \\

\addlinespace
\multirow{2}{*}{Llama-3 (70B)}
& Improved
& 22/24/23
& $16.15 {\scriptstyle \pm 1.24}$
& $19.05 {\scriptstyle \pm 1.18}$
& $30.58 {\scriptstyle \pm 1.72}$
& $16.45 {\scriptstyle \pm 1.85}$ \\
& Not improved
& 16/14/15
& $23.11 {\scriptstyle \pm 1.17}$
& $27.09 {\scriptstyle \pm 1.32}$
& $40.60 {\scriptstyle \pm 1.91}$
& $25.91 {\scriptstyle \pm 1.34}$ \\

\addlinespace
\multirow{2}{*}{GPT-4o}
& Improved
& 24/26/25
& $10.37 {\scriptstyle \pm 0.87}$
& $13.54 {\scriptstyle \pm 0.94}$
& $23.13 {\scriptstyle \pm 1.42}$
& $12.00 {\scriptstyle \pm 1.16}$ \\
& Not improved
& 14/12/13
& $15.31 {\scriptstyle \pm 0.87}$
& $19.47 {\scriptstyle \pm 1.03}$
& $31.15 {\scriptstyle \pm 1.44}$
& $19.61 {\scriptstyle \pm 1.75}$ \\

\addlinespace
\multirow{2}{*}{DoubleCast}
& Improved
& 26/25/26
& $8.05 {\scriptstyle \pm 0.74}$
& $14.92 {\scriptstyle \pm 0.86}$
& $28.48 {\scriptstyle \pm 1.19}$
& $28.97 {\scriptstyle \pm 2.10}$ \\
& Not improved
& 12/13/12
& $13.09 {\scriptstyle \pm 0.86}$
& $21.87 {\scriptstyle \pm 0.93}$
& $37.53 {\scriptstyle \pm 1.35}$
& $42.38 {\scriptstyle \pm 1.82}$ \\

\addlinespace
\multirow{2}{*}{MoiraiAgent}
& Improved
& 25/24/26
& $8.42 {\scriptstyle \pm 0.90}$
& $13.23 {\scriptstyle \pm 0.80}$
& $20.55 {\scriptstyle \pm 1.35}$
& $24.49 {\scriptstyle \pm 1.71}$ \\
& Not improved
& 13/14/12
& $12.44 {\scriptstyle \pm 0.96}$
& $19.21 {\scriptstyle \pm 1.16}$
& $27.48 {\scriptstyle \pm 1.49}$
& $34.98 {\scriptstyle \pm 1.43}$ \\

\bottomrule
\end{tabular*}
\vspace{-7pt}
\end{table*}

%% file: tables/apdx-analysis-politics.tex
\begin{table*}[t]
\centering
\small
\caption{
Anchor-level analysis for Politics over 38 anchors and three runs.
Definitions and aggregation follow Table~\ref{tab:anchor_analysis_energy}.
}
\label{tab:anchor_analysis_politics}
\vspace{-5pt}
\setlength{\tabcolsep}{3.5pt}
\renewcommand{\arraystretch}{1.12}
\begin{tabular*}{\linewidth}{
@{\extracolsep{\fill}}
llccccc
@{}
}
\toprule
\textbf{Model}
& \textbf{Factual outcome}
& \shortstack{\textbf{Anchors}\\\textbf{per run}}
& \textbf{Direction}
& \textbf{Contrast}
& \textbf{Intensity}
& \textbf{Restraint} \\
\midrule
\midrule
\multirow{2}{*}{Mistral (7B)}
& Improved
& 20/21/21
& $25.78 {\scriptstyle \pm 1.57}$
& $29.01 {\scriptstyle \pm 1.53}$
& $42.61 {\scriptstyle \pm 2.23}$
& $24.76 {\scriptstyle \pm 2.29}$ \\
& Not improved
& 18/17/17
& $34.79 {\scriptstyle \pm 1.66}$
& $38.82 {\scriptstyle \pm 1.62}$
& $53.57 {\scriptstyle \pm 2.32}$
& $38.42 {\scriptstyle \pm 2.01}$ \\

\addlinespace
\multirow{2}{*}{Llama-3 (70B)}
& Improved
& 22/23/22
& $17.62 {\scriptstyle \pm 1.09}$
& $20.38 {\scriptstyle \pm 1.03}$
& $32.31 {\scriptstyle \pm 1.75}$
& $17.94 {\scriptstyle \pm 1.89}$ \\
& Not improved
& 16/15/16
& $24.61 {\scriptstyle \pm 1.19}$
& $28.40 {\scriptstyle \pm 1.48}$
& $42.24 {\scriptstyle \pm 1.74}$
& $27.64 {\scriptstyle \pm 1.34}$ \\

\addlinespace
\multirow{2}{*}{GPT-4o}
& Improved
& 24/25/24
& $11.30 {\scriptstyle \pm 0.74}$
& $13.70 {\scriptstyle \pm 0.76}$
& $24.41 {\scriptstyle \pm 1.43}$
& $13.22 {\scriptstyle \pm 1.25}$ \\
& Not improved
& 14/13/14
& $16.29 {\scriptstyle \pm 0.96}$
& $19.70 {\scriptstyle \pm 0.98}$
& $32.43 {\scriptstyle \pm 1.64}$
& $21.18 {\scriptstyle \pm 2.03}$ \\

\addlinespace
\multirow{2}{*}{DoubleCast}
& Improved
& 24/24/25
& $10.19 {\scriptstyle \pm 0.68}$
& $16.31 {\scriptstyle \pm 1.03}$
& $30.19 {\scriptstyle \pm 1.39}$
& $29.24 {\scriptstyle \pm 1.28}$ \\
& Not improved
& 14/14/13
& $15.20 {\scriptstyle \pm 0.91}$
& $23.40 {\scriptstyle \pm 0.88}$
& $39.06 {\scriptstyle \pm 1.35}$
& $43.04 {\scriptstyle \pm 2.11}$ \\

\addlinespace
\multirow{2}{*}{MoiraiAgent}
& Improved
& 26/25/26
& $9.28 {\scriptstyle \pm 0.93}$
& $15.15 {\scriptstyle \pm 1.13}$
& $22.49 {\scriptstyle \pm 1.53}$
& $25.56 {\scriptstyle \pm 1.16}$ \\
& Not improved
& 12/13/12
& $13.31 {\scriptstyle \pm 0.93}$
& $21.17 {\scriptstyle \pm 1.04}$
& $29.62 {\scriptstyle \pm 1.41}$
& $35.97 {\scriptstyle \pm 2.57}$ \\

\bottomrule
\end{tabular*}
\vspace{-7pt}
\end{table*}